\PassOptionsToPackage{table}{xcolor}
\documentclass[10pt]{article} % For LaTeX2e
\usepackage[preprint]{tmlr}
\usepackage{amsmath,amsfonts,bm}

\def\eqref#1{equation~\ref{#1}}
\def\1{\bm{1}}

\DeclareMathAlphabet{\mathsfit}{\encodingdefault}{\sfdefault}{m}{sl}
\SetMathAlphabet{\mathsfit}{bold}{\encodingdefault}{\sfdefault}{bx}{n}

\definecolor{navyblue}{rgb}{0.0,0.0,0.5}

\usepackage{hyperref}
\hypersetup{
  colorlinks=true,
  linkcolor=navyblue,
  citecolor=navyblue,
  urlcolor=navyblue
}
\usepackage{url}
\usepackage{booktabs}
\usepackage{xcolor}
\usepackage{makecell}
\usepackage{adjustbox}
\usepackage{amssymb}
\usepackage{graphicx} % for \scalebox
\usepackage{longtable}
\usepackage{multirow}
\usepackage{array}
\usepackage{bbm}
 \usepackage{enumitem}
\usepackage{amsthm}
\newtheorem{claim}{Claim}
\newtheorem{lemma}[claim]{Lemma}

\newtheorem{corollary}[claim]{Corollary}
\newtheorem{definition}[claim]{Definition}
\newtheorem{proposition}[claim]{Proposition}

\definecolor{blue}{HTML}{33598f}
\definecolor{bluebg}{HTML}{DFE5EE}
\definecolor{green}{HTML}{3A7018}
\definecolor{greenbg}{HTML}{E1EADC}
\definecolor{red}{HTML}{BA4242}
\definecolor{redbg}{HTML}{F5E3E3}
\definecolor{purple}{HTML}{825580}
\definecolor{purplebg}{HTML}{ECE6EC}
\definecolor{yellow}{HTML}{C49520}
\definecolor{yellowbg}{HTML}{F6EFDE}
\definecolor{teal}{HTML}{1f747a}
\definecolor{tealbg}{HTML}{DCE9E6}
\definecolor{orange}{HTML}{945A33}
\definecolor{orangebg}{HTML}{f5e7e0}
\definecolor{navy}{HTML}{1c1c4c}
\definecolor{navybg}{HTML}{DFE2EC}
\definecolor{sky}{HTML}{4d839e}
\definecolor{skybg}{HTML}{E1EEF4}
\definecolor{slate}{HTML}{3A3D42}
\definecolor{slatebg}{HTML}{E4E5E6}
\definecolor{darkbrown}{HTML}{5C4033}
\definecolor{darkbrownbg}{HTML}{EDE6E1}
\definecolor{violet}{HTML}{60468d}
\definecolor{violetbg}{HTML}{E9E4F0}

\newcommand{\ical}{{\tiny\sffamily\textcolor{teal}{\textbf{CAL}}}\,} %calibration
\newcommand{\irnk}{{\tiny\sffamily\textcolor{navy}{\textbf{RNK}}}\,} %ranking
\newcommand{\iscr}{{\tiny\sffamily\textcolor{yellow}{\textbf{SCR}}}\,} %scoring
\newcommand{\ifth}{{\tiny\sffamily\textcolor{sky}{\textbf{FTH}}}\,} %faithfulness
\newcommand{\idec}{{\tiny\sffamily\textcolor{violet}{\textbf{DEC}}}\,} %decision
\newcommand{\ireg}{{\tiny\sffamily\textcolor{orange}{\textbf{REG}}}\,} %regression
\newcommand{\icov}{{\tiny\sffamily\textcolor{slate}{\textbf{COV}}}\,} %coverage

\newcommand{\cCOR}{{\tiny\sffamily\textcolor{navy}{\textbf{COR}}}\,} %correctness
\newcommand{\cFTH}{{\tiny\sffamily\textcolor{sky}{\textbf{FTH}}}\,} %faithfulness
\newcommand{\cGRD}{{\tiny\sffamily\textcolor{teal}{\textbf{GRD}}}\,} %grounding
\newcommand{\cUNS}{{\tiny\sffamily\textcolor{yellow}{\textbf{UNS}}}\,} %underspecification

\usepackage{xcolor}

\usepackage[most]{tcolorbox}
\definecolor{panelframe}{HTML}{CBD3DC} 
\newtcolorbox[auto counter]{takeaway}[1]{%
  enhanced, breakable,
  colback=navybg!50, colframe=navy!70,
  coltext=black,
  boxrule=0.8pt, arc=4pt,
  left=5pt, right=5pt, top=3pt, bottom=3pt,
  toptitle=0.5pt, bottomtitle=0.5pt,
  before skip=7pt, after skip=7pt,
  fonttitle=\bfseries, coltitle=white, colbacktitle=navy!70,
  title={Takeaway~\thetcbcounter: #1}%
}

\newcommand{\bsrv}{$\circ$}        % survey of benchmarks/datasets
\newcommand{\bevl}{$\bullet$}      % empirical benchmark evaluation

\newcommand{\accB}{{\scriptsize$\blacksquare$}} % black-box
\newcommand{\accG}{{\scriptsize$\boxtimes$}}    % gray-box
\newcommand{\accW}{{\scriptsize$\square$}}      % white-box

\title{Where Is Uncertainty Observed? A Mechanistic\\ Understanding of Uncertainty Quantification in LLMs}

\author{%
  \begin{center}
  \name Xiaoyu Liang\textsuperscript{1} \hspace{1.2em}
  \name Virginia Ceccatelli\textsuperscript{1} \hspace{1.2em}
  \name Aliai Eusebi\textsuperscript{1} \\[2pt]
  \name Tobias Braun\textsuperscript{1,2} \hspace{1.2em}
  \name Nils Loose\textsuperscript{1,3} \hspace{1.2em}
  \name Carlo Malagnino\textsuperscript{1} \hspace{1.2em}
  \name Lorenzo Cavallaro\textsuperscript{1,4} \\[8pt]
  {\normalfont\normalsize
  \textsuperscript{1}University College London \hspace{1em}
  \textsuperscript{2}Technische Universit\"at Darmstadt \hspace{1em} \\
  \textsuperscript{3}Universit\"at zu L\"ubeck \hspace{1em} 
  \textsuperscript{4}Mohamed bin Zayed University of Artificial Intelligence}
  \end{center}%
  }

\def\month{MM}  % Insert correct month for camera-ready version
\def\year{YYYY} % Insert correct year for camera-ready version
\def\openreview{\url{https://openreview.net/forum?id=XXXX}} % Insert correct link to OpenReview for camera-ready version

\begin{document}

\maketitle

\begin{center}\url{https://s2lab.org/UQ-Lenses}\end{center}

\begin{abstract}
Knowing when a large language model's (LLM's) response can be trusted is critical for its safe use. Uncertainty quantification (UQ) can help LLMs recognize if their responses may be unreliable. Existing taxonomies organize UQ methods by uncertainty source or estimation mechanism. They leave unresolved the question of \emph{where uncertainty is observed in the model's computation}. The answer matters because the observation point determines whether an estimator has access to enough information to recover a given target, and how well it can recover it. Drawing on mechanistic interpretability, we define four mechanistic observation lenses: the \emph{structural} lens observes component updates, the \emph{representational} lens observes hidden states, the \emph{distributional} lens observes token probabilities, and the \emph{semantical} lens observes distributions over meanings. We relate these lenses through the residual stream and order them by the information they carry, which gives a principle for selecting UQ methods for different targets. We survey existing methods through the four lenses. Furthermore, we distinguish evaluation targets, criteria, and metrics to clarify different UQ evaluation protocols. Finally, we identify open problems and research directions to connect uncertainty estimation to the internal computation and downstream use of LLMs.
\end{abstract}
% The answer matters because the observation point determines whether the observation carries enough information to recover a given target, and how well an estimator can recover it.
%% ─── Body ────────────────────────────────────────────────────────────
% !TEX root = tmlr.tex
\section{Introduction}\label{sec:introduction}

Large language models (LLMs) have shown strong performance on challenging tasks in mathematics, reasoning, and coding~\citep{guoDeepSeekR1IncentivizesReasoning2025, lewkowyczSolvingQuantitativeReasoning2022, cobbeTrainingVerifiersSolve2021,chenEvaluatingLargeLanguage2021}. Recently, LLMs have been used to find counterexamples to the Jacobian conjecture~\citep{gaoCounterexamplesJacobianConjecture2026}, address the Millennium Prize Navier--Stokes existence and smoothness problem~\citep{openaiNavierStokesMillennium2026, claymathematicsinstituteMillenniumPrizeProblems2000}, and improve their own code~\citep{zhangDarwinGodelMachine2026}. However, LLMs may generate \emph{hallucinations}, plausible content that conflicts with established facts or is unsupported by the provided context~\citep{huangSurveyHallucinationLarge2025, kangUncertaintyQuantificationHallucination2026}. Unreliable responses of LLMs can be particularly harmful when models are used for high-stakes applications such as medical diagnosis~\citep{hagerEvaluationMitigationLimitations2024}, legal advice~\citep{dahlLargeLegalFictions2024}, or scientific research~\citep{boikoAutonomousChemicalResearch2023}. Knowing when an LLM's response can be trusted is critical for its safe use~\citep{jiangTrustNotTrust2018, europeanparliamentandofthecounciloftheeuropeanunionRegulationEU20242026}.

\begin{figure}[ht!]
    \centering
    \includegraphics[width=0.88\linewidth]{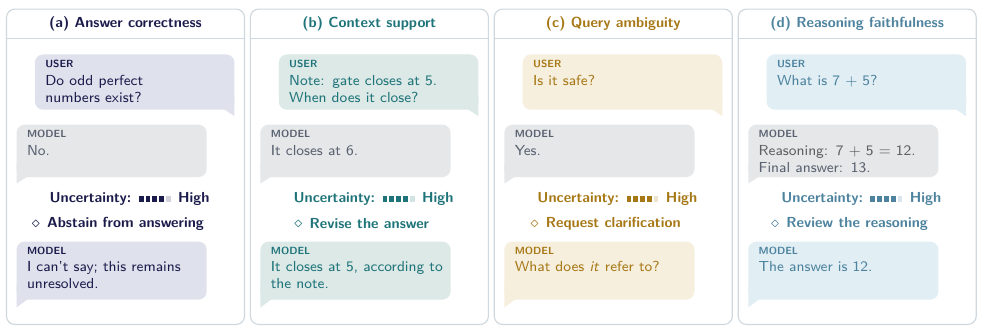}
    \caption{Illustrative failure modes and uncertainty-guided actions: (a) abstain when uncertain about answer reliability; (b) revise an answer when it may not be supported by the context; (c) request clarification when the query is underspecified; and (d) review reasoning when uncertain about the reasoning process.}
    \label{fig:uq-use-cases}
\end{figure}

Prior work has developed methods to detect hallucinations and reduce downstream decision errors~\citep{chuangLookbackLensDetecting2024, vishwakarmaPrunePredictOptimizing2025}. LLMs have been shown to be overconfident in their outputs, generating plausible yet unreliable responses~\citep{xiongCanLLMsExpress2024}. Uncertainty quantification (UQ) is used to improve the trustworthiness of LLMs by measuring the model's level of uncertainty about its response~\citep{hullermeierAleatoricEpistemicUncertainty2021, heSurveyUncertaintyQuantification2026, liuUncertaintyQuantificationConfidence2025, kirchhofPositionUncertaintyQuantification2025, shorinwaSurveyUncertaintyQuantification2025}. The signal of model uncertainty can be used for downstream actions such as abstaining~\citep{tomaniUncertaintyBasedAbstentionLLMs2024}, grounding an answer in context~\citep{chuangLookbackLensDetecting2024}, requesting clarification~\citep{zhangClarifyWhenNecessary2025}, or invoking extended reasoning~\citep{yinReasoningFluxEnhancing2024}. Figure~\ref{fig:uq-use-cases} illustrates four such uses in LLM systems. 

\begin{figure}[ht!]
    \centering
    \includegraphics[width=0.87\linewidth]{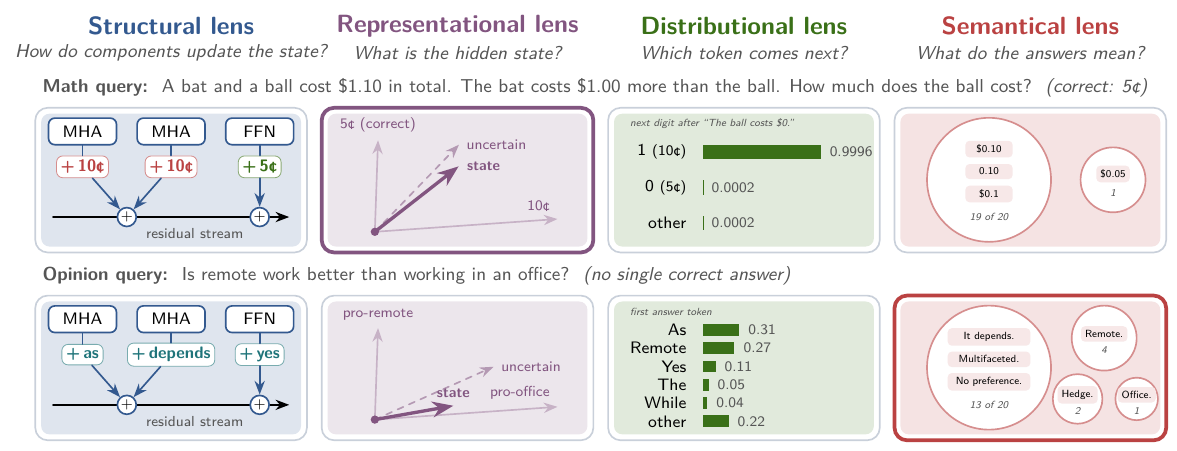}
    \caption{Four lenses applied to two queries. For the math query, the sampled answers agree on a wrong answer, so semantic uncertainty is low, but the hidden state lies near the uncertain direction. For the opinion query, the token distribution spreads over many phrasings, but the answers share one stance. A colored frame marks the suited lens. Illustrative values from an experiment on Qwen2.5-0.5B-Instruct~\citep{qwenQwen25TechnicalReport2025}.}
    \label{fig:intro-lenses}
\end{figure}

Prior surveys of UQ for LLMs organize methods by the source of uncertainty~\citep{heSurveyUncertaintyQuantification2026}, how uncertainty arises during model development and interaction~\citep{kirchhofPositionUncertaintyQuantification2025, beigiRethinkingUncertaintyCritical2024}, or the estimation mechanism~\citep{shorinwaSurveyUncertaintyQuantification2025}. These taxonomies leave open where in the \emph{model computation} uncertainty is observed. Estimators read their signal at different points, and each point reveals different information. As shown in Figure~\ref{fig:intro-lenses}, for a math question, only observing the meaning misses the uncertain hidden state~\citep{chenLLMsInternalStates2024}. For an opinion question, only observing the output tokens mostly reads lexical differences and misses the shared stance~\citep{kuhnSemanticUncertaintyLinguistic2023}. We therefore ask: \emph{where in the model's computation should uncertainty be observed, and what information does each observation point reveal?} The answer matters because the observation point determines whether a target can be recovered and how well.

We address this question with a mechanistic framework for UQ~\citep{barezUnderstandingAIState2026, bereskaMechanisticInterpretabilityAI2024}. We operationalize uncertainty as the dispersion of the model's predictive state for a query, observed at four successive points in the model's computation. We call these points mechanistic observation lenses: \emph{structural}, \emph{representational}, \emph{distributional}, and \emph{semantical}. The four lenses are connected through the residual stream~\citep{elhageMathematicalTransformerCircuits2021}. The structural and representational lenses observe how the residual stream is updated and the residual stream state, respectively. The distributional and semantical lenses observe the residual stream readout distribution over vocabulary tokens and over meanings, respectively. The four lenses are ordered by the information they carry, which we examine empirically on correctness prediction.

Based on the four lenses, we build a framework to characterize uncertainty estimators by the lens they apply, the variables they sample across repeated inferences, and the auxiliary structure they rely on. The lens determines \emph{sufficiency}, and the sampling and auxiliary structure determine \emph{efficiency}. Probing methods read uncertainty from the hidden state, adding supervision and training cost~\citep{azariaInternalStateLLM2023, hanSemanticEntropyProbes2024}, while semantics methods require resampling~\citep{farquharDetectingHallucinationsLarge2024}. The suited lens is sufficient for the target and admits an efficient estimator. We survey existing uncertainty estimators using the three axes. The estimators can target different quantities and require different evaluation criteria, but existing work largely evaluates uncertainty for correctness prediction~\citep{gengSurveyConfidenceEstimation2024,vashurinBenchmarkingUncertaintyQuantification2025}. We identify specific evaluation targets, criteria, and metrics, explain how they relate, and survey UQ methods accordingly. We then discuss open problems and directions for future UQ research.

In summary, our contributions are as follows. We (1) identify four mechanistic observation lenses on the residual stream, each revealing different information about uncertainty, and show that they are ordered by the information they carry (Section~\ref{sec:lenses}); (2) characterize uncertainty estimators along three axes, the lens, the variation, and the auxiliary structure, and survey existing estimators along these axes (Section~\ref{sec:survey}); (3) identify evaluation targets, criteria, and metrics for uncertainty estimation, and survey existing evaluation practice accordingly (Section~\ref{sec:evaluation}); and (4) discuss open problems and future directions for UQ in LLMs (Section~\ref{sec:discussion}).

% !TEX root = tmlr.tex
\section{Background}\label{sec:background}

We begin with the background on UQ in LLMs. We first define uncertainty and its evaluation in Sections~\ref{sec:background-uncertainty} and~\ref{sec:background-eval}. We then introduce the LLM preliminaries in Section~\ref{sec:background-llm}, which provide the foundation for the four lenses defined in Section~\ref{sec:lenses}. Table~\ref{tab:notation} summarizes the notation used throughout the paper.

\begin{table}[ht]
\centering
\caption{Summary of notation.}\label{tab:notation}
\renewcommand{\arraystretch}{1.3}
\setlength{\tabcolsep}{6pt}
\begin{tabular}{c p{0.66\linewidth}}
\toprule
\textsc{Symbol} & \textsc{Meaning} \\
\midrule
$\mathcal X, \mathcal Y, \mathcal H, \mathcal P$ & Input space, output space, hypothesis space, probability distribution\\
$f_\theta: \mathcal X \to \mathcal Y$ & The LLM, parameterized by $\theta$ \\
$x$,\ $y$ & Input and output sequence, with $x^t$ and $y^t$ as the token at position $t$ \\
$h^{(\ell)} \in \mathbb{R}^d$ & Residual-stream state after block $\ell$, of hidden dimension $d$ \\
$\Delta h_c^{(\ell)}$ & Update written to the residual stream by component $c$ at layer $\ell$ \\
$\ell\in\{0,\dots,L\}$ & Residual-stream index, $\ell=0$ is the embedding output \\
$c\in\{0,\dots,C\}$ & Transformer component index, an attention head or the FFN \\
$t\in\{1,\dots,T\}$ & Token position index \\
$W_U$ & Unembedding matrix \\
$\mathrm{norm}$ & Final normalization applied before the unembedding \\
$\rho$ & Readout map applied before the unembedding, $\rho=I$ by default \\
$\Pi(h^{(\ell)})$ & Token distribution readout $\mathrm{softmax}(W_U\,\mathrm{norm}(\rho h^{(\ell)}))$ of a hidden state \\
$V$,\ $V^*$ & Token vocabulary; sequences over the vocabulary \\
$R: V^*\times V^* \to [0,1]\,\,\text{or}\,\,\mathbb R$ & Externally defined semantic relation on vocabulary sequences \\
$\mathcal{M},\ \mathcal{M}_R$ & Meaning space of outputs, set of meanings defined by $R$ \\
$\Pi_R(h^{(\ell)})$ & Semantic distribution readout of a hidden state \\
$\lambda\in\Lambda$ & Observation lens, in the lens set $\Lambda=\{\text{str,\,rep,\,dis,\,sem}\}$ \\
$O_\lambda\in\mathcal O_\lambda$ & Observation at lens $\lambda$, in its observation space $\mathcal O_\lambda$ \\
$g_\text{sum}$,\ $g_\text{unemb}$,\ $g_R$ & Lens maps $O_\text{str}\to O_\text{rep}\to O_\text{dis}\to O_\text{sem}$ in Section~\ref{sec:lenses-hierarchy} \\
$\mathcal{U}(\cdot)$ & Uncertainty function applied to the observation \\
$U_\lambda(x)$ & Uncertainty estimator given input $x$ reading lens $\lambda$ \\
$Z$ & Variant across the sampled observations \\
$S$ & Auxiliary structure used for measuring uncertainty \\
$Z_S$,\ $\rho_S$,\ $\mathcal U_S$,\ $\psi_S$ &  Four auxiliary structures $S$ in Section~\ref{sec:auxiliary-axis} \\
\bottomrule
\end{tabular}
\end{table}

\subsection{Uncertainty Definition}\label{sec:background-uncertainty}

In this work, we operationalize uncertainty as the dispersion of a model's predictive state. In risk analysis, uncertainty is generally categorized as either aleatoric or epistemic~\citep{kiureghianAleatoryEpistemicDoes2009}. Aleatoric uncertainty arises from randomness that a given model cannot resolve and is irreducible, whereas epistemic uncertainty arises from a lack of knowledge of the underlying input-output relationship and is reducible with more data. This characterization was carried over to machine learning models~\citep{kendallWhatUncertaintiesWe2017} through Bayesian deep learning~\citep{nealBayesianLearningNeural2012}. \citet{hullermeierAleatoricEpistemicUncertainty2021} stress that the two notions are defined only relative to a context setting, given by an input space $\mathcal X$, an output space $\mathcal Y$, a hypothesis space $\mathcal H$, and a probability distribution $\mathcal P$ over $\mathcal X \times \mathcal Y$. For example, two classes that overlap under a single feature exhibit aleatoric uncertainty, but may become separable with an added second feature, turning it into epistemic uncertainty~\citep{hullermeierAleatoricEpistemicUncertainty2021}. The boundary between aleatoric and epistemic (A/E) uncertainty moves with the context setting $(\mathcal X, \mathcal Y, \mathcal H, \mathcal P)$.

With a fixed context, the Bayesian decomposition~\citep{malininUncertaintyEstimationAutoregressive2021} splits total uncertainty into an aleatoric term, the expected conditional entropy, and an epistemic term, the mutual information between the output and the parameters. Formally, $H[y \mid x] = \mathbb{E}_{p(\theta)}[H[y \mid x, \theta]] + \mathbb{I}[y;\theta \mid x]$. This decomposition requires a distribution over model parameters $p(\theta)$, usually approximated by Bayesian neural networks~\citep{nealBayesianLearningNeural2012}. However, the parameters of LLMs are extremely large and often inaccessible for approximation. Recent works~\citep{beigiRethinkingUncertaintyCritical2024, kirchhofPositionUncertaintyQuantification2025, baanUncertaintyNaturalLanguage2023} criticized the classic Bayesian reading of A/E uncertainty in LLMs and argued for new definitions of uncertainty types, which we discuss in Section~\ref{sec:related}.

This work does not focus on the A/E distinction. The distinction concerns the \emph{type} of uncertainty, and it has already been examined in depth, both in classical Bayesian form and in the revisions proposed for LLMs~\citep{hullermeierAleatoricEpistemicUncertainty2021, beigiRethinkingUncertaintyCritical2024, kirchhofPositionUncertaintyQuantification2025, baanUncertaintyNaturalLanguage2023, shorinwaSurveyUncertaintyQuantification2025}. Our contribution instead asks where uncertainty can be \emph{observed} in an LLM\@. Formally, this varies the output space $\mathcal Y$ over the four lenses' observation spaces $\mathcal O_\lambda$ defined in Section~\ref{sec:lenses}, while previous works usually fix $\mathcal Y$ to the vocabulary sequence space $V^*$. Two sequences with different wording but the same meaning are separable in vocabulary space, but inseparable in meaning space. So their difference is epistemic in one lens but aleatoric in another lens.

\paragraph{Uncertainty and related concepts.} \emph{Uncertainty} and \emph{confidence} are frequently used interchangeably in the UQ literature~\citep{liuUncertaintyQuantificationConfidence2025}. In this work, we distinguish them as follows. Uncertainty is treated as the dispersion of the model's predictive state, which has no ground truth and is observed through the lenses. Confidence is the model's belief that its prediction is correct, which requires a ground truth and is decided at the output. It is defined as a prediction target for UQ in Section~\ref{sec:eval-target}. \emph{Calibration} refers to the agreement between the estimated uncertainty and the empirical correctness frequency. Calibration names a family of evaluation metrics in Section~\ref{sec:eval-metrics}. Finally, UQ is widely used as a signal for \emph{hallucination} detection in LLMs, where a hallucination is generated content that is factually incorrect or ungrounded by the context~\citep{kangUncertaintyQuantificationHallucination2026}. We return to the relationship between UQ and hallucination in Section~\ref{sec:eval-target}.

\subsection{Uncertainty Evaluation}\label{sec:background-eval}

Evaluating a UQ method means asking how well its uncertainty score serves a downstream task, such as correctness prediction~\citep{shorinwaSurveyUncertaintyQuantification2025} or contextual hallucination detection~\citep{kangUncertaintyQuantificationHallucination2026}. Since uncertainty has no ground truth, the score is evaluated against a \emph{target} that does. Certainty may be read as the model's confidence in answering correctly, as a reflection of its parametric knowledge, or as its faithfulness to the context. Whether the target is met is decided by a \emph{criterion} that checks the output against a reference, and evaluation \emph{metrics} then quantify how well the uncertainty score tracks that target.

Most work evaluates UQ methods against the correctness target, quantifying how well the score can be read as the model's confidence. However, the criterion for correctness is often left implicit~\citep{janiakIllusionProgressReevaluating2025, santilliRevisitingUncertaintyQuantification2025}. UQ papers label outputs with different correctness functions, such as exact match or meaning equivalence~\citep{williamsBroadCoverageChallengeCorpus2018}, each fixing a different notion of correctness, but the results are reported and compared together. For metrics, we group the families discussed in related surveys into ranking, calibration, and scoring rules, and add four families based on the literature review. Table~\ref{tab:survey-comparison} summarizes the evaluation setups of related UQ surveys, and Section~\ref{sec:evaluation} discusses the evaluation of UQ methods in detail.

\subsection{Uncertainty in Large Language Models}\label{sec:background-llm}

\paragraph{Transformer architecture.} 

The LLM model $f_\theta$ maps an input $x$ to an output $y$ through layered computations. The architecture operates in three main phases: (1) The embedding first maps each input token to a vector and forms the initial state $h^{(0)}$ together with positional information. (2) The $L$ transformer blocks then transform and contextualize these vectors, and finally (3) an unembedding turns the final vector back into a distribution over the vocabulary. The blocks are indexed by $\ell \in \{1,\dots,L\}$. We use $h^{(\ell)} \in \mathbb{R}^d$ to denote the residual-stream state after block $\ell$, where $d$ is the hidden dimension. This vector carries information from one layer to the next, updated by the transformer blocks. In our framework, the representational lens reads $h^{(\ell)}$ itself and the structural lens reads the updates written into it, defined in Section~\ref{sec:lenses-four}. 

Each block contains a multi-head attention (MHA) block and a feed-forward network (FFN)\@. We refer to the individual attention heads of the MHA and the FFN as the \emph{components} of a block, indexed by $c \in \{0,\dots, C\}$, where $c \in \{0,\dots, C-1\}$ are the $C$ attention heads and $c = C$ is the FFN, following their computation order. Every component reads from the residual stream and writes an additive update $\Delta h_c^{(\ell)}$ back into it. 

The component update $\Delta h_c^{(\ell)}$ is defined as follows. When $c$ is an attention head, it is given by Eq.~\ref{eq:mha-update}.
\begin{equation}
    \Delta h_c^{(\ell)} = W_O^{(\ell,c)} V_c^{(\ell)} a_c^{(\ell)},
    \qquad
    a_c^{(\ell)} = \mathrm{softmax}\!\left( \frac{\big(K_c^{(\ell)}\big)^{\!\top} q_c^{(\ell)}}{\sqrt{d_k}} \right),
    \qquad
    \Delta h_{\text{MHA}}^{(\ell)} := \sum_{c=0}^{C-1} \Delta h_c^{(\ell)},
    \label{eq:mha-update}
\end{equation}
where $t$ is the current position, $d_k$ is the head dimension, $q_c^{(\ell)} \in \mathbb{R}^{d_k}$ is the query projection of $h^{(\ell-1)}$, $K_c^{(\ell)}, V_c^{(\ell)} \in \mathbb{R}^{d_k\times t}$ stack the key and value projections at positions $1,\dots,t$, $W_O^{(\ell,c)} \in \mathbb{R}^{d\times d_k}$ is the output projection, and $\Delta h_{\text{MHA}}^{(\ell)}$ is the summed write of the layer's attention heads. The attention weights $a_c^{(\ell)} \in \mathbb{R}^{t}$ of every position form the lower-triangular attention pattern $A_c^{(\ell)} \in \mathbb{R}^{T\times T}$.

When $c$ is the feed-forward network, the write is given by Eq.~\ref{eq:ffn-update}.
\begin{equation}
    \Delta h_{\text{FFN}}^{(\ell)} := \Delta h_C^{(\ell)} = W_2^{(\ell)} m^{(\ell)} + b_2^{(\ell)},
    \qquad
    m^{(\ell)} = \sigma\!\left( W_1^{(\ell)} \big( h^{(\ell-1)} + \Delta h_{\text{MHA}}^{(\ell)} \big) + b_1^{(\ell)} \right),
    \label{eq:ffn-update}
\end{equation}
with $\sigma$ a pointwise nonlinearity. Block $\ell$ adds both writes to $h^{(\ell-1)}$.\footnote{\label{fn:parallel-blocks}We describe the serial block used by most current LLMs. We also consider parallel blocks, as in PaLM~\citep{chowdheryPaLMScalingLanguage2022} where the FFN read $h^{(\ell-1)}$ only, and common attention and FFN variants, including multi-query and grouped-query attention~\citep{shazeerFastTransformerDecoding2019, ainslieGQATrainingGeneralized2023}, multi-head latent attention~\citep{deepseek-aiDeepSeekV2StrongEconomical2024}, sliding-window attention~\citep{jiangMistral7B2023}, and mixture-of-experts layers~\citep{jiangMixtralExperts2024}.} At the update level, the component writes $\Delta h_c^{(\ell)}$ further factorizes into the intermediate state, the attention weights $a_c^{(\ell)}$ for a head in Eq.~\ref{eq:mha-update} and the activation $m^{(\ell)}$ for the FFN in Eq.~\ref{eq:ffn-update}. These factors are read as parts of the same component update.

Modern transformers place normalization before each component (pre-norm) and apply a final normalization $\mathrm{norm}$ before the unembedding~\citep{zhangRootMeanSquare2019, baLayerNormalization2016}. The unembedding matrix $W_U$ maps a normalized hidden state to logits over the vocabulary $V$, and a softmax turns those logits into a token distribution. We write it as Eq.~\ref{eq:unembedding}.
\begin{equation}
    \Pi(h^{(\ell)}) := \mathrm{softmax}\big(W_U\,\mathrm{norm}(\rho h^{(\ell)})\big)
    \label{eq:unembedding}
\end{equation}
where $\rho$ is a fixed map applied to the hidden state, and $\rho=I$ by default. The choices of $\rho$ can be externally defined~\citep{belroseElicitingLatentPredictions2025, blankRlensMakingJlens2026, gurnee2026verbalizable}. The readout factorizes through the logits $z(h^{(\ell)}) := W_U\,\mathrm{norm}(\rho h^{(\ell)})$. The direct output is $\Pi(h^{(L)})$, and the same map applied to an intermediate state gives $\Pi(h^{(\ell)})$. This token distribution is read through the distributional lens of Section~\ref{sec:lenses-four}. We omit normalizations in the following, absorbing each component's input normalization into $\Delta h_c^{(\ell)}$ and writing $\Pi(h^{(\ell)}) = \mathrm{softmax}\big(W_Uh^{(\ell)}\big)$ if $\rho=I$.

\paragraph{Generation and sampling.} 

Given the input $x$ and the tokens $y^{<t}$ generated so far, the readout at the last position is the next-token distribution, $p(y^t \mid x, y^{<t}, \theta)$. LLM generation is autoregressive, meaning that each sampled token is appended to the context and the forward pass is repeated, so the probability of a whole sequence is the product of these conditionals, as in Eq.~\ref{eq:autoregressive-factorisation}.
\begin{equation}
    p(y \mid x, \theta) = \prod_{t=1}^{T} p(y^t \mid x, y^{<t}, \theta),
    \label{eq:autoregressive-factorisation}
\end{equation}
where $y \in V^*$ is a sequence over the vocabulary of length $T$, and $t$ indexes token position within the sequence. We distinguish the token-level quantity $p(y^t \mid x, y^{<t}, \theta)$ from the sequence-level quantity $p(y \mid x, \theta)$ in the distributional lens in Section~\ref{sec:lens-dis}. 

A decoding procedure turns the predicted probabilities into text. The simplest greedy decoding always selects the single most likely token. Methods like temperature sampling~\citep{ficlerControllingLinguisticStyle2017} and top-$p$ sampling~\citep{holtzmanCuriousCaseNeural2020} introduce randomness by picking tokens probabilistically. Under such stochastic decoding, the same input can yield different outputs. This enables output resampling with a fixed model and input, yielding the \emph{output} variation shown in Section~\ref{sec:variation-axis}.

The parameters $\theta$ can also be treated as random variables. Under a Bayesian view, the parameters $\theta$ are random variables drawn from a posterior $p(\theta \mid \mathcal{D})$ given the training data $\mathcal{D}$~\citep{nealBayesianLearningNeural2012}. Resampling $\theta$ yields different predictions for the same input~\citep{galDropoutBayesianApproximation2016, lakshminarayananSimpleScalablePredictive2017}, which realizes the \emph{parameter} variation of Section~\ref{sec:variation-axis}.

Consider the generated outputs: ``10 cents'' and ``\$0.10'' differ in form but give the same answer. Outputs that differ token by token can share a meaning, so a semantic measure is needed. Natural language inference via bidirectional entailment~\citep{farquharDetectingHallucinationsLarge2024} treats two sequences as equivalent when each entails the other. Equivalence can also be judged by an LLM~\citep{manakulSelfCheckGPTZeroResourceBlackBox2023}, returning an equivalence label or a graded score. Embedding similarity~\citep{reimersSentenceBERTSentenceEmbeddings2019} embeds sequences and measures closeness. These instantiate a semantic relation $R: V^* \times V^* \to [0,1]\,\,\text{or}\,\,\mathbb{R}$ on pairs of sequences, providing the foundation for the semantical lens of Section~\ref{sec:lenses-four}. Input resampling within a semantic neighbourhood relies on the same relation and realizes the \emph{input} variation of Section~\ref{sec:variation-axis}.

% !TEX root = tmlr.tex
\section{Related Work}\label{sec:related}

Table~\ref{tab:survey-comparison} organizes existing surveys and position papers by the research question each work poses, including \emph{what} uncertainty is, \emph{which} sources produce it, \emph{when} in the life cycle it arises, \emph{how} to estimate it, and \emph{where} it is observed. Appendix~\ref{sec:appendix-selection} describes how we selected these works and the methods surveyed in Section~\ref{sec:survey}. Across these works, the research question often differs from the organizing axis it uses. Several surveys frame their scope around sources or stages of uncertainty but still group methods by algorithmic family~\citep{gawlikowskiSurveyUncertaintyDeep2023, huUncertaintyNaturalLanguage2023, beigiRethinkingUncertaintyCritical2024}. We review these clusters below and locate the gap that motivates our lens framework. To the best of our knowledge, no prior taxonomy treats where the uncertainty is observed as a dimension of variation.

\begin{table}[ht]
\caption{Comparison of UQ surveys, benchmarks, and position papers.
\textit{Target} in Section~\ref{sec:eval-target}: \colorbox{navybg}{\cCOR} correctness, \colorbox{tealbg}{\cGRD} grounding, \colorbox{yellowbg}{\cUNS} underspecification, \colorbox{skybg}{\cFTH} faithfulness. \textit{Eval. Metrics} in Section~\ref{sec:eval-metrics}: \irnk ranking, \ical calibration, \iscr scoring rules, \idec decision, \ireg regression, \icov coverage, \ifth faithfulness. \textit{Benchmark}: $\circ$ survey of datasets, $\bullet$ empirical evaluation. ``---'' denotes not addressed items. \textit{Abbreviations}: MF = method family, DL = deep learning, LLM = large language model, HCI = human-computer interaction.}
\label{tab:survey-comparison}
\vspace{4pt}
\centering
\adjustbox{max width=\textwidth}{%
\renewcommand{\arraystretch}{1.25}%
\setlength{\tabcolsep}{5pt}%
\small
\begin{tabular}{l l l c c c c}
\toprule
\textsc{Survey} & \textsc{Research Question} & \textsc{Organizing Axis} & \textsc{Scope} & \textsc{Target} & \textsc{Eval.} & \textsc{Bench.} \\
\midrule

% ── What ──────────────────────────────────────────────
\rowcolor{skybg}
\multicolumn{7}{l}{\textcolor{sky}{\textsc{What is uncertainty}}} \\
\makecell[tl]{\citet{hullermeierAleatoricEpistemicUncertainty2021}\\ {\tiny Mach.\ Learn.}}
  & What is A/E uncertainty
  & Formal framework
  & DL
  & ---
  & ---
  & --- \\

\midrule
% ── Which ─────────────────────────────────────────────
\rowcolor{yellowbg}
\multicolumn{7}{l}{\textcolor{yellow}{\textsc{Which uncertainty source}}} \\
\makecell[tl]{\citet{heSurveyUncertaintyQuantification2026}\\ {\tiny ACM Comput.\ Surv.}}
  & Which uncertainty source
  & Data / Model
  & DL
  & \colorbox{navybg}{\cCOR}\colorbox{yellowbg}{\cUNS}
  & \ical\irnk
  & --- \\
\makecell[tl]{\citet{liuUncertaintyQuantificationConfidence2025}\\ {\tiny KDD Tutorial}}
  & Which source in LLM
  & Source $\times$ Computation
  & LLM
  & \colorbox{navybg}{\cCOR}\colorbox{yellowbg}{\cUNS}
  & \ical\irnk
  & \bsrv \\
\makecell[tl]{\citet{gawlikowskiSurveyUncertaintyDeep2023}\\ {\tiny AI Review}}
  & Which source in DL
  & MF-DL
  & DL
  & \colorbox{navybg}{\cCOR}\colorbox{yellowbg}{\cUNS}
  & \ical
  & \bsrv \\

\midrule
% ── When ──────────────────────────────────────────────
\rowcolor{tealbg}
\multicolumn{7}{l}{\textcolor{teal}{\textsc{When in pipeline}}} \\
\makecell[tl]{\citet{beigiRethinkingUncertaintyCritical2024}\\ {\tiny arXiv}}
  & When in LLM life cycle
  & MF-LLM
  & LLM
  & \colorbox{navybg}{\cCOR}
  & ---
  & --- \\
\makecell[tl]{\citet{kirchhofPositionUncertaintyQuantification2025}\\ {\tiny ICML Position}}
  & When in agent interaction
  & HCI Stages
  & \makecell[c]{LLM\\Agents}
  & \colorbox{navybg}{\cCOR}\colorbox{yellowbg}{\cUNS}
  & ---
  & --- \\
\makecell[tl]{\citet{huUncertaintyNaturalLanguage2023}\\ {\tiny arXiv}}
  & When in NLP pipeline
  & MF-Modeling
  & NLP
  & \colorbox{navybg}{\cCOR}\colorbox{yellowbg}{\cUNS}
  & \ical\irnk
  & --- \\

\midrule
% ── How ───────────────────────────────────────────────
\rowcolor{slatebg}
\multicolumn{7}{l}{\textcolor{slate}{\textsc{How to estimate}}} \\
\makecell[tl]{\citet{shorinwaSurveyUncertaintyQuantification2025}\\ {\tiny ACM Comput.\ Surv.}}
  & How to estimate in LLMs
  & MF-LLM
  & LLM
  & \colorbox{navybg}{\cCOR}
  & \ical
  & \bsrv \\
\makecell[tl]{\citet{xiaSurveyUncertaintyEstimation2025}\\ {\tiny ACL Findings}}
  & How to estimate at inference
  & MF-LLM
  & LLM
  & \colorbox{navybg}{\cCOR}
  & \irnk
  & \bevl \\
\makecell[tl]{\citet{vashurinBenchmarkingUncertaintyQuantification2025}\\ {\tiny ACL Tutorial}}
  & How to benchmark UQ
  & MF $\times$ Access
  & LLM
  & \colorbox{navybg}{\cCOR}
  & \ical\irnk
  & \bevl \\
\makecell[tl]{\citet{huangLookYouLeap2025}\\ {\tiny IEEE Trans.\ Softw.\ Eng.}}
  & How to identify risks
  & MF-Inference
  & LLM
  & \colorbox{navybg}{\cCOR}\colorbox{tealbg}{\cGRD}
  & \irnk\ireg
  & \bevl \\
\makecell[tl]{\citet{kangUncertaintyQuantificationHallucination2026}\\ {\tiny IEEE Inform. Theory Mag.}}
  & How to detect hallucinations
  & MF $\times$ Access
  & LLM
  & \colorbox{navybg}{\cCOR}\colorbox{tealbg}{\cGRD}
  & \ical\irnk
  & \bevl \\
\makecell[tl]{\citet{gengSurveyConfidenceEstimation2024}\\ {\tiny NAACL}}
  & How to calibrate confidence
  & Task $\times$ Access
  & LLM
  & \colorbox{navybg}{\cCOR}\colorbox{tealbg}{\cGRD}
  & \ical
  & --- \\
\makecell[tl]{\citet{huangSurveyUncertaintyEstimation2024}\\ {\tiny arXiv}}
  & How does theory meet practice
  & MF-Theory
  & DL \& LLM
  & \colorbox{navybg}{\cCOR}\colorbox{yellowbg}{\cUNS}
  & ---
  & --- \\
\makecell[tl]{\citet{abdarReviewUncertaintyQuantification2021}\\ {\tiny Inf.\ Fusion}}
  & How to catalog DL UQ
  & MF-DL
  & DL
  & \colorbox{navybg}{\cCOR}
  & \ical
  & --- \\
\makecell[tl]{\citet{fakourStructuredReviewLiterature2024}\\ {\tiny arXiv}}
  & How to catalog DL UQ
  & MF-DL
  & DL
  & \colorbox{navybg}{\cCOR}
  & \ical\iscr
  & --- \\
\makecell[tl]{\citet{wangAleatoricEpistemicExploring2025}\\ {\tiny arXiv}}
  & How to estimate by mechanism
  & MF-Mechanism
  & DL
  & \colorbox{navybg}{\cCOR}
  & \irnk\ical\iscr
  & --- \\
\midrule
% ── Where (this work) ────────────────────────────────
\rowcolor{navybg}
\multicolumn{7}{l}{\textcolor{navy}{\textsc{Where in observation}}} \\
\textbf{This work}
  & Where in mechanistic observation
  & Mechanistic observation
  & LLM
  & \makecell[c]{\colorbox{navybg}{\cCOR}\colorbox{tealbg}{\cGRD}\\\colorbox{yellowbg}{\cUNS}\colorbox{skybg}{\cFTH}}
  & \makecell[c]{\irnk\ical\iscr\\\idec\ireg\icov\ifth}
  & \bsrv \\
\bottomrule
\end{tabular}
}
\end{table}

\paragraph{What is uncertainty.}
One line of work asks what uncertainty is and answers with a formal framework. \citet{hullermeierAleatoricEpistemicUncertainty2021} define the A/E decomposition relative to a context setting $(\mathcal X, \mathcal Y, \mathcal H, \mathcal P)$, discussed in Section~\ref{sec:background-uncertainty}. Later surveys adopt this vocabulary but fix the output space $\mathcal Y$ to the task output, a predicted label or a generated sequence in $V^*$, so uncertainty is only observed at a single point. We instead treat $\mathcal Y$ as a variable representing different observation spaces of the model computation.

\paragraph{Which source.}
With uncertainty defined, the next question is which part of the modeling process generates it. This line of surveys organizes the literature by the \emph{source} of uncertainty.
\citet{heSurveyUncertaintyQuantification2026} partition uncertainty sources into data and model and equates data uncertainty with aleatoric and model uncertainty with epistemic, a different definition of A/E~\citep{kirchhofPositionUncertaintyQuantification2025}. They organize UQ methods by the source.
\citet{liuUncertaintyQuantificationConfidence2025} define uncertainty sources as: input uncertainty in the prompt, reasoning uncertainty in the reasoning process, parameter uncertainty from approximating the weights, and prediction uncertainty from stochastic decoding. They cross-cut sources with a computation efficiency axis.

Earlier, \citet{gawlikowskiSurveyUncertaintyDeep2023} used the source framing for deep neural networks by decomposing uncertainty into data, model, and distribution, but grouped methods by method family.
These surveys examines the source of uncertainty in deep learning and LLM scenarios without considering the observation point.

\paragraph{When in pipeline.}
Another line of work criticizes the A/E decomposition in LLMs and proposes to decompose uncertainty along the modeling pipeline, aiming to answer \emph{when} in the model's life cycle or interaction loop uncertainty arises. \citet{beigiRethinkingUncertaintyCritical2024} argue that the classical A/E dichotomy is insufficient for LLMs and reorganizes uncertainty along LLM life cycle stages, including the pretraining and training stage, instruction tuning and alignment stage, inference stage, and output stage. \citet{kirchhofPositionUncertaintyQuantification2025} focus on human-computer interaction (HCI), proposing that standard uncertainty definitions ``directly contradict each other'' in an interactive LLM agent setting. They analyze uncertainty along three phases of HCI, including underspecification uncertainty from ambiguity in the user's intent, interactive learning through agent follow-up, and output uncertainty in the agent's response.

The earlier work in Natural Language Processing (NLP)~\citep{huUncertaintyNaturalLanguage2023} traces uncertainty through the NLP pipeline, separating A/E uncertainty in input, system, and output stages.
These taxonomies focus on the occurrence of uncertainty in the LLM, agent, or NLP pipeline, but fix the observation to the task output.

\paragraph{How to estimate.}
The largest group of surveys asks \emph{how} to estimate uncertainty and organizes methods by algorithmic family. Among these, \citet{shorinwaSurveyUncertaintyQuantification2025} stand closest to our work, as the only LLM-focused UQ survey that elevates mechanistic interpretability (MI) to a top-level category. Their taxonomy classifies methods into token-level methods operating on output logits, self-verbalized confidence giving verbal judgments, semantic-similarity methods comparing meaning across outputs, and MI methods reading internal model states. They explicitly note that the MI category is unexplored. Our four observation lenses are grounded in MI and map onto their taxonomy. Token-level and semantic-similarity methods correspond to the distributional and semantical lenses, and self-verbalized confidence is a subtype of semantics. The MI methods split into the structural and representational lenses. Unlike disjoint method families, the lenses are connected by the residual stream and ordered by the information they read.

The remaining method-family surveys vary in scope. \citet{xiaSurveyUncertaintyEstimation2025} focus on inference-time uncertainty estimation and evaluate twelve methods. \citet{vashurinBenchmarkingUncertaintyQuantification2025} implement the LM-Polygraph framework for benchmarking. \citet{huangLookYouLeap2025} conduct an exploratory analysis of UQ methods for risk identification tasks. \citet{kangUncertaintyQuantificationHallucination2026} classify hallucination detection methods. \citet{gengSurveyConfidenceEstimation2024} cross-cut methods by task type and model-access level. \citet{huangSurveyUncertaintyEstimation2024} examine the gap between UQ theory and practice. Several earlier surveys catalog DL uncertainty methods without specializing to language models~\citep{abdarReviewUncertaintyQuantification2021, fakourStructuredReviewLiterature2024, wangAleatoricEpistemicExploring2025}. These works classify methods by the procedure.

\paragraph{Where in observation.}
Existing taxonomies vary along three axes, including the source of uncertainty, the occurrence of uncertainty, and the algorithmic family. One survey classifies UQ methods reading LLM internal states as a method family, but never distinguishes which internal observation is read. No prior survey treats the observation itself as a taxonomic axis. The remainder of this paper introduces the mechanistic observation as a fourth axis, that is, where in the model's computation the uncertainty is observed.

% !TEX root = tmlr.tex
\section{Observation Lenses in LLMs}\label{sec:lenses}

The taxonomies reviewed in Section~\ref{sec:related} differ in the research questions and the organizing axis, but none of them explicitly considers the observation point of uncertainty in LLM computation~\citep{hullermeierAleatoricEpistemicUncertainty2021, heSurveyUncertaintyQuantification2026, shorinwaSurveyUncertaintyQuantification2025}. The observation point determines how much internal model information can be accessed, so it should be treated as a design choice rather than left implicit.

We organize UQ methods along a \emph{lens} axis, specifying where they read in the LLM~\citep{elhageMathematicalTransformerCircuits2021, somvanshiBridgingBlackBox2026a}. For each input, the model $f_\theta$ maps the residual stream to an output. Each lens provides a different view of this computation. We identify four observation lenses based on mechanistic interpretability, called the structural, representational, distributional, and semantical lenses.

Section~\ref{sec:lenses-residual} introduces the primacy of the residual stream, and Section~\ref{sec:lenses-definition} defines the four lenses formally. Section~\ref{sec:lenses-estimator} gives the uncertainty estimator at each lens. Section~\ref{sec:lenses-hierarchy} establishes the information loss relation between the four lenses, and Section~\ref{sec:lenses-poc} validates that relation empirically.

\subsection{Residual-stream Primacy}\label{sec:lenses-residual}

We first establish the architecture that grounds every observation. Drawing from mechanistic interpretability, all components of a transformer, including token embedding, attention heads, feed-forward neural network, and unembedding blocks, communicate with each other by reading and writing to the residual stream. The residual stream is the shared communication channel between components~\citep{elhageMathematicalTransformerCircuits2021}.

\begin{proposition}[Residual stream primacy]\label{claim:residual-stream-primacy}
Let $f_\theta$ be a standard pre-norm transformer language model with $L$ blocks indexed by $\ell\in\{1,\dots,L\}$, with normalization absorbed. Let $h^{(0)}$ denote the initial state. Each block has $C+1$ components, $C$ attention heads followed by one FFN, each reading the current residual stream and writing additively back to it. At a given token position, the residual-stream state after block $\ell$ is
\begin{equation}
    h^{(\ell)} = h^{(\ell-1)} + \Delta h_{\text{MHA}}^{(\ell)}\big( h^{(\ell-1)} \big) + \Delta h_{\text{FFN}}^{(\ell)}\big( h^{(\ell-1)} + \Delta h_{\text{MHA}}^{(\ell)} \big),
    \label{eq:residual-update}
\end{equation}
where $\Delta h_{\text{MHA}}^{(\ell)}(\cdot)$ are the write operations by the attention blocks and $\Delta h_{\text{FFN}}^{(\ell)}(\cdot)$ that of the FFN, stated in Eqs.~\ref{eq:mha-update} and~\ref{eq:ffn-update}. The FFN processes each position independently, and the attention heads read the states $h^{(\ell-1)}$ at the current and all preceding positions. Accordingly, the residual stream is the \emph{unique} channel through which components communicate across layers and positions. The residual stream state $h^{(\ell)} = h^{(0)} + \sum_{\ell'=1}^{\ell} \sum_{c=0}^{C} \Delta h_c^{(\ell')}$ consists of the initial state and all preceding component updates, and every component reads from it.\footnote{\label{fn:parallel-primacy}The proposition applies directly to most current LLMs, including variants in Footnote~\ref{fn:parallel-blocks}, since every component reads a residual-stream state and writes additively to it. Architectures with cross-layer connections, like cross-layer key-value sharing~\citep{brandonReducingTransformerKeyValue2024, sunYouOnlyCache2024} and multi-stream or depth-weighted residual connections~\citep{pagliardiniDenseFormerEnhancingInformation2024, zhuHyperConnections2025} are left open for future work.}
\end{proposition}

\begin{takeaway}{Communication channel in transformers}
    All communication in a transformer is mediated by the residual stream, since every component at every layer and position reads from and writes to it. 
\end{takeaway}

\subsection{Mechanistic Observation Lenses}\label{sec:lenses-definition}

\begin{figure}[h!]
    \centering
    \includegraphics[width=0.85\linewidth]{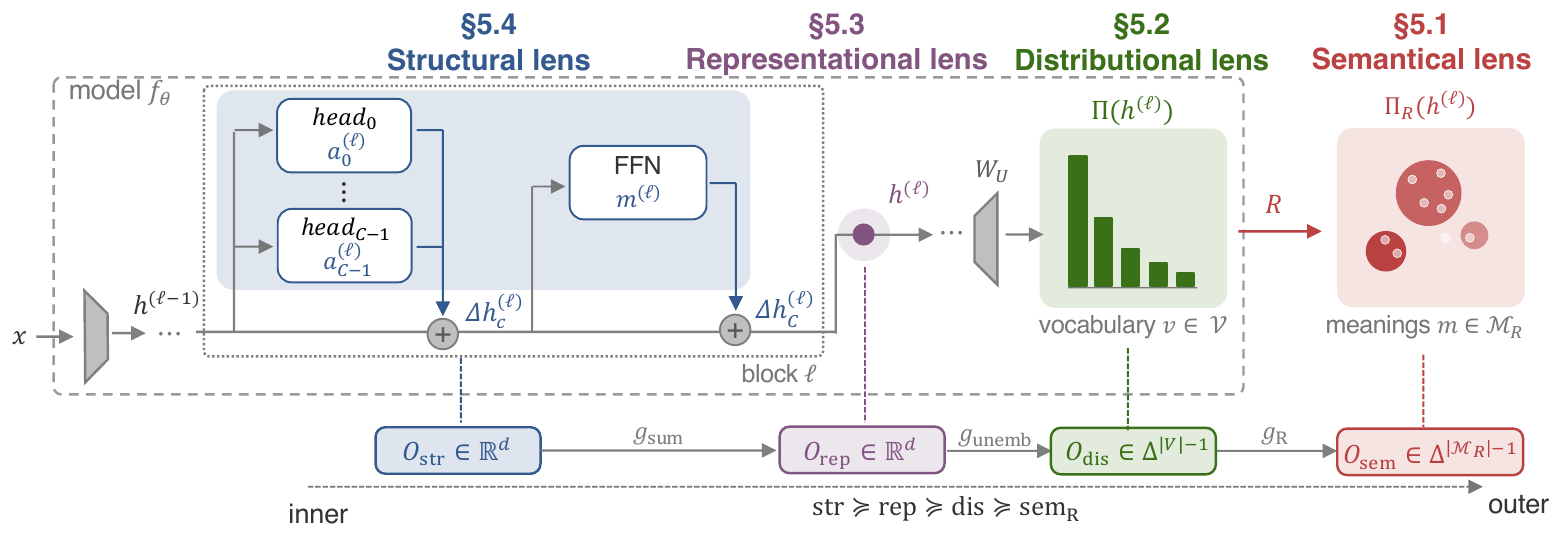}
    \caption{Four mechanistic observation lenses in the residual stream.}
    \label{fig:four-lenses}
\end{figure}

We define a \emph{mechanistic observation lens} by the type of object it reads from the model's computation. A lens specifies what is observed and an uncertainty estimator turns these observations into a score. Figure~\ref{fig:four-lenses} locates the four lenses along the residual stream. The structural lens observes the individual attention-head and FFN writes to the residual stream, including their attention weights and activations. The representational lens observes the hidden state. The distributional lens observes the token probabilities. The semantical lens observes a distribution over meanings, obtained with a semantic relation $R$.

\subsubsection{Residual Stream Observations}\label{sec:lenses-observations}

By Proposition~\ref{claim:residual-stream-primacy}, every observation of the model's computation is obtained from the residual stream state $h^{(\ell)}$ or from its component updates $\Delta h_c^{(\ell)}$ and their factorization. A \emph{mechanistic object} is what the residual stream exposes at one point of the computation.

\begin{definition}[Residual Stream Lenses]
    The mechanistic lenses should satisfy the following properties:
    \begin{itemize}
        \item \textbf{Residual naturality}. Each lens reads the model $f_\theta$ residual stream alone up to a fixed external map. Its readout takes either the component updates $\Delta h_c^{(\ell)}$, possibly factorized into $a_c^{(\ell)}$ and $m^{(\ell)}$, or the stream state $h^{(\ell)}$ as input and returns an observation $O_\lambda \in \mathcal O_\lambda$.
        \item \textbf{Object identification}. Each lens is identified by the mechanistic object it observes, or equivalently by the type of its canonical observation space $\mathcal O_\lambda$. Two readouts that observe the same object belong to the same lens.
        \item \textbf{Observation determinism}. The readout of each lens is a fixed map from the residual stream or the components to the observation space $\mathcal O_\lambda$. It should not fit parameters to the input or perform sampling over the LLM\@.
    \end{itemize}\label{claim:residual-lenses}
\end{definition}

Current mechanistic interpretability research~\citep{bereskaMechanisticInterpretabilityAI2024, barezUnderstandingAIState2026} studies four kinds of objects in the residual stream: the individual components~\citep{elhageMathematicalTransformerCircuits2021, somvanshiBridgingBlackBox2026a}, the features encoded in the hidden state~\citep{elhageToyModelsSuperposition2022, parkLinearRepresentationHypothesis2024, olahZoomIntroductionCircuits2020}, the predictions over the vocabulary~\citep{nostalgebraistInterpretingGPTLogit2020, gevaTransformerFeedForwardLayers2022}, and the meanings of the generated outputs~\citep{kuhnSemanticUncertaintyLinguistic2023, benderClimbingNLUMeaning2020}.

Each mechanistic object has one canonical observation space $\mathcal O_\lambda$, defined formally in Section~\ref{sec:lenses-four}. The \emph{structural lens} is identified by observing the \underline{individual components} $\Delta h_c^{(\ell)}$. The \emph{representational lens} is identified by observing the \underline{hidden state} $h^{(\ell)}$. The \emph{distributional lens} is identified by observing the \underline{token distribution} $\Pi(h^{(\ell)})$, obtained by unembedding the state with $W_U$. The \emph{semantical lens} is identified by observing the \underline{meaning distribution} $\Pi_R(h^{(\ell)})$, obtained by a semantic relation $R$.

Residual naturality restricts all lenses to the model $f_\theta$. The unembedding $W_U$ and the normalization are part of $\theta$, so the structural, representational, and distributional lenses require no specification beyond the model itself. The semantical observation is defined on meanings through a semantic relation $R$ between outputs. The $R$ is the only structure external to $\theta$ used to define a lens, discussed in Section~\ref{sec:lens-sem}.

The forward pass computation determines the order of the lenses. Component writes are summed into the residual stream state, the state is unembedded into a prediction over $V$, and the prediction is coarsened by $R$ into a meaning distribution. We call a lens \emph{inner} to another if its object comes earlier in the computation. Section~\ref{sec:lenses-hierarchy} shows that every inner observation determines every \emph{outer} one.

A new interpretability tool can only change \emph{how} a lens is read. Each tool applies a readout map $\rho$ to the observation before uncertainty estimation. For the distributional lens, the map acts on $h^{(\ell)}$ before unembedding, giving a prediction $\mathrm{softmax}(W_U\,\mathrm{norm}(\rho h^{(\ell)}))$ over $V$. The Logit Lens~\citep{nostalgebraistInterpretingGPTLogit2020} reads $\Pi(h^{(\ell)})$ with $\rho = I$. The J-Lens~\citep{gurnee2026verbalizable} reads $h^{(\ell)}$ with a Jacobian-based map $J_\ell$, the R-Lens~\citep{blankRlensMakingJlens2026} uses a more faithful map at early layers, and the tuned lens~\citep{belroseElicitingLatentPredictions2025} learns an affine map. All of these return a token prediction and read the distributional lens. The R-Lens shows that a different $\rho$ map can make the prediction more legible. For the representational lens, the map $\rho$ applies to the hidden state $h^{(\ell)}$ directly. The sparse autoencoder~\citep{hubenSparseAutoencodersFind2024} rewrites $h^{(\ell)}$ in a learned feature basis, the J-space~\citep{gurnee2026verbalizable} decomposes $h^{(\ell)}$ against the overcomplete dictionary $W_U J_\ell$, the U-space~\citep{braunUSpaceUncoveringWhen2026} decomposes it against certainty anchor basis vectors $U_\ell$. They all read the representational lens.

\subsubsection{Four Observation Lenses}\label{sec:lenses-four}

This subsection defines the four lenses formally. For each lens, we define the observation $O_\lambda$, the type of its observation space $\mathcal O_\lambda$, and its mapping from the adjacent inner lens. 

\paragraph{Structural Lens.} The structural object is the individual component of the transformer~\citep{elhageMathematicalTransformerCircuits2021}. The structural lens reads the summands of the residual-stream update: the head writes $\Delta h_c^{(\ell)},\,\,c<C$ in Eq.~\ref{eq:mha-update}, and the FFN writes $\Delta h_C^{(\ell)}$ in Eq.~\ref{eq:ffn-update}. Following \citet{elhageMathematicalTransformerCircuits2021}, we treat the embedding as a component that writes $\Delta h_0^{(0)} := h^{(0)}$ into the residual stream at layer $0$. The structural observation is the per-component write in Eq.~\ref{eq:observation-str}. Each observation is indexed by its component $(\ell, c)$.
\begin{equation}
    O_{\text{str}} := \Delta h_c^{(\ell)} \in \mathbb R^d, \qquad c\in\{0,\dots,C\},\, \ell\in\{1,\dots,L\},\,(\ell, c)=(0, 0)
    \label{eq:observation-str}
\end{equation}

Each component write further factorizes into an intermediate state of the same component. A head write is $\Delta h_c^{(\ell)}$ with the attention weights $a_c^{(\ell)}$ of Eq.~\ref{eq:mha-update}, and the FFN write is $\Delta h_C^{(\ell)}$ with the activation $m^{(\ell)}$ of Eq.~\ref{eq:ffn-update}. We treat $a_c^{(\ell)}$ and $m^{(\ell)}$ as parts of the component update, and a method that reads either of them reads the structural lens. The structural lens gives the finest-grained view of the computation. It is also the hardest lens to analyze, because observations scale with $L(C+1)$, while attention patterns scale with context length.

\paragraph{Representational Lens.} The representational object is the set of features encoded in the hidden state. Under the superposition and linear representation views, features are directions in $\mathbb R^d$, and the hidden state is approximately a sparse linear combination of them~\citep{elhageToyModelsSuperposition2022, parkLinearRepresentationHypothesis2024}. The representational lens reads the residual-stream state, the sum of the component writes in Eq.~\ref{eq:residual-update}, and its observation is a point in $\mathbb R^d$ as in Eq.~\ref{eq:observation-rep}.
\begin{equation}
    O_{\text{rep}} := h^{(\ell)} \in \mathbb R^d, \qquad \ell\in\{0,\dots,L\}
    \label{eq:observation-rep}
\end{equation}

The map $g_\text{sum}$ from the structural lens to the representational lens is the residual sum of Eq.~\ref{eq:residual-update}. It keeps the total update and discards its split among component writes, so the representational view is coarser than the structural one. Uncertainty estimators at this lens read geometric or topological properties of $h^{(\ell)}$, such as distances to reference states or a learnt probe. A fixed readout map $\rho$, such as an SAE encoder, rewrites $h^{(\ell)}$ in another basis, but still reads the representational lens.

\paragraph{Distributional Lens.} The distributional object is the model's prediction over the vocabulary~\citep{malininUncertaintyEstimationAutoregressive2021, gevaTransformerFeedForwardLayers2022}. The distributional lens unembeds the state by $W_U$ and reads the next-token probability distribution, as in Eq.~\ref{eq:unembedding}. The distributional observation is defined as Eq.~\ref{eq:observation-dis}. This distribution lies on the simplex $\Delta^{|V|-1}$, where the entries correspond to vocabulary tokens, each entry is between 0 and 1, and all entries sum to 1. The sequence-level distribution is the accumulated product in Eq.~\ref{eq:autoregressive-factorisation}.
\begin{equation}
    O_{\text{dis}}:=\Pi(h^{(\ell)}) \in \Delta^{|V|-1}, \qquad \ell\in\{0,\dots,L\}
    \label{eq:observation-dis}
\end{equation}

The map $g_\text{unemb}$ from the representational lens to the distributional lens is the readout in Eq.~\ref{eq:unembedding}, composed of the final normalization, the unembedding $W_U$, and the softmax. The normalization discards the norm $\lVert h^{(\ell)}\rVert$, and also the mean of the coordinates in LayerNorm~\citep{baLayerNormalization2016}. The unembedding $W_U$ is often assumed to be full-rank and preserves the normalized state in general~\citep{xiaoReadoutScalingShapes2026,gurnee2026verbalizable, fengRankRandomMatrix2007, nazirBetterLanguageModel2025}. The softmax then removes a common additive shift of the logits $z(h^{(\ell)})$. The logits and $\Pi(h^{(\ell)})$ describe the same prediction, and a method that reads $z(h^{(\ell)})$ still reads the distributional lens~\citep{maEstimatingLLMUncertainty2025}. A fixed readout map $\rho$~\citep{gurnee2026verbalizable, belroseElicitingLatentPredictions2025} changes how $h^{(\ell)}$ is unembedded but still reads distributional observations, discussed in Section~\ref{sec:lenses-observations}. Functions on the token distribution~\citep{hendrycksBaselineDetectingMisclassified2017, malininUncertaintyEstimationAutoregressive2021}, read as the model's predictive confidence, are the standard baselines in the UQ literature.

\paragraph{Semantical Lens.} The semantical object is the meaning of the generated outputs. Following \citet{benderClimbingNLUMeaning2020}, we take meaning as a relation between an output and the intent it conveys. Here the intent is the response to the query for the task, such as the answer in question answering or the chosen action in decision making. We denote the space of these intents as $\mathcal M$. The meaning relation cannot be observed, and semantical UQ approximates it by a pairwise semantic relation $R$ that judges whether two outputs have the same meaning~\citep{kuhnSemanticUncertaintyLinguistic2023}. The relation $R$ defines the set of possible meanings $\mathcal M_R$. The semantical lens reads the sequence-level prediction distributed over $\mathcal M_R$, as in Eq.~\ref{eq:observation-sem}.
\begin{equation}
    O_{\text{sem}}:=\Pi_R(h^{(\ell)}) \in \Delta^{|\mathcal{M}_R|-1}, \qquad \ell\in\{0,\dots,L\}
    \label{eq:observation-sem}
\end{equation}

The distribution of meanings $\Pi_R(h^{(\ell)})$ is formed from output sequences, which come from the residual stream. The semantical space is the simplex $\Delta^{|\mathcal M_R|-1}$. The set $\mathcal M_R$ is finite in practice because it is built from finite outputs. The map $g_R$ from the distributional lens to the semantical lens moves the sequence-level prediction onto the semantical space. It discards the lexical form of the outputs and keeps only what $R$ reads as their meaning, depending on the implementation of $R$. The observation $O_{\text{sem}}$ is this exact distribution over $\mathcal M_R$, approximated from a finite set of sampled outputs. The relation $R$ can be realized by NLI models~\citep{kuhnSemanticUncertaintyLinguistic2023}, LLM judges~\citep{manakulSelfCheckGPTZeroResourceBlackBox2023}, or embedding models~\citep{aoCSSContrastiveSemantic2024}. The verbalized uncertainty uses an $R$ internal to the LLM\@. The four types of $\mathcal M_R$, cluster, graph, manifold, and report, and their instantiations are explained in Section~\ref{sec:lens-sem} and summarized in Table~\ref{tab:sem-types}. Semantical observation is closer to human interpretation, but is only as faithful as the implementation of $R$.

\begin{takeaway}{Definition of mechanistic observation lenses}
    The mechanistic observation lenses are identified by the mechanistic object they read from the residual stream. From innermost to outermost, the structural lens reads the component updates; the representational lens reads the hidden state; the distributional lens reads the token distribution; and the semantical lens reads the distribution over meanings induced by a semantic relation $R$ between outputs.
\end{takeaway}

\subsection{Uncertainty Estimators at Lenses}\label{sec:lenses-estimator}

\subsubsection{The Estimator as a Point \texorpdfstring{$(\lambda, Z, S)$}{(lambda, Z, S)}}\label{sec:estimator-as-point}

Section~\ref{sec:lenses-definition} defines the four lenses, and each lens $\lambda$ returns an observation $O_\lambda$ of the model's computation. An uncertainty \emph{estimator} turns such observations into a scalar score, predicting how uncertain the LLM is about the response. The uncertainty estimator is written as Eq.~\ref{eq:uncertainty-estimator}.
\begin{equation}
    U_\lambda(x) = \mathcal U\big({\{\,O_\lambda(\theta,x)\,\}}_{Z}, S\big)
    \label{eq:uncertainty-estimator}
\end{equation}
The function $\mathcal{U}$ measures the spread of the observations and returns a scalar uncertainty score. Equation~\ref{eq:uncertainty-estimator} represents every estimator by its coordinates $(\lambda, Z, S)$ on three distinct axes. The lens $\lambda$ in Section~\ref{sec:lens-axis} specifies which observation object the estimator reads from the model's computation. The variation $Z$ in Section~\ref{sec:variation-axis} specifies what varies across multiple forward passes, i.e., the resampling strategy for the estimation. The auxiliary structure $S$ in Section~\ref{sec:auxiliary-axis} specifies how the observations are turned into a scalar, covering the external structure the estimator uses. 

\subsubsection{The Lens Axis}\label{sec:lens-axis}

The lens $\lambda$ specifies where in the model computation the estimator $U_\lambda$ observes uncertainty. We instantiate the axis at the four lenses by substituting each observation $O_\lambda$ into Eq.~\ref{eq:uncertainty-estimator} in Table~\ref{tab:lens-estimator-summary}. 

Each observation $O_\lambda$ is an \emph{atom}, read at a single point of the forward pass: a layer $\ell$, a token position $t$, and a component $c$. A \emph{field} is a set of atoms of the same lens, read from one forward pass over a fixed sequence $(x,y)$ at different $(c,\ell,t)$, for example the hidden states of all layers, or the token distributions along a generated sequence. The function $\mathcal U$ measures the spread within a single atom or across the field. Every lens admits an atom estimator and a field estimator.

\begin{table}[ht]
\centering
\caption{Lens axis summary. Estimator reads $O_\lambda$ and predicts with an atom estimator or a field estimator.}\label{tab:lens-estimator-summary}
\renewcommand{\arraystretch}{1.4}
\vspace{4pt}
\adjustbox{max width=\textwidth}{%
\begin{tabular}{l c l c c}
\toprule
\textsc{Lens} $\lambda$ & \textsc{Observation} $O_\lambda$ & \textsc{Observation Space} $\mathcal O_\lambda$ & \textsc{Atom Estimator} & \textsc{Field Estimator} \\
\midrule
Structural       & $\Delta h_c^{(\ell)}$        & Component write, $\mathbb R^d$    & $\mathcal U\!\left(\Delta h_c^{(\ell)}(x)\right)$        & $\mathcal U\!\left({\{\Delta h_c^{(\ell)}(x^t)\}}_{c,\ell,t}\right)$ \\
Representational & $h^{(\ell)}$                 & Hidden state, $\mathbb R^d$        & $\mathcal U\!\left(h^{(\ell)}(x)\right)$                 & $\mathcal U\!\left({\{h^{(\ell)}(x^t)\}}_{\ell,t}\right)$ \\
Distributional   & $\Pi(h^{(\ell)})$            & Token distribution, $\Delta^{|V|-1}$  & $\mathcal U\!\left(\Pi(h^{(\ell)}(x))\right)$             & $\mathcal U\!\left({\{\Pi(h^{(\ell)}(x^t))\}}_{\ell,t}\right)$ \\
Semantical       & $\Pi_R(h^{(\ell)})$          & Meaning distribution, $\Delta^{|\mathcal M_R|-1}$ & $\mathcal U\!\left(\Pi_R(h^{(\ell)}(x))\right)$           & $\mathcal U\!\left({\{\Pi_R(h^{(\ell)}(x^t))\}}_{\ell,t}\right)$ \\
\bottomrule
\end{tabular}
}
\end{table}

\paragraph{Assigning a method to a lens.} A method's assignment is determined by the mechanistic object it observes. Methods reading the same object read the same lens, whatever their implementation. Differences in the measurement function $\mathcal U$, the variation $Z$, and the auxiliary structure $S$ do not change the lens assignment. Under Definition~\ref{claim:residual-lenses}, every uncertainty estimator in this survey has a well-defined lens, since it reads one of the four types of objects. A method that reads two or more of them is a \emph{cross-lens} estimator, which Section~\ref{sec:survey} presents under the innermost lens it reads.

\paragraph{Native and surrogate uncertainty estimators.} The lens axis classifies $U_\lambda$ by where it reads, making our taxonomy estimator-centric. The estimand is defined by what the estimator aims to predict. The estimand and estimator locations may not coincide, so we define two types of uncertainty estimators.

When the uncertainty estimator reads from $O_\lambda$ and estimates the uncertainty of $\lambda$, it is a \emph{native} uncertainty estimator, defined in Eq.~\ref{eq:uncertainty-estimator-native}.
\begin{equation}
    U_\lambda^{\text{nat}}(x) = \mathcal U\big({\{\,O_\lambda(\theta,x)\,\}}_{Z}, S\big)
    \label{eq:uncertainty-estimator-native}
\end{equation}

When the uncertainty estimator reads from $O_\lambda$ and estimates the uncertainty of a different lens $\lambda'$, it is a \emph{surrogate} uncertainty estimator, defined in Eq.~\ref{eq:uncertainty-estimator-surrogate}.
\begin{equation}
    U_\lambda^{\text{sur}}(x) = \mathcal U\big({\{\,O_\lambda(\theta,x)\,\}}_{Z}, S\big),\qquad \mathrm{s.t.} \quad U_\lambda^{\text{sur}}(x) \approx \mathcal U'\big({\{\,O_\lambda'(\theta,x)\,\}}_{Z'}, S'\big)
    \label{eq:uncertainty-estimator-surrogate}
\end{equation}

The native--surrogate axis provides an estimand-level perspective. An inner lens may serve as a surrogate for an outer lens, although the converse need not hold. Section~\ref{sec:lenses-hierarchy} explains the reasons for this asymmetry.

\subsubsection{The Variation Axis}\label{sec:variation-axis}

The variation axis $Z$ specifies what the estimator varies to obtain multiple observations. Each forward pass yields one atom or field of $O_\lambda$, and resampling $Z$ yields $O_\lambda$ across passes. We summarize four values of $Z$. 

\begin{itemize}
    \item $Z = \varnothing$ denotes no resampling. The estimator reads a single forward pass, and $\mathcal U$ measures the spread within one atom or across a field, e.g., the entropy of a token distribution.
    \item $Z = y \sim p(\cdot \mid x, \theta)$ denotes that the output is resampled with $\theta$ and $x$ fixed, as in stochastic decoding in Section~\ref{sec:background-llm}. $\mathcal U$ measures the spread across the sampled generations.
    \item $Z = \theta \sim p(\theta \mid \mathcal D)$ denotes that the parameters are resampled from a posterior given the training data $\mathcal D$, and $\mathcal U$ measures the spread across models. It is the Bayesian predictive uncertainty in Section~\ref{sec:background-llm}.
    \item $Z = x' \sim p(\cdot \mid x)$ denotes that the input is resampled, usually within a semantic neighbourhood of $x$ by paraphrasing or reformulation, and $\mathcal U$ measures the spread across inputs.
\end{itemize}

\subsubsection{The Auxiliary Axis}\label{sec:auxiliary-axis}

The observations alone may not give the most useful uncertainty score. An estimator may sample multiple observations, transform them before $\mathcal U$, fit a $\mathcal U$ function, or calibrate the score for better performance. The auxiliary axis $S$ covers the structure needed for these operations beyond the observation $O_\lambda$ and the variation $Z$. We show four roles of $S$ below. When $S=\varnothing$, the estimator uses the observations alone.

\begin{itemize}
    \item \textbf{$S$ defines the distribution of $Z$.} The estimator draws $Z$ from a distribution $Z_S$. For $Z=\theta$, $Z_S$ can be a fitted approximate posterior $q(\theta)$ over the weights~\citep{blundellWeightUncertaintyNeural2015} or a finite ensemble of trained models~\citep{lakshminarayananSimpleScalablePredictive2017}. For $Z=x'$, $Z_S$ defines a family of input transformations, such as paraphrases, dummy tokens, or prompt variants~\citep{gaoSPUQPerturbationBasedUncertainty2024}.
    
    \item \textbf{$S$ defines the readout map $\rho$.} A readout map $\rho_S$ transforms each observation before $\mathcal U$. The map $\rho_S$ is built offline and does not depend on the query. It can be an affine map~\citep{belroseElicitingLatentPredictions2025}, a Jacobian-based map~\citep{gurnee2026verbalizable}, or a sparse autoencoder~\citep{hubenSparseAutoencodersFind2024}.

    \item \textbf{$S$ parameterizes $\mathcal U$.} The function $\mathcal U$ is implemented as an external model $\mathcal U_S$. It can be a trained linear probe on hidden states~\citep{azariaInternalStateLLM2023} or a Dirichlet head over the token distribution~\citep{malininUncertaintyEstimationAutoregressive2021}. $\mathcal U_S$ is trained against a target defined at a particular lens.

    \item \textbf{$S$ post-processes the score with $\psi$.} A lens-agnostic structure $\psi_S$ post-processes the output of $\mathcal U$. Conformal prediction uses an exchangeable calibration set to give coverage guarantees~\citep{quachConformalLanguageModeling2024}.
\end{itemize}

The four roles compose as in Eq.~\ref{eq:uncertainty-estimator-composition}. The default structure is under no resampling $Z_S=\varnothing$, $\rho_S = I$, $\mathcal U_S$ as a fixed function, and $\psi_S$ is the identity. 
\begin{equation}
    U_\lambda(x) = \psi_S\Big(\mathcal U_S\big({\{\,\rho_S(O_\lambda(\theta,x))\,\}}_{Z\sim Z_S}\big)\Big)
    \label{eq:uncertainty-estimator-composition}
\end{equation}

\begin{takeaway}{Characterizing an uncertainty estimator}
    An uncertainty estimator is characterized by the lens $\lambda$ it reads, the variation $Z$ it resamples, and the auxiliary structure $S$ it uses. The lens is set by the mechanistic object the estimator reads. The variation changes which observations are collected and the auxiliary structure changes how they are scored, but neither changes the lens. An estimator is \emph{native} if it estimates the uncertainty of the lens it reads, and \emph{surrogate} if it estimates the uncertainty of another lens.
\end{takeaway}

\subsection{The Lens Hierarchy}\label{sec:lenses-hierarchy}

We show that the four lens observations form a hierarchy. Each outer observation is a function of every inner one (Lemma~\ref{claim:lens-determination}). Each such function discards some information outward along the process (Corollary~\ref{claim:information-loss}). For a given uncertainty target, we show whether a lens is sufficient to recover it (Corollary~\ref{claim:lens-sufficiency}).

Section~\ref{sec:lenses-four} introduces the lens map from each lens to the adjacent outer one. Lemma~\ref{claim:lens-determination} links these maps into a chain. Fix the input $x$ and a token position $t$. The map $g_\text{sum}$ takes all component writes at position $t$ up to layer $\ell$, including embedding write at layer $0$, and returns their residual sum, $g_\text{sum}\big(\{\Delta h_c^{(\ell')}\}_{c,\,0\le\ell'\le\ell}\big) = \sum_{\ell'=0}^{\ell}\sum_{c}\Delta h_c^{(\ell')}$. The map $g_\text{unemb}$ is the readout in Eq.~\ref{eq:unembedding}. The map $g_R$ takes a set of outputs $\{y_k\}_{k=1}^{K}$ sampled from the model, and maps the sequence-level probability $p(y_k\mid x,\theta)$ of each sample to the semantics distribution. The $g_\text{sum}$ and $g_\text{unemb}$ are determined by $\theta$, and $g_R$ is fixed by $\theta$ and the semantic relation $R$.

\begin{lemma}[Lens Determination]
    For a fixed input $x$, the four lens observations are related by the lens maps, each taken over the collection defined above for its map:
    \begin{equation}
        O_\text{str} \xrightarrow{\;g_\text{sum}\;} O_\text{rep}
        \xrightarrow{\;g_\text{unemb}\;} O_\text{dis} \xrightarrow{\;g_R\;} O_\text{sem},
    \end{equation}
    where each lens map is a well-defined function of its argument.\label{claim:lens-determination}
\end{lemma}

We write $\lambda\succeq\lambda'$, read as $\lambda$ \emph{determines} $\lambda'$, when $O_{\lambda'}=g(O_\lambda)$. The observation $O_\lambda$ contains all the information in $O_{\lambda'}$.\footnote{Formally, let $\sigma(O)$ denote the $\sigma$-algebra generated by an observation $O$, not to be confused with the nonlinearity $\sigma$ of Eq.~\ref{eq:ffn-update}. By the Doob--Dynkin lemma~\citep{kallenbergFoundationsModernProbability1997}, $O_{\lambda'}$ is a measurable function of $O_\lambda$ if and only if $\sigma(O_{\lambda'})\subseteq\sigma(O_\lambda)$, so $\lambda\succeq\lambda'$ is this inclusion.} Hence any uncertainty estimator of $O_{\lambda'}$ can be computed from $O_\lambda$. The lenses are ordered $\text{str}\succeq\text{rep}\succeq\text{dis}\succeq\text{sem}_R$, where the semantical lens is indexed by $R$.

\begin{corollary}[Lens Information Loss]
    For $\lambda_\text{in}\succeq\lambda_\text{out}$ with $O_\text{out}=g(O_\text{in})$, the lens map $g$ is non-injective on $\mathcal O_{\lambda_\text{in}}$, so no estimator that reads only $O_\text{out}$ can recover $O_\text{in}$. The non-injectivity follows from the form of $g$, independently of the model and the input. \label{claim:information-loss}
\end{corollary}
The map $g_\text{sum}$ maps different splits of component update to the same state, $g_\text{unemb}$ maps $h^{(\ell)}$ and any positive rescaling of $h^{(\ell)}$ to the same prediction, and $g_R$ maps different forms of outputs to the same meaning, depending on $R$. An outer lens cannot in general be a surrogate for an inner one. Corollary~\ref{claim:information-loss} concerns the information available in the observation, not the information usable by an estimator. The usable information typically shrinks outward, but this depends on the estimator family~\citep{xuTheoryUsableInformation2020}. In Section~\ref{sec:lenses-poc}, we measure the $g_\text{unemb}$ and $g_R$ information loss empirically.

The lost information in Corollary~\ref{claim:information-loss} matters only when the target depends on it. Corollary~\ref{claim:lens-sufficiency} fixes a target $T$ and asks whether $T$ depends on what $g$ discards.
\begin{corollary}[Lens Sufficiency]
    Let $g:\mathcal O_{\lambda_\text{in}}\to\mathcal O_{\lambda_\text{out}}$ be one of the lens maps in Lemma~\ref{claim:lens-determination}, and let $T$ be an uncertainty estimation target.
    \begin{itemize}
        \item Permission. If $T=f(O_\text{out})$ for some function $f$, then reading $\lambda_\text{out}$, or any lens inner to it, is sufficient to recover $T$.
        \item Prohibition. If two inner observations $a,b\in\mathcal O_{\lambda_\text{in}}$ have the same outer observation, $g(a)=g(b)$, but different targets, $T(a)\neq T(b)$, then $T$ is not a function of $O_\text{out}$. No estimator that reads only $\lambda_\text{out}$, or a lens outer to it, can recover $T$.
    \end{itemize}\label{claim:lens-sufficiency}
\end{corollary}

For a target $T$, let $\lambda^*$ be the outermost sufficient lens for $T$. By Permission, every lens $\lambda\succeq\lambda^*$ can be read to recover $T$. By Prohibition, no estimator reading only a lens $\lambda\prec\lambda^*$ can recover it. Lens \emph{sufficiency} restricts the lens to $\lambda^*$ or a lens inner to it. Existing uncertainty estimators each read a lens sufficient for their target. Semantic entropy reads $O_\text{sem}$ for meaning-level correctness~\citep{farquharDetectingHallucinationsLarge2024}, predictive entropy reads $O_\text{dis}$ for token-level correctness~\citep{malininUncertaintyEstimationAutoregressive2021}, linear probes read $O_\text{rep}$ for hidden-state features~\citep{azariaInternalStateLLM2023}, and component attribution requires $O_\text{str}$~\citep{sunReDeEPDetectingHallucination2025}.

Corollary~\ref{claim:lens-sufficiency} determines which lenses are sufficient. Among the estimators that read a sufficient lens, the choice depends on \emph{efficiency}, which is set by the variation $Z$ and the auxiliary structure $S$. The lens maps to $\lambda^*$ discard only information irrelevant to $T$, while an estimator reading $\lambda\succ\lambda^*$ must learn to recover $T$ in the presence of this information. Beyond this, estimators reading $\lambda^*$ and those reading an inner lens can differ in $Z$ and $S$. For meaning-level correctness, reading $O_\text{sem}$ usually requires resampling the output, $Z=y$~\citep{farquharDetectingHallucinationsLarge2024, kuhnSemanticUncertaintyLinguistic2023}, while reading $O_\text{rep}$ usually requires a $\mathcal U_S$ fitted on labeled data~\citep{hanSemanticEntropyProbes2024}. The two estimators trade sampling cost against supervision and training cost.

\begin{takeaway}{Choosing a lens by sufficiency and efficiency}
    The four lenses are ordered by the information they carry, and each outer lens discards some information from the adjacent inner lens. For an uncertainty target, whether the target depends on the discarded information determines the \emph{sufficiency} of the lens, and what variable is resampled and what auxiliary structure is used determine the \emph{efficiency} of the estimator. The suited lens is sufficient for the target and admits an efficient estimator.
\end{takeaway}

\subsection{Proof of Concept Experiments}\label{sec:lenses-poc}

This section tests whether the correctness signal decreases along the lens hierarchy. We aim to answer the following questions. \textbf{RQ1}: How much correctness signal can a probe decode from each lens? \textbf{RQ2}: How does this signal change with the model's confidence? \textbf{RQ3}: How much of the decodable signal do established uncertainty estimators use? \textbf{RQ4}: What do the $g_{\text{unemb}}$ and $g_{R}$ gaps contain?

\paragraph{Models and Datasets.} We evaluate Qwen2.5-7B-Instruct, Qwen2.5-7B-Base~\citep{qwenQwen25TechnicalReport2025}, and Llama-3.1-8B-Instruct~\citep{grattafioriLlama3Herd2024} on TriviaQA~\citep{joshiTriviaQALargeScale2017}, PopQA~\citep{mallenWhenNotTrust2023}, and MMLU~\citep{hendrycksMeasuringMassiveMultitask2021}. MMLU is a sanity check. Its answers are single-option letters, so $g_R$ has no lexical form to discard from the distributional lens. Details are in Appendix~\ref{sec:appendix-poc-models-datasets}.

\paragraph{Feature construction.} We build nested feature sets $F_\text{sem} \subset F_\text{dis} \subset F_\text{rep} \subset F_\text{str}$, so the gain of each lens relative to the next outer lens is the contribution of the new features.  The three inner lens features are built on one teacher-forced pass over the greedy answer. The semantical feature is built from the greedy answer and $k=20$ samples, since the $R$ we use reads $O_\text{sem}$ from sampled outputs. The gains to semantical lens measure what single-pass inner features add to sampled semantical features. Details are in Appendix~\ref{sec:appendix-poc-features}.

\begin{table}[ht]
\centering
\caption{Decodability baseline (DecBase) AUROC for correctness prediction. Cell \colorbox{blue!30}{blue} shading is the improvement from the semantical lens in that column.
\textit{Abbreviations}: Qwen-Inst/Qwen-Base = Qwen2.5-7B-Instruct/Base, Llama-Inst = Llama-3.1-8B-Instruct. }\label{tab:decbase}
\vspace{4pt}
\adjustbox{max width=\textwidth}{%
\begin{tabular}{l ccc ccc ccc}
\toprule
& \multicolumn{3}{c}{TriviaQA} & \multicolumn{3}{c}{MMLU} & \multicolumn{3}{c}{PopQA} \\
\cmidrule(lr){2-4}\cmidrule(lr){5-7}\cmidrule(lr){8-10}
\textsc{Lens} & Qwen-Inst & Qwen-Base & Llama-Inst & Qwen-Inst & Qwen-Base & Llama-Inst & Qwen-Inst & Qwen-Base & Llama-Inst \\
\midrule
Structural       & \cellcolor{blue!39}0.910 & \cellcolor{blue!17}0.901 & \cellcolor{blue!18}0.900 & \cellcolor{blue!11}0.807 & \cellcolor{blue!4}0.800  & \cellcolor{blue!7}0.775 & \cellcolor{blue!38}0.960 & \cellcolor{blue!12}0.921 & \cellcolor{blue!20}0.913 \\
Representational & \cellcolor{blue!37}0.908 & \cellcolor{blue!13}0.897 & \cellcolor{blue!17}0.899 & \cellcolor{blue!11}0.807 & \cellcolor{blue!1}0.797  & \cellcolor{blue!6}0.774 & \cellcolor{blue!43}0.965 & \cellcolor{blue!3}0.912  & \cellcolor{blue!14}0.907 \\
Distributional   & \cellcolor{blue!13}0.884 & \cellcolor{blue!9}0.893 & \cellcolor{blue!12}0.894 & \cellcolor{blue!7}0.803  & \cellcolor{blue!4}0.800  & \cellcolor{blue!5}0.773 & \cellcolor{blue!7}0.929  & \cellcolor{blue!0}0.906  & \cellcolor{blue!6}0.899  \\
Semantical       & 0.871 & 0.884 & 0.882 & 0.796 & 0.796 & 0.768 & 0.922 & 0.909 & 0.893 \\
\bottomrule
\end{tabular}
}
\end{table}

\paragraph{Decodability baseline.} The decodability baseline (DecBase) is a supervised probe to describe the correctness information usable by this probe family~\citep{xuTheoryUsableInformation2020}. For TriviaQA and PopQA, we use an $\ell_2$-regularized logistic regression model followed by a one-hidden-layer MLP\@. For MMLU, we use only the logistic-regression layer. We evaluate the probes with AUROC and AUPRC, as detailed in Appendix~\ref{sec:appendix-poc-baseline}.

\paragraph{Uncertainty estimators.} We compare DecBase with one established estimator per lens, including a minimal RAUQ~\citep{vazhentsevEfficientHallucinationDetection2026} that keeps one head per layer, INSIDE~\citep{chenLLMsInternalStates2024}, mean token entropy (TE)~\citep{malininUncertaintyEstimationAutoregressive2021}, and semantic entropy (SE)~\citep{farquharDetectingHallucinationsLarge2024}.

\begin{figure}[!htb]
    \centering
    \includegraphics[width=\linewidth]{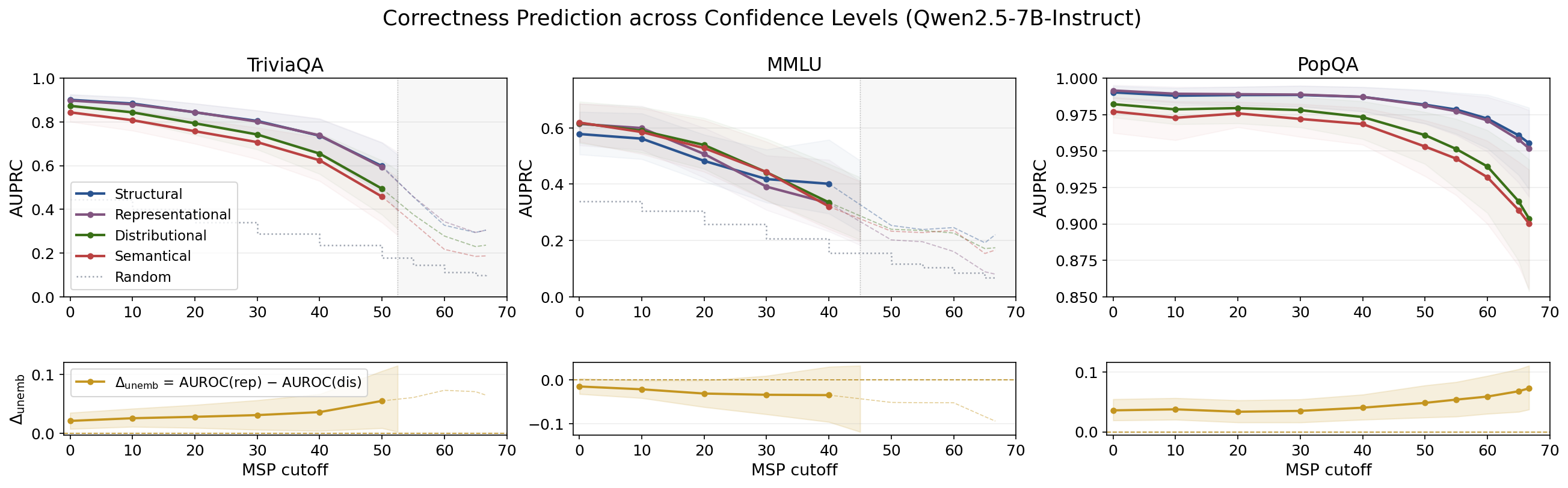}
    \caption{DecBase performance of Qwen-Inst across confidence levels. Top: AUPRC as a function of the MSP lower-percentile cutoff. The gray dashed line marks the chance level baseline, and gray shading marks cutoffs with fewer than 50 incorrect answers. Bottom: marginal gain of $g_\text{unemb}$.}
    \label{fig:correct-confidence}
\end{figure}

\begin{figure}[!htb]
    \centering
    \includegraphics[width=\linewidth]{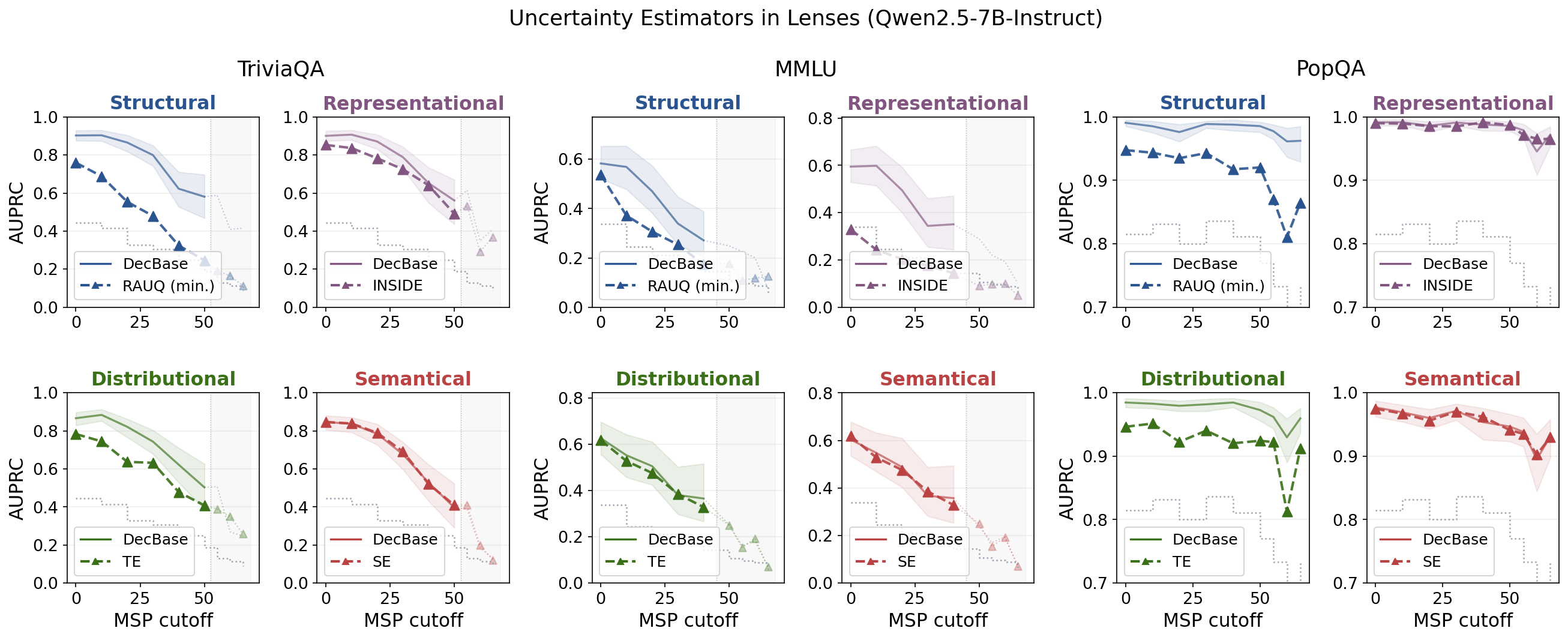}
    \caption{AUPRC of uncertainty estimators and DecBases across confidence levels. The gray dashed line marks the chance level baseline, and gray shading marks cutoffs with fewer than 50 incorrect answers.}
    \label{fig:estimators}
\end{figure}

\begin{figure}[!htb]
    \centering
    \includegraphics[width=0.95\linewidth]{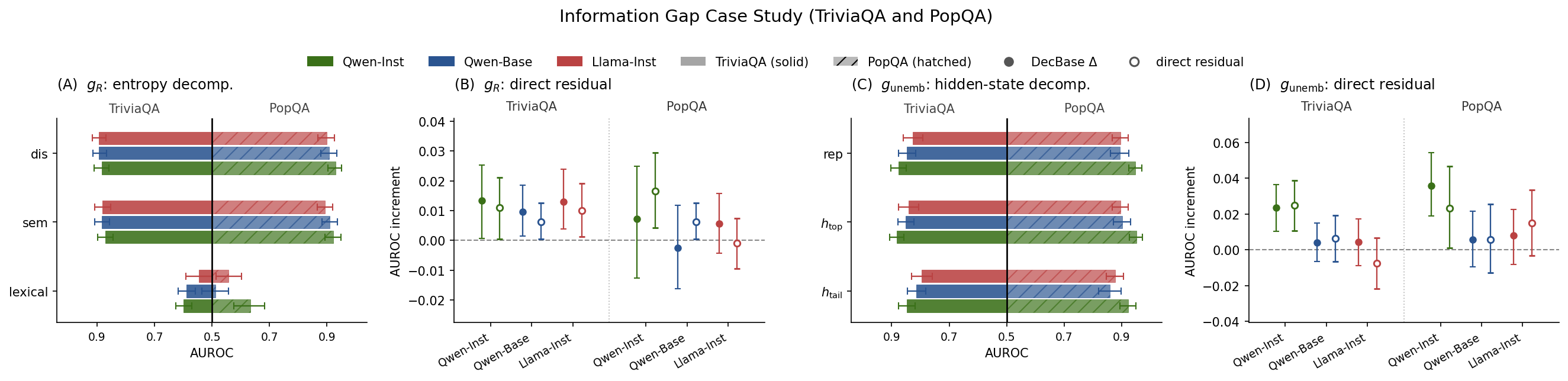}
    \caption{Information gap case study. Panels (A, C) decompose the $g_R$ and $g_\text{unemb}$ gaps, and show the AUROC of each component. Panels (B, D) show the direct residual predictions for the same gaps.}\label{fig:case-study}
\end{figure}

\paragraph{Results.} Table~\ref{tab:decbase} reports the DecBase of each lens for all models and datasets. Figure~\ref{fig:correct-confidence} shows how DecBase changes as confidence rises. Figure~\ref{fig:estimators} compares each estimator with the corresponding DecBase. Figure~\ref{fig:case-study} decomposes the two gaps. For $g_R$, we ask how much correctness signal lives in the lexical variation. For $g_\text{unemb}$, we ask how much lives in the directions of the hidden state that the unembedding expresses weakly.

In Table~\ref{tab:decbase}, the DecBase results are monotonically decreasing from the structural lens to the semantical lens in general. The semantical lens already achieves a high AUROC, and the total gain of the single-pass inner features is real but small using this probe. On MMLU, all increments collapse to near zero because the multiple-choice format suppresses the output diversity. Qwen-Inst shows larger gains, mainly at the $g_\text{unemb}$ gap. Instruction tuning may change how correctness information is spread across the hidden state.

In Figure~\ref{fig:correct-confidence}, we use the mean maximum softmax probability (MSP) of the last token as a confidence score. Raising the MSP cutoff removes less confident questions, leaving fewer errors and harder cases. Confidence and correctness are correlated at every lens. On open-ended QA, the ordering is preserved at every confidence level, and the gaps are wider. On MMLU, the four lines cross. The collapsed output space makes distributional signal the same as the semantical one. The lower panels show the $g_\text{unemb}$ gap $\Delta_\text{unemb}$. The representational lens helps most on confident errors.

Figure~\ref{fig:estimators} shows that each estimator scores at or below the corresponding DecBase. The estimators use only part of the signal that the probe decodes. 
The largest gap is for the minimal RAUQ~\citep{vazhentsevEfficientHallucinationDetection2026} reading only one head per layer. On MMLU, INSIDE~\citep{chenLLMsInternalStates2024} falls to chance level, because the samples produce nearly identical hidden states. TE~\citep{malininUncertaintyEstimationAutoregressive2021} cannot capture the signal of more diverse outputs. SE~\citep{farquharDetectingHallucinationsLarge2024} already captures most of the correctness signal at the semantical lens. DecBase indicates the signal available at each lens, and the estimator's $Z$ and $S$ determine how much of it is recovered.

Figure~\ref{fig:case-study} splits the $g_R$ and $g_\text{unemb}$ gaps into interpretable components, and checks each gap with a direct residual predictor. For $g_R$, the lexical term is the conditional entropy of the surface form given the meaning. It measures how much the phrasing varies within one meaning. The lexical gap is only weakly correlated with correctness. For $g_\text{unemb}$, we split the last-token hidden state into $h_\text{top}$, its projection onto the top-$r$ right singular vectors of $W_U$, and the tail $h_\text{tail} = h - h_\text{top}$, detailed in Appendix~\ref{sec:appendix-poc-additional-results}. The tail directions have small but nonzero singular values. A probe on $h_\text{tail}$ alone reaches a high AUROC, but the probe on the full $h$ is barely better than the one on $h_\text{top}$. The $h_\text{tail}$ carries correctness signal, but is redundant given $h_\text{top}$. The direct residual predictor fits a logit-offset model on the inner features on top of a fixed probe on the outer lens. It tests whether the inner features correct errors the outer prediction misses. Its increments agree with the DecBase increments.

\paragraph{Summary.} We address all four research questions. For \textbf{RQ1}, all four lenses carry correctness signal, and the DecBase mostly decreases outward. Given the sampled semantical features, the inner gains are small. For \textbf{RQ2}, the DecBase of every lens falls as confidence rises. The ordering of the lenses is preserved for open-ended QA and breaks down for multiple-choice QA\@. For \textbf{RQ3}, every estimator uses less signal than the corresponding DecBase. For \textbf{RQ4}, the lexical variation in the $g_R$ gap is only weakly correlated with correctness. The weakly expressed tail $h_\text{tail}$ in the $g_\text{unemb}$ gap predicts correctness on its own, but adds almost nothing given $h_\text{top}$.

% !TEX root = tmlr.tex
\section{UQ Methods Through the Four Lenses}\label{sec:survey}

In this section, we survey uncertainty estimators for four lenses in the reverse order of their hierarchy, from the semantical lens to the structural lens. This does not conflict with the $\text{str}\succeq\text{rep}\succeq\text{dis}\succeq\text{sem}_R$ ordering. The order here follows the literature. Current uncertainty estimation concentrates on the semantical and distributional lenses, while the representational and structural lenses remain under-explored.

We classify each method along the three axes of Section~\ref{sec:lenses-estimator}. The lens $\lambda$ is the observation point. A cross-lens estimator observing two or more lenses is assigned to the innermost lens, and stated explicitly in its description. The variation $Z$ is the resampling variable, and the auxiliary structure $S$ is used to obtain the score. The target column defines what the estimator is designed to measure, and the evaluation column shows the evaluation metrics used. The description of target and evaluation is in Section~\ref{sec:evaluation}.

\subsection{Semantical Lens}\label{sec:lens-sem}

The semantical lens reads $O_\text{sem}=\Pi_R(h^{(\ell)})$ in the semantical space $\Delta^{|\mathcal M_R|-1}$. The set of meanings $\mathcal M_R$ is defined by a semantic relation $R$ between two outputs, and approximates the intent space $\mathcal M$ in Section~\ref{sec:lenses-four}. Depending on what $R$ returns, $\mathcal M_R$ is structured as clusters, a graph, a manifold, or a report. 

We first show how $R$ is defined in practice. The meaning relation in Section~\ref{sec:lenses-four} cannot be observed, so semantical UQ approximates it by comparing outputs with each other. Following Semantic Entropy~\citep{kuhnSemanticUncertaintyLinguistic2023}, two sampled outputs have the same meaning when a model judges them equivalent. Natural Language Inference (NLI) models, LLM judges, and embedding models all realize $R$.

\paragraph{NLI-based semantics.} Natural Language Inference (NLI)~\citep{williamsBroadCoverageChallengeCorpus2018} determines the relationship between two sentences as contradiction, neutral, or entailment. NLI returns a probability vector $(p_c, p_n, p_e)$ over the three labels, produced by small encoder models such as RoBERTa~\citep{liuRoBERTaRobustlyOptimized2019} and DeBERTa~\citep{heDEBERTADECODINGENHANCEDBERT2021}. The pairwise NLI output is graded and directional. So it can be used as a binary relation or a soft score, with or without preserving its asymmetry.

\paragraph{LLM-based semantics.} Equivalence can also be judged by an LLM~\citep{manakulSelfCheckGPTZeroResourceBlackBox2023}. Two responses are placed in a prompt template that asks whether they carry the same meaning, like \emph{``Context: \{response\}. Sentence: \{sentence\}. Is the sentence supported by the context above? Answer Yes or No.''}, and the reply is parsed into a binary label or a score. 

\paragraph{Embedding-based semantics.} The embedding methods map each response to a vector with an embedding model such as Sentence-BERT~\citep{reimersSentenceBERTSentenceEmbeddings2019} or CLIP~\citep{radfordLearningTransferableVisual2021}. The relation is read from the geometry of the embedding space, usually as a similarity or a distance between two representations, so $R$ is continuous.

The three types of models instantiate $R$, and the structure of meanings $\mathcal M_R$ depends on the codomain of $R$ and the conditions it satisfies, summarized in Table~\ref{tab:sem-types}. The cluster type defines $R$ as an equivalence relation that is binary, symmetric, and transitive, such as bidirectional entailment~\citep{kuhnSemanticUncertaintyLinguistic2023, farquharDetectingHallucinationsLarge2024}, so $R$ partitions the outputs into classes. The graph type drops transitivity and allows graded or asymmetric scores, such as graded entailment~\citep{linGeneratingConfidenceUncertainty2024, coxMappingMeaningAddressing2025}, and the manifold type replaces the pairwise judgment with similarity in a shared embedding space~\citep{aoCSSContrastiveSemantic2024}. The report type drops the external $R$, as the model states its own belief in the generated text~\citep{linTeachingModelsExpress2022} and $\mathcal M_R$ is left implicit.
The following subsections survey the estimators built on each structure of $\mathcal M_R$, and the methods are summarized in Table~\ref{tab:sem-survey}.

\begin{table}[ht]
\centering
\caption{Types of the set of meanings $\mathcal M_R$ and semantic relations $R$. ``---'' denotes not applicable.}\label{tab:sem-types}
\renewcommand{\arraystretch}{1.3}
\begin{tabular}{l c l l l}
\toprule
\textsc{Type} & $R$ & $R$ \textsc{Conditions} & $\mathcal M_R$ & $R$ \textsc{Implementations} \\
\midrule
Cluster  & $\{0,1\}$              & Symmetric, transitive & Equivalence classes         & Bidirectional NLI \\
Graph    & $\{0,1\}$ or $[0,1]$   & Sym.\ or asym.        & Nodes in weighted graph     & Graded NLI        \\
Manifold & $\mathbb R$            & Symmetric             & Outputs in metric space   & Embedding similarity         \\
Report   & Internal               & ---                   & Implicit                    & Verbalized confidence        \\
\bottomrule
\end{tabular}
\end{table}

\begin{table}[!htb]
\centering
\small
\caption{Semantical methods survey. \textit{Access}: \accB{} black-box, \accG{} gray-box, \accW{} white-box. \textit{N/S}: Native and surrogate estimators in Section~\ref{sec:lens-axis}, $\circ$ native, $\bullet_{\text{dis}}$ surrogate for the distributional lens, $\bullet_{\text{sem}}$ surrogate for the semantical lens. $Z$: The variation axis in Section~\ref{sec:variation-axis}. $S$: The auxiliary structure in Section~\ref{sec:auxiliary-axis}. \textit{Target} in Section~\ref{sec:eval-target}: \colorbox{navybg}{\cCOR} correctness, \colorbox{yellowbg}{\cUNS} underspecification, \colorbox{skybg}{\cFTH} faithfulness. \textit{Eval. Metrics} in Section~\ref{sec:eval-metrics}: \irnk ranking, \ical calibration, \iscr scoring, \idec decision, \icov coverage, \ifth faithfulness.}
\label{tab:sem-survey}
\begin{tabular}{@{} l p{6.0cm} c >{\scriptsize}c >{\footnotesize}c >{\footnotesize}c c l @{}}
\toprule
\textsc{Type} & \textsc{Method} & \textsc{Access} & \textsc{N/S} & $Z$ & $S$ & \textsc{Target} & \textsc{Eval} \\
\midrule
\multirow{11}{*}{\textbf{Cluster}} & Semantic Entropy \citep{kuhnSemanticUncertaintyLinguistic2023} & \accG & $\circ$ & $y$ & $\varnothing$ & \colorbox{navybg}{\cCOR} & \irnk \\
 & SE Blackbox~\citep{farquharDetectingHallucinationsLarge2024} & \accB & $\circ$ & $y$ & $\varnothing$ & \colorbox{navybg}{\cCOR} & \irnk \\
 & NumSets~\citep{linGeneratingConfidenceUncertainty2024} & \accB & $\circ$ & $y$ & $\varnothing$ & \colorbox{navybg}{\cCOR} & \irnk \\
 & Alphabet Size~\citep{mccabeEstimatingSemanticAlphabet2026} & \accB & $\circ$ & $y$ & $\varnothing$ & \colorbox{navybg}{\cCOR} & \irnk\idec \\
 & Self-Assessment~\citep{caoClusteredSelfAssessmentSimple2026} & \accB & $\circ$ & $y$ & $\varnothing$ & \colorbox{navybg}{\cCOR} & \irnk\iscr \\
 & SRE-UQ~\citep{vipulanandanSemanticUncertaintyQuantification2026} & \accG & $\circ$ & $y$ & $\varnothing$ & \colorbox{navybg}{\cCOR} & \irnk \\
 & SDLG~\citep{aichbergerImprovingUncertaintyEstimation2025} & \accB & $\circ$ & $y$ & $Z_S$ & \colorbox{navybg}{\cCOR} & \irnk \\
 & ConU~\citep{wangConUConformalUncertainty2024} & \accG & $\circ$ & $y$ & $\psi_S$ & \colorbox{navybg}{\cCOR} & \icov \\
 & CP~\citep{mohriLanguageModelsConformal2024} & \accB & $\circ$ & $y$ & $\psi_S$ & \colorbox{navybg}{\cCOR} & \icov \\
 & Input Clarification~\citep{houDecomposingUncertaintyLarge2024} & \accB & $\circ$ & $y,x'$ & $\varnothing$ & \colorbox{yellowbg}{\cUNS} & \irnk \\
 & SAC3~\citep{zhangSAC^3ReliableHallucination2023} & \accB & $\circ$ & $y,x',\theta$ & $Z_S$, $\mathcal U_S$ & \colorbox{navybg}{\cCOR} & \irnk\idec \\
\midrule
\multirow{18}{*}{\textbf{Graph}} & SelfCheckGPT~\citep{manakulSelfCheckGPTZeroResourceBlackBox2023} & \accB & $\circ$ & $y$ & $\varnothing$ & \colorbox{navybg}{\cCOR} & \irnk \\
 & Degree \citep{linGeneratingConfidenceUncertainty2024} & \accB & $\circ$ & $y$ & $\varnothing$ & \colorbox{navybg}{\cCOR} & \irnk \\
 & Eccentricity~\citep{linGeneratingConfidenceUncertainty2024} & \accB & $\circ$ & $y$ & $\varnothing$ & \colorbox{navybg}{\cCOR} & \irnk \\
 & EigV~\citep{linGeneratingConfidenceUncertainty2024} & \accB & $\circ$ & $y$ & $\varnothing$ & \colorbox{navybg}{\cCOR} & \irnk \\
 & KLE~\citep{nikitinKernelLanguageEntropy2024} & \accB/\accG & $\circ$ & $y$ & $\varnothing$ & \colorbox{navybg}{\cCOR} & \irnk \\
 & SD~\citep{qiuSemanticDensityUncertainty2024} & \accB & $\circ$ & $y$ & $\varnothing$ & \colorbox{navybg}{\cCOR} & \irnk \\
 & DEG~\citep{daLLMUncertaintyQuantification2024} & \accB & $\circ$ & $y$ & $\varnothing$ & \colorbox{navybg}{\cCOR} & \irnk \\
 & CLUE~\citep{wangCLUEConceptLevelUncertainty2024} & \accB & $\circ$ & $y$ & $\varnothing$ & \colorbox{navybg}{\cCOR} & \irnk \\
 & LUQ~\citep{zhangLUQLongtextUncertainty2024} & \accB & $\circ$ & $y$ & $\varnothing$ & \colorbox{navybg}{\cCOR} & \irnk \\
 & Graph Uncertainty~\citep{jiangGraphbasedUncertaintyMetrics2024} & \accB & $\circ$ & $y$ & $\varnothing$ & \colorbox{navybg}{\cCOR} & \irnk \\
 & Reasoning Graph~\citep{zhangAllRoadsLead2025} & \accB & $\circ$ & $y$ & $\varnothing$ & \colorbox{navybg}{\cCOR} & \irnk\ical\iscr \\
 & AGSC~\citep{luoAGSCAdaptiveGranularity2026} & \accB & $\circ$ & $y$ & $\mathcal U_S$ & \colorbox{navybg}{\cCOR} & \irnk \\
 & MD-UQ~\citep{chenUncertaintyQuantificationLarge2025} & \accB & $\circ$ & $y$ & $\mathcal U_S$ & \colorbox{navybg}{\cCOR} & \irnk \\
 & TopologyUQ~\citep{daUnderstandingUncertaintyLLM2025} & \accB & $\circ$ & $y$ & $\mathcal U_S$ & \colorbox{skybg}{\cFTH} & \ifth \\
 & Variance~\citep{coxMappingMeaningAddressing2025} & \accB & $\circ$ & $y,x'$ & $\varnothing$ & \colorbox{navybg}{\cCOR} & \irnk \\
 & IUQ~\citep{fanIUQInterrogativeUncertainty2026} & \accB & $\circ$ & $y,x'$ & $\mathcal U_S$ & \colorbox{navybg}{\cCOR} & \irnk \\
 & BTProp~\citep{houProbabilisticFrameworkLLM2025} & \accG & $\circ$ & $y,x'$ & $\mathcal U_S$ & \colorbox{navybg}{\cCOR} & \irnk \\
 & CoTA~\citep{tanneruQuantifyingUncertaintyNatural2024} & \accB & $\circ$ & $y,x'$ & $Z_S$, $\mathcal U_S$ & \colorbox{skybg}{\cFTH} & \ifth \\
\midrule
\multirow{6}{*}{\textbf{Manifold}} & SelfCheckGPT~\citep{manakulSelfCheckGPTZeroResourceBlackBox2023} & \accB & $\circ$ & $y$ & $\varnothing$ & \colorbox{navybg}{\cCOR} & \irnk \\
 & CSS~\citep{aoCSSContrastiveSemantic2024} & \accB & $\circ$ & $y$ & $\varnothing$ & \colorbox{navybg}{\cCOR} & \irnk \\
 & MATU~\citep{chenEveryResponseCounts2026} & \accB & $\circ$ & $y$ & $\varnothing$ & \colorbox{navybg}{\cCOR} & \irnk \\
 & GeoU~\citep{phillipsGeometricUncertaintyDetecting2026} & \accB & $\circ$ & $y$ & $\mathcal U_S$ & \colorbox{navybg}{\cCOR} & \irnk \\
 & SpecU~\citep{walhaFinegrainedUncertaintyDecomposition2026} & \accB & $\circ$ & $y,x'$ & $Z_S$ & \colorbox{navybg}{\cCOR} & \irnk\iscr \\
 & SPUQ~\citep{gaoSPUQPerturbationBasedUncertainty2024} & \accB & $\circ$ & $y,x'$ & $Z_S$ & \colorbox{navybg}{\cCOR} & \ical \\
\bottomrule
\end{tabular}
\end{table}

\begin{table}[!htb]
\centering
\small
\begin{tabular}{@{} l p{6.0cm} c >{\scriptsize}c >{\footnotesize}c >{\footnotesize}c c l @{}}
\multicolumn{8}{l}{\small\textit{Table~\ref{tab:sem-survey} continued from previous page}} \\
\toprule
\textsc{Type} & \textsc{Method} & \textsc{Access} & \textsc{N/S} & $Z$ & $S$ & \textsc{Target} & \textsc{Eval} \\
\midrule
\multirow{18}{*}{\textbf{Report}} & Verbalized Confidence~\citep{linTeachingModelsExpress2022} & \accB & $\circ$ & $\varnothing$ & $\varnothing$ & \colorbox{navybg}{\cCOR} & \ical \\
 & Knowledge Transfer~\citep{yangCanWeTrust2024} & \accB & $\circ$ & $\varnothing$ & $\varnothing$ & \colorbox{navybg}{\cCOR} & \ical \\
 & SaySelf~\citep{xuSaySelfTeachingLLMs2024} & \accB & $\circ$ & $\varnothing$ & $\varnothing$ & \colorbox{navybg}{\cCOR} & \ical\irnk \\
 & Ling. Calib.~\citep{bandLinguisticCalibrationLongForm2024} & \accB & $\circ$ & $\varnothing$ & $\varnothing$ & \colorbox{navybg}{\cCOR} & \ical \\
 & CONQORD~\citep{taoWhenTrustLLMs2024} & \accB & $\circ$ & $\varnothing$ & $\varnothing$ & \colorbox{navybg}{\cCOR} & \ical \\
 & Ask for Calibration~\citep{tianJustAskCalibration2023} & \accB & $\circ$ & $\varnothing$ & $\varnothing$ & \colorbox{navybg}{\cCOR} & \ical\irnk\iscr \\
 & UaIT~\citep{liuCanLLMsLearn2024} & \accB & $\circ$ & $\varnothing$ & $\varnothing$ & \colorbox{navybg}{\cCOR} & \irnk \\
 & MetaFaith~\citep{liuMetaFaithFaithfulNatural2025} & \accB & $\circ$ & $\varnothing$ & $\varnothing$ & \colorbox{skybg}{\cFTH} & \ifth\ical\iscr \\
 & RLMF~\citep{liuReinforcementLearningMetacognitive2026} & \accB & $\circ$ & $\varnothing$ & $\varnothing$ & \colorbox{skybg}{\cFTH} & \ifth\ical\iscr \\
 & CoTA~\citep{tanneruQuantifyingUncertaintyNatural2024} & \accB & $\circ$ & $\varnothing$ & $\varnothing$ & \colorbox{skybg}{\cFTH} & \ifth \\
 & DINCO~\citep{wangCalibratingVerbalizedConfidence2026} & \accB & $\circ$ & $y$ & $Z_S$ & \colorbox{navybg}{\cCOR} & \ical\irnk\iscr \\
 & LM vs LM~\citep{cohenLMVsLM2023} & \accB & $\circ$ & $y$ & $\mathcal U_S$ & \colorbox{navybg}{\cCOR} & \idec \\
 & HCPD~\citep{yangZerosourceLLMHallucination2026} & \accB & $\circ$ & $y$ & $\mathcal U_S$ & \colorbox{navybg}{\cCOR} & \irnk \\
 & BSDetector~\citep{chenQuantifyingUncertaintyAnswers2024} & \accB & $\circ$ & $y$ & $\mathcal U_S$ & \colorbox{navybg}{\cCOR} & \irnk \\
 & Decisiveness~\citep{yonaCanLargeLanguage2024} & \accB & $\circ$ & $y$ & $\mathcal U_S$ & \colorbox{skybg}{\cFTH} & \ifth \\
 & Enhanced CP~\citep{cherianLargeLanguageModel2024} & \accB & $\circ$ & $y$ & $\psi_S$ & \colorbox{navybg}{\cCOR} & \irnk\icov \\
 & T3~\citep{liThinkTwiceTrusting2024} & \accB & $\circ$ & $x'$ & $\varnothing$ & \colorbox{navybg}{\cCOR} & \irnk\ical \\
 & SPUQ~\citep{gaoSPUQPerturbationBasedUncertainty2024} & \accB & $\circ$ & $y,x'$ & $Z_S$ & \colorbox{navybg}{\cCOR} & \ical \\
\bottomrule
\end{tabular}
\end{table}

\subsubsection{Cluster Type}\label{sec:lens-sem-cluster}

The cluster type makes the strongest assumption on $R$ and constructs the simplest $\mathcal M_R$. $R$ is binary, symmetric, and transitive, so the semantic models partition the responses into equivalence classes, and estimators read the distribution over these classes.

\paragraph{NLI-based semantics.}
Semantic Entropy (SE)~\citep{kuhnSemanticUncertaintyLinguistic2023} is the foundational work. Two sampled responses are equivalent if each entails the other, and the entropy is computed over the semantic clusters. Sampling with a fixed prompt gives $Z=y$, and the entropy reads $\Pi_R$ directly, giving $S=\varnothing$. The original formulation weights clusters by the sequence likelihood of its members and so needs gray-box access. A black-box variant weights with the empirical cluster frequency~\citep{farquharDetectingHallucinationsLarge2024}. NumSets~\citep{linGeneratingConfidenceUncertainty2024} counts the number of clusters. \citet{mccabeEstimatingSemanticAlphabet2026} show that both approximations underestimate semantic entropy, and their alphabet-size estimator corrects this bias.

The other NLI-based methods keep the partition and change another step of the pipeline. Before sampling, SRE-UQ~\citep{vipulanandanSemanticUncertaintyQuantification2026} recalibrates the token distribution by entropy maximization and then computes the semantic entropy, making it \emph{cross-lens}. It folds a distributional signal into a semantical readout. At sampling, Semantically Diverse Language Generation (SDLG)~\citep{aichbergerImprovingUncertaintyEstimation2025} steers the LLM towards semantically diverse responses, an instance of $S=Z_S$ reshaping the $Z=y$ distribution. After clustering, Self-Assessment~\citep{caoClusteredSelfAssessmentSimple2026} turns the clusters of an open-ended generation into the options of a multiple-choice question and reads the choice probabilities.

\paragraph{LLM-based semantics.}
An LLM judge is more flexible. Input Clarification~\citep{houDecomposingUncertaintyLarge2024} generates clarifications of the input, feeds them to the LLM, and ensembles the predictions, varying $Z=x'$ and $Z=y$. It targets underspecification, i.e., ambiguity of the prompt. SAC3~\citep{zhangSAC^3ReliableHallucination2023} uses a similar judge over question--answer pairs and samples across paraphrased inputs and across models, giving $Z=y,x',\theta$.

Beyond equivalence checking, Conformal Prediction~\citep{mohriLanguageModelsConformal2024} and ConU~\citep{wangConUConformalUncertainty2024} take LLM-judged semantic consistency as the nonconformity score. Here $S=\psi_S$ wraps the semantical observation in a conformal framework, giving a prediction set with coverage.

\subsubsection{Graph Type}\label{sec:lens-sem-graph}

The graph type keeps the asymmetry or graded features of $R$ or both. Responses are defined as nodes, and the edges may be weighted or unweighted and directed or undirected. Estimators build this graph explicitly or implicitly and read its spectral or centrality functionals. $\mathcal U$ reads the structure of $\mathcal M_R$ as well as $\Pi_R$.

\paragraph{NLI-based semantics.}
The graded symmetric graph is the direct successor to the cluster type. \citet{linGeneratingConfidenceUncertainty2024} use the softmax probability of entailment or contradiction as the edge weight, and measure uncertainty by the Degree matrix, Eccentricity, and sum of eigenvalues of the graph Laplacian (EigV)\@. Later estimators keep this construction and change the readout. The variance method~\citep{coxMappingMeaningAddressing2025} reads the spectral embedding of the affinity graph, KLE~\citep{nikitinKernelLanguageEntropy2024} computes the von Neumann entropy of a positive semi-definite heat kernel built on the graph, and Semantic Density~\citep{qiuSemanticDensityUncertainty2024} reweights the edges as $\mathbb{E}(\parallel v_1 - v_2 \parallel^2)=p_c+\frac{1}{2}p_n$ before readout. SelfCheckGPT~\citep{manakulSelfCheckGPTZeroResourceBlackBox2023} evaluates consistency with different semantics models. Its NLI and LLM variants form an undirected graph, its BERTScore~\citep{zhangBERTScoreEvaluatingText2020} variant belongs to the manifold type, and its $n$-gram variant measures lexical similarity, so it observes the distribution rather than the meanings.

To account for the asymmetry of entailment, the Directional Entailment Graph (DEG)~\citep{daLLMUncertaintyQuantification2024} builds a directed graph whose edge weights are the softmax of $(p_c, p_n, p_e)$, and reads uncertainty from the eigenvalues of its random-walk Laplacian.

The nodes can be finer-grained than the whole response. Long-text Uncertainty Quantification (LUQ)~\citep{zhangLUQLongtextUncertainty2024} builds the graph over sentences and pools upward, and CLUE~\citep{wangCLUEConceptLevelUncertainty2024} splits the output into concepts scored by a one-directional relation from a sequence to a concept. Adaptive Granularity and GMM-based Semantic Clustering (AGSC)~\citep{luoAGSCAdaptiveGranularity2026} adopts adaptive granularity. It decomposes NLI-neutral pairs into atomic claims and pools them with a Gaussian mixture model. Belief Tree Propagation (BTProp)~\citep{houProbabilisticFrameworkLLM2025} builds a belief tree over decomposed statements whose logical relations are given by NLI, and Chain-of-Thought Agreement (CoTA)~\citep{tanneruQuantifyingUncertaintyNatural2024} applies the same machinery to reasoning chains. CoTA also changes the target to the faithfulness of reasoning explanations, and scores their consistency either with an NLI graph or with the verbalized report.

\paragraph{LLM-based semantics.}

An LLM judge can also define a graded, asymmetric relation. Graph Uncertainty~\citep{jiangGraphbasedUncertaintyMetrics2024} uses it to relate responses to the claims, giving a bipartite graph read by a family of centrality metrics. Interrogative Uncertainty Quantification (IUQ)~\citep{fanIUQInterrogativeUncertainty2026} turns each claim into a self-contained query and reads the claim sequence. Multi-Dimensional Uncertainty Quantification (MD-UQ)~\citep{chenUncertaintyQuantificationLarge2025} stacks the NLI-based and LLM-based semantics into a tensor and reads uncertainty from its decomposition. Reasoning graphs can also be built with an LLM judge. TopologyUQ~\citep{daUnderstandingUncertaintyLLM2025} extracts an explanation graph per reasoning chain and scores their disagreement by graph edit distance. Reasoning Graph~\citep{zhangAllRoadsLead2025} merges all reasoning traces into one graph.

\subsubsection{Manifold Type}\label{sec:lens-sem-manifold}

The manifold type places a response among all possible meanings. An embedding model maps every response to a vector, and $R$ becomes a distance or an inner product in the embedding space. $\mathcal U$ reads a geometric property of the metric space, i.e., the structure of $\mathcal M_R$ and $\Pi_R$.

\paragraph{Embedding-based semantics.}

The earliest manifold type UQ is the BERTScore variant of SelfCheckGPT~\citep{manakulSelfCheckGPTZeroResourceBlackBox2023}, which compares responses by their BERTScore cosine similarity.
SPUQ~\citep{gaoSPUQPerturbationBasedUncertainty2024} uses $S=Z_S$ to define the perturbation sets that generate $Z=x'$, and scores the outputs' consitency with BERTScore, SentenceBERT, verbalized score, and RougeL\@. Each model defines a different $R$. BERTScore and SentenceBERT lie in the manifold type, the verbalized score lies in the report type, and RougeL belongs to the distributional lens. The same $Z$ and $S$ can serve different observations. Beyond BERT-based embeddings, Contrastive Semantic Similarity (CSS)~\citep{aoCSSContrastiveSemantic2024} embeds responses with CLIP features and computes centrality metrics (Ecc, EigV, Deg) of the graph Laplacian. Geometric Uncertainty (GeoU)~\citep{phillipsGeometricUncertaintyDetecting2026} and Spectral Uncertainty (SpecU)~\citep{walhaFinegrainedUncertaintyDecomposition2026} embed responses with a sentence encoder and read archetypal analysis and spectral decomposition, respectively.

\paragraph{LLM-based semantics.}
Multi-Agent Tensor Uncertainty (MATU)~\citep{chenEveryResponseCounts2026} embeds the interactions between agents with an LLM and decomposes the tensor.

\subsubsection{Report Type}\label{sec:lens-sem-report}

LLMs can be prompted to state their uncertainty in natural language, and $R$ is internal to the model's generation. The LLM converts its belief about semantic-level confidence into text, either a linguistic hedge such as \emph{``almost certain''} or a number such as \emph{``90\% confident''}, and $\mathcal M_R$ is not constructed explicitly. The report type expresses the semantical uncertainty as output. The observation carries the model's belief about its own certainty, but it is read from the surface form of the generated text.

Verbalized uncertainty was first elicited by supervised fine-tuning~\citep{linTeachingModelsExpress2022} and then by prompting~\citep{tianJustAskCalibration2023}. Later work changes the training signal to improve calibration, with reinforcement learning against self-reflective rationales in SaySelf~\citep{xuSaySelfTeachingLLMs2024}, reinforcement learning from human feedback in CONQORD~\citep{taoWhenTrustLLMs2024}, distillation from a larger, better-calibrated model in Knowledge Transfer~\citep{yangCanWeTrust2024}, distillation from an attention-relevance estimator in UaIT~\citep{liuCanLLMsLearn2024}, and a reward from a surrogate reader in linguistic calibration~\citep{bandLinguisticCalibrationLongForm2024}. At inference, all of them reduce to $(Z, S)=(\varnothing,\varnothing)$, since verbalization needs no resampling and no external structure.

Verbalized uncertainty is often overconfident, and another line of work resolves this at inference. One route is to apply resampling. DINCO~\citep{wangCalibratingVerbalizedConfidence2026} uses $Z=y$ to generate distractor responses. HCPD~\citep{yangZerosourceLLMHallucination2026} uses $Z=y$ to decompose judgments into human-interpretable parts. T3~\citep{liThinkTwiceTrusting2024} uses $Z=x'$ to prompt the model to think twice. The other route adds an external source, including another model's verbalization in LM vs LM~\citep{cohenLMVsLM2023}, an NLI-based semantic similarity in BSDetector~\citep{chenQuantifyingUncertaintyAnswers2024}, and a conformal post-processing in enhanced conformal prediction~\citep{cherianLargeLanguageModel2024}.

These methods target correctness. One can also ask whether the verbalized uncertainty is faithful to the model's performance. Decisiveness~\citep{yonaCanLargeLanguage2024} is defined as the model's belief in its own correctness and faithfulness is its agreement with the consistency-based uncertainty~\citep{manakulSelfCheckGPTZeroResourceBlackBox2023}. MetaFaith~\citep{liuMetaFaithFaithfulNatural2025} and RLMF~\citep{liuReinforcementLearningMetacognitive2026} improve this agreement by prompt-based calibration and reinforcement learning.

\begin{takeaway}{Semantical estimators measure the spread of meanings}
    All semantical methods are native estimators that read a semantical observation and measure the spread of meanings. The relation $R$ moves from cluster to graph to manifold, which weakens the assumptions on meaning. Verbalization branches off this line and lets the model itself judge meaning.
\end{takeaway}

\subsection{Distributional Lens}\label{sec:lens-dis}

The distributional lens reads the model's prediction over the vocabulary $O_\text{dis}=\Pi(h^{(\ell)})$. Methods that read the logits or apply a readout map $\rho$ read the same lens. Table~\ref{tab:dis-survey} summarizes them by observation granularity. 

We divide the distributional methods into token, sequence, and claim levels. A token-level method reads the atom $\Pi(h^{(\ell)}(x^t))$ at one position $t$. A sequence-level method reads the probability of a full output in Eq.~\ref{eq:autoregressive-factorisation}. A claim-level method splits the output into claims and reads the product over the tokens of claims, which is coarser than a token and finer than the sequence. Within each level, we start from the closed-form functions at $S=\varnothing$ that define distributional uncertainty, and then group other methods by the role of $S$.

\begin{table}[!htb]
\centering
\small
\caption{Distributional methods survey. \textit{Access}: \accB{} black-box, \accG{} gray-box, \accW{} white-box. \textit{N/S}: Native and surrogate estimators in Section~\ref{sec:lens-axis}, $\circ$ native, $\bullet_{\text{dis}}$ surrogate for the distributional lens, $\bullet_{\text{sem}}$ surrogate for the semantical lens. $Z$: The variation axis in Section~\ref{sec:variation-axis}. $S$: The auxiliary structure in Section~\ref{sec:auxiliary-axis}. \textit{Target} in Section~\ref{sec:eval-target}: \colorbox{navybg}{\cCOR} correctness. \textit{Eval. Metrics} in Section~\ref{sec:eval-metrics}: \irnk ranking, \ical calibration, \iscr scoring, \idec decision, \icov coverage.}
\label{tab:dis-survey}
\begin{tabular}{@{} l p{6.3cm} c >{\scriptsize}c >{\footnotesize}c >{\footnotesize}c c l @{}}
\toprule
\textsc{Level} & \textsc{Method} & \textsc{Access} & \textsc{N/S} & $Z$ & $S$ & \textsc{Target} & \textsc{Eval} \\
\midrule
\multirow{22}{*}{\textbf{Token}} & MSP~\citep{hendrycksBaselineDetectingMisclassified2017} & \accG & $\circ$ & $\varnothing$ & $\varnothing$ & \colorbox{navybg}{\cCOR} & \irnk \\
 & LogTokU~\citep{maEstimatingLLMUncertainty2025} & \accG & $\circ$ & $\varnothing$ & $\varnothing$ & \colorbox{navybg}{\cCOR} & \irnk \\
 & U-LINE~\citep{vashurinUNCERTAINTYLINELengthInvariantEstimation2025} & \accG & $\circ$ & $\varnothing$ & $\psi_S$ & \colorbox{navybg}{\cCOR} & \irnk \\
 & Semantic Energy~\citep{maSemanticEnergyDetecting2025} & \accG & $\circ$ & $y$ & $\varnothing$ & \colorbox{navybg}{\cCOR} & \irnk \\
 & LoFree CP~\citep{suAPIEnoughConformal2024} & \accB & $\circ$ & $y$ & $\psi_S$ & \colorbox{navybg}{\cCOR} & \icov \\
 & CAPE~\citep{jiangCalibratingLanguageModels2023} & \accG & $\circ$ & $x'$ & $Z_S$ & \colorbox{navybg}{\cCOR} & \ical\irnk\iscr \\
 & Agreement~\citep{portillowightmanStrengthNumbersEstimating2023} & \accG & $\circ$ & $x'$ & $Z_S$ & \colorbox{navybg}{\cCOR} & \ical\irnk \\
 & Stable Explanation~\citep{beckerCyclesThoughtMeasuring2024} & \accG & $\circ$ & $x'$ & $Z_S$ & \colorbox{navybg}{\cCOR} & \irnk\ical \\
 & MCQA CP~\citep{kumarConformalPredictionLarge2023} & \accG & $\circ$ & $x'$ & $\psi_S$ & \colorbox{navybg}{\cCOR} & \ical\icov \\
 & LLM Judges~\citep{jungTrustEscalateLLM2025} & \accB & $\circ$ & $x'$ & $\psi_S$ & \colorbox{navybg}{\cCOR} & \irnk\ical \\
 & MC~\citep{mora-crossUncertaintyEstimationLarge2024} & \accG & $\circ$ & $\theta$ & $Z_S$ & \colorbox{navybg}{\cCOR} & \ical \\
 & LoRA Ens.~\citep{balabanovUncertaintyQuantificationFinetuned2025} & \accG & $\circ$ & $\theta$ & $Z_S$ & \colorbox{navybg}{\cCOR} & \irnk\ical \\
 & TokUR~\citep{zhangTokURTokenLevelUncertainty2026} & \accW & $\circ$ & $\theta$ & $Z_S$ & \colorbox{navybg}{\cCOR} & \irnk\idec \\
 & BLoRA~\citep{yangBayesianLowrankAdaptation2024} & \accG & $\circ$ & $\theta$ & $Z_S$ & \colorbox{navybg}{\cCOR} & \ical\iscr \\
 & BLoB~\citep{wangBLoBBayesianLowRank2024} & \accG & $\circ$ & $\theta$ & $Z_S$ & \colorbox{navybg}{\cCOR} & \ical\iscr \\
 & MUSE~\citep{kruseSimpleEffectiveInformationTheoretic2025} & \accG & $\circ$ & $\theta$ & $Z_S$ & \colorbox{navybg}{\cCOR} & \irnk\ical\iscr \\
 & PE ICL~\citep{lingUncertaintyQuantificationInContext2024} & \accG & $\circ$ & $x',\theta$ & $Z_S$ & \colorbox{navybg}{\cCOR} & \irnk \\
 & P(True)~\citep{kadavathLanguageModelsMostly2022} & \accG & $\bullet_{\text{sem}}$ & $\varnothing$ & $\varnothing$ & \colorbox{navybg}{\cCOR} & \ical \\
 & Self-Evaluation~\citep{renSelfEvaluationImprovesSelective2023} & \accG & $\bullet_{\text{sem}}$ & $\varnothing$ & $\varnothing$ & \colorbox{navybg}{\cCOR} & \irnk\ical \\
 & COOPERATE~\citep{fengDontHallucinateAbstain2024} & \accG & $\bullet_{\text{sem}}$ & $\varnothing$ & $\varnothing$ & \colorbox{navybg}{\cCOR} & \idec \\
 & TokenSAR~\citep{duanShiftingAttentionRelevance2024} & \accG & $\bullet_{\text{sem}}$ & $y$ & $\mathcal U_S$ & \colorbox{navybg}{\cCOR} & \irnk \\
 & CoT-UQ~\citep{zhangCoTUQImprovingResponsewise2025} & \accG & $\bullet_{\text{sem}}$ & $y$ & $\mathcal U_S$ & \colorbox{navybg}{\cCOR} & \irnk \\
\midrule
\multirow{24}{*}{\textbf{Sequence}} & Leng.-Norm. SCR~\citep{malininUncertaintyEstimationAutoregressive2021} & \accG & $\circ$ & $\varnothing$ & $\varnothing$ & \colorbox{navybg}{\cCOR} & \irnk \\
 & Selective Answering~\citep{coleSelectivelyAnsweringAmbiguous2023} & \accG & $\circ$ & $\varnothing$ & $\varnothing$ & \colorbox{navybg}{\cCOR} & \irnk\ical \\
 & Perplexity~\citep{fomichevaUnsupervisedQualityEstimation2020} & \accG & $\circ$ & $\varnothing$ & $\varnothing$ & \colorbox{navybg}{\cCOR} & \irnk \\
 & G-NLL~\citep{aichbergerRethinkingUncertaintyEstimation2026} & \accG & $\circ$ & $\varnothing$ & $\varnothing$ & \colorbox{navybg}{\cCOR} & \irnk \\
 & Rényi Divergence~\citep{darrinRainProofUmbrellaShield2023} & \accG & $\circ$ & $\varnothing$ & $\varnothing$ & \colorbox{navybg}{\cCOR} & \irnk \\
 & Fisher-Rao Dist.~\citep{darrinRainProofUmbrellaShield2023} & \accG & $\circ$ & $\varnothing$ & $\varnothing$ & \colorbox{navybg}{\cCOR} & \irnk \\
 & UProp~\citep{duanUPropInvestigatingUncertainty2025} & \accG & $\circ$ & $\varnothing$ & $\varnothing$ & \colorbox{navybg}{\cCOR} & \irnk \\
 & Internal Consistency~\citep{xieCalibratingReasoningLanguage2024} & \accW & $\circ$ & $\varnothing$ & $\varnothing$ & \colorbox{navybg}{\cCOR} & \ical \\
 & Self-Certainty~\citep{kangScalableBestofNSelection2025} & \accG & $\circ$ & $\varnothing$ & $\varnothing$ & \colorbox{navybg}{\cCOR} & \idec \\
 & SAUP~\citep{zhaoUncertaintyPropagationLLM2025} & \accG & $\circ$ & $\varnothing$ & $\mathcal U_S$ & \colorbox{navybg}{\cCOR} & \irnk \\
 & DisAAD~\citep{cuiEstimatingBlackboxLLM2026} & \accB & $\circ$ & $\varnothing$ & $\mathcal U_S$ & \colorbox{navybg}{\cCOR} & \irnk\ical\iscr \\
 & Consistency~\citep{lyuCalibratingLargeLanguage2025} & \accB & $\circ$ & $y$ & $\varnothing$ & \colorbox{navybg}{\cCOR} & \irnk\ical\iscr \\
 & SelfCheckGPT~\citep{manakulSelfCheckGPTZeroResourceBlackBox2023} & \accB & $\circ$ & $y$ & $\mathcal U_S$ & \colorbox{navybg}{\cCOR} & \irnk \\
 & Subjective UQ~\citep{wangSubjectiveUncertaintyQuantification2025} & \accB & $\circ$ & $y$ & $\mathcal U_S$ & \colorbox{navybg}{\cCOR} & \irnk\ical \\
 & Mutual Information~\citep{yadkoriBelieveNotBelieve2024} & \accB & $\circ$ & $y$ & $\mathcal U_S$ & \colorbox{navybg}{\cCOR} & \irnk \\
 & CoCoA~\citep{vashurinCoCoAMinimumBayes2025} & \accG & $\circ$ & $y$ & $\mathcal U_S$ & \colorbox{navybg}{\cCOR} & \irnk \\
 & SNNE~\citep{nguyenSemanticEntropyBoosting2025} & \accB & $\circ$ & $y$ & $\mathcal U_S$ & \colorbox{navybg}{\cCOR} & \irnk \\
 & CoRAP~\citep{liQuantifyingUnderstandingUncertainty2026} & \accG & $\circ$ & $y$ & $\psi_S$ & \colorbox{navybg}{\cCOR} & \icov \\
 & DeepConf~\citep{fuDeepThinkConfidence2026} & \accG & $\circ$ & $y$ & $\psi_S$ & \colorbox{navybg}{\cCOR} & \idec \\
 & LI~\citep{kimDetectingLLMHallucination2025} & \accW & $\circ$ & $x'$ & $\varnothing$ & \colorbox{navybg}{\cCOR} & \irnk\ical \\
 & SentenceSAR~\citep{duanShiftingAttentionRelevance2024} & \accG & $\bullet_{\text{sem}}$ & $y$ & $\mathcal U_S$ & \colorbox{navybg}{\cCOR} & \irnk \\
 & LARS~\citep{yaldizNotDesignLearn2025} & \accG & $\bullet_{\text{sem}}$ & $y$ & $\mathcal U_S$ & \colorbox{navybg}{\cCOR} & \irnk \\
 & COMPETE~\citep{fengDontHallucinateAbstain2024} & \accG & $\bullet_{\text{sem}}$ & $x'$ & $\varnothing$ & \colorbox{navybg}{\cCOR} & \idec \\
 & MARS~\citep{bakmanMARSMeaningAwareResponse2024} & \accG & $\bullet_{\text{sem}}$ & $\varnothing / y$ & $\mathcal U_S$ & \colorbox{navybg}{\cCOR} & \irnk \\
\midrule
\multirow{2}{*}{\textbf{Claim}} & CCP~\citep{fadeevaFactCheckingOutputLarge2024} & \accG & $\circ$ & $\varnothing$ & $\mathcal U_S$ & \colorbox{navybg}{\cCOR} & \irnk \\
 & Conformal LM~\citep{quachConformalLanguageModeling2024} & \accG & $\circ$ & $y$ & $\psi_S$ & \colorbox{navybg}{\cCOR} & \icov \\
\bottomrule
\end{tabular}
\end{table}

\subsubsection{Token-level Distribution}\label{sec:lens-dis-token}

The token distribution $\Pi(h^{(\ell)}(x^t))$ is the simplest and most common baseline for UQ\@. The softmax output is spread over $V$, and can be used for single-token tasks such as multiple-choice question answering, or score a longer output by its last token.

\paragraph{Closed-form functions $S=\varnothing$.}
The simplest distributional estimators are closed-form functions of $\Pi(h^{(\ell)}(x^t))$ at $Z=\varnothing$ and $S=\varnothing$, like maximum softmax probability (MSP)~\citep{hendrycksBaselineDetectingMisclassified2017}, which takes the largest entry of the distribution as a confidence score.

LogTokU~\citep{maEstimatingLLMUncertainty2025} uses the logits $z(h^{(\ell)}(x^t))$ as the evidence of a Dirichlet distribution. Semantic Energy~\citep{maSemanticEnergyDetecting2025} reads the logit with output resampling $Z=y$. Logit signals can steer decoding away from hallucinations~\citep{vanderpoelMutualInformationAlleviates2022,yinReasoningFluxEnhancing2024}.

P(True)~\citep{kadavathLanguageModelsMostly2022} and Self-Evaluation~\citep{renSelfEvaluationImprovesSelective2023} read the probability of one specific token, such as \emph{``True''} or \emph{``Correct''}. COOPERATE~\citep{fengDontHallucinateAbstain2024} uses the same probability to decide whether to abstain, and COMPETE~\citep{fengDontHallucinateAbstain2024} is its sequence-level extension. The token is chosen manually based on its meaning, so these methods are surrogates for the semantics.

\paragraph{Sampling structures $S=Z_S$.}
Many methods introduce $S=Z_S$ to define the distribution of input prompts $x'$ or model parameters $\theta$, and aggregate the token-level score over the samples. The prompt variants can be human-crafted in CAPE~\citep{jiangCalibratingLanguageModels2023} and Stable Explanation~\citep{beckerCyclesThoughtMeasuring2024}, or generated by another language model in Agreement~\citep{portillowightmanStrengthNumbersEstimating2023}. The same strategy extends naturally to sequence-level sampling.

Sampling $\theta$ is the Bayesian view~\citep{malininUncertaintyEstimationAutoregressive2021}. Due to the number of parameters, the posterior over $\theta$ of an LLM can only be an approximation. $Z_S$ defines the approximation via Monte-Carlo Dropout~\citep{mora-crossUncertaintyEstimationLarge2024}, low-rank adaptation (LoRA) ensembles~\citep{balabanovUncertaintyQuantificationFinetuned2025}, Bayesian LoRA for reasoning~\citep{zhangTokURTokenLevelUncertainty2026}, Bayesian Laplace LoRA~\citep{yangBayesianLowrankAdaptation2024}, Bayesian Low-Rank Adaptation by Backpropagation (BLoB)~\citep{wangBLoBBayesianLowRank2024}, or the information-theoretic subset method MUSE~\citep{kruseSimpleEffectiveInformationTheoretic2025}. These approximations transfer to the sequence level. Predictive Entropy In-Context Learning~\citep{lingUncertaintyQuantificationInContext2024} combines input and parameter sampling in one estimator.

\paragraph{External functions $S=\mathcal U_S$.}
Some methods implement $\mathcal U_S$ as an external model that converts $\Pi(h^{(\ell)}(x^t))$ into a score. TokenSAR~\citep{duanShiftingAttentionRelevance2024} weights each token distribution by its relevance to the meaning, judged by NLI, and SentenceSAR extends it to sequence level. CoT-UQ~\citep{zhangCoTUQImprovingResponsewise2025} uses the LLM to extract keywords and weights the corresponding tokens. Both methods aim for semantics.

\paragraph{Post-hoc wrappers $S=\psi_S$.}
The last group wraps the score with $\psi_S$. Conformal prediction provides coverage guarantees. Multiple-Choice Question Answering CP~\citep{kumarConformalPredictionLarge2023} uses the softmax probabilities as the conformity score, and later work applies it to LLM judges~\citep{jungTrustEscalateLLM2025} and code generation~\citep{xuUncertaintyQuantificationLLMbased2026}. Logit-Free CP~\citep{suAPIEnoughConformal2024} uses sample frequencies, building a black-box variant. U-LINE~\citep{vashurinUNCERTAINTYLINELengthInvariantEstimation2025} proposes a post-processing for length normalization. 

\subsubsection{Sequence-level Distribution}\label{sec:lens-dis-sequence}

Sequence-level methods read the probability of a full response, and $\mathcal U$ returns one score over the response. 

\paragraph{Closed-form functions $S=\varnothing$.}
The closed-form functions read the sequence probability directly, including the negative log-likelihood (NLL)~\citep{coleSelectivelyAnsweringAmbiguous2023}, its length-normalized form~\citep{malininUncertaintyEstimationAutoregressive2021}, perplexity~\citep{fomichevaUnsupervisedQualityEstimation2020}, and the exponential of the length-normalized NLL\@. G-NLL~\citep{aichbergerRethinkingUncertaintyEstimation2026} approximates the NLL by greedy decoding. Two further estimators measure its distance to the uniform distribution, by the Rényi divergence and the Fisher--Rao distance~\citep{darrinRainProofUmbrellaShield2023}.

UProp~\citep{duanUPropInvestigatingUncertainty2025} and SAUP~\citep{zhaoUncertaintyPropagationLLM2025} apply them to multi-agent systems. \citet{zhangFixedBiasesDecoding2026} use them in multimodal models, and Self-Certainty~\citep{kangScalableBestofNSelection2025} applies to chain-of-thought. Consistency~\citep{lyuCalibratingLargeLanguage2025} computes the entropy of the agreement among the samples. Internal Consistency~\citep{xieCalibratingReasoningLanguage2024} and LI~\citep{kimDetectingLLMHallucination2025} apply the Logit Lens~\citep{nostalgebraistInterpretingGPTLogit2020} to the intermediate states $h^{(\ell)}$ with $\rho=I$, reading the distributional lens at $\ell<L$. The former measures the agreement across layers, and the latter compares distributions across inputs with the actual or a null context.

\paragraph{External functions $S=\mathcal U_S$.}
$\mathcal U_S$ can replace the closed-form entropy~\citep{lyuCalibratingLargeLanguage2025} with lexical similarity. Lexical models include the n-gram model in SelfCheckGPT~\citep{manakulSelfCheckGPTZeroResourceBlackBox2023}, the chrF score in Subjective UQ~\citep{wangSubjectiveUncertaintyQuantification2025}, the token-overlap F1 score in Mutual Information~\citep{yadkoriBelieveNotBelieve2024}, and ROUGE-L in CoCoA~\citep{vashurinCoCoAMinimumBayes2025} and SNNE~\citep{nguyenSemanticEntropyBoosting2025}. 

If a semantics model replaces the lexical similarity, the estimator becomes a surrogate for semantics. These methods weight the distribution to approximate the spread over $\mathcal M_R$. SentenceSAR~\citep{duanShiftingAttentionRelevance2024} uses a SentenceTransformer, LARS~\citep{yaldizNotDesignLearn2025} trains an encoder for semantic correctness, and MARS~\citep{bakmanMARSMeaningAwareResponse2024} fine-tunes a BERT-like model to capture meaning. DisAAD~\citep{cuiEstimatingBlackboxLLM2026} reads the logits of a surrogate model to predict the uncertainty of a black-box LLM\@.

\paragraph{Post-hoc wrappers $S=\psi_S$.}
CoRAP~\citep{liQuantifyingUnderstandingUncertainty2026} applies conformal prediction $\psi_S$ to reasoning tasks. DeepConf~\citep{fuDeepThinkConfidence2026} applies $\psi_S$ as a filter on confidence to filter out low-quality responses. 

\subsubsection{Claim-level Distribution}\label{sec:lens-dis-claim}

The sequence distribution is often too coarse for correctness evaluation, and a single token is too fine to carry a meaning. Claim-level methods split the output into claims and operate on the sub-sequence distribution.

Claim-Conditioned Probability (CCP)~\citep{fadeevaFactCheckingOutputLarge2024} uses an NLI model to split the output into claims and aggregates the token-level probabilities within each claim. Conformal LM~\citep{quachConformalLanguageModeling2024} splits the output with a simple sentence splitter and applies conformal prediction $\psi_S$ for a claim-level guarantee.

\begin{takeaway}{Distributional estimators as baselines}
    The distributional lens is the earliest and most studied lens, and its estimators are the usual baselines. Most methods read token probabilities with gray-box access, and black-box methods use sampled text or a proxy model. Most estimators are native. The surrogate estimators select or weight tokens with a semantics model to approximate the spread over meanings.
\end{takeaway}

\subsection{Representational Lens}\label{sec:lens-rep}

The representational lens reads the residual-stream state $O_\text{rep}=h^{(\ell)}$. We organize the methods by the layers they read, summarized in Table~\ref{tab:rep-survey}. Most representational estimators fit an external model $\mathcal U_S$ on $h^{(\ell)}$, and we separate them by the source of the training label. The reference label is the ground truth of the dataset. The LLM label comes from an LLM judge or the model's self-assessment. The proxy label is the score of an estimator at another lens. The choice of supervision makes the methods surrogates for the outer lens.

\begin{table}[!htb]
\centering
\small
\caption{Representational methods survey. \textit{Access}: \accB{} black-box, \accG{} gray-box, \accW{} white-box. \textit{N/S}: Native and surrogate estimators in Section~\ref{sec:lens-axis}, $\circ$ native, $\bullet_{\text{dis}}$ surrogate for the distributional lens, $\bullet_{\text{sem}}$ surrogate for the semantical lens. $Z$: The variation axis in Section~\ref{sec:variation-axis}. $S$: The auxiliary structure in Section~\ref{sec:auxiliary-axis}. \textit{Target} in Section~\ref{sec:eval-target}: \colorbox{navybg}{\cCOR} correctness, \colorbox{tealbg}{\cGRD} grounding, \colorbox{yellowbg}{\cUNS} underspecification. \textit{Eval. Metrics} in Section~\ref{sec:eval-metrics}: \irnk ranking, \ical calibration, \iscr scoring, \idec decision, \ireg regression.}
\label{tab:rep-survey}
\begin{tabular}{@{} l p{6.3cm} c >{\scriptsize}c >{\footnotesize}c >{\footnotesize}c c l @{}}
\toprule
\textsc{Layer} & \textsc{Method} & \textsc{Access} & \textsc{N/S} & $Z$ & $S$ & \textsc{Target} & \textsc{Eval} \\
\midrule
\multirow{13}{*}{\textbf{Fixed}} & Rel. Mahalanobis Dist.~\citep{renOutofDistributionDetectionSelective2023} & \accW & $\circ$ & $\varnothing$ & $\varnothing$ & \colorbox{navybg}{\cCOR} & \irnk \\
 & U-Space~\citep{braunUSpaceUncoveringWhen2026} & \accW & $\circ$ & $\varnothing$ & $\rho_S$ & \colorbox{navybg}{\cCOR} & \irnk \\
 & INSIDE~\citep{chenLLMsInternalStates2024} & \accW & $\circ$ & $y$ & $\varnothing$ & \colorbox{navybg}{\cCOR} & \irnk \\
 & SemGrad~\citep{liGradientsRespectSemantics2026} & \accW & $\bullet_{\text{dis}}$ & $x'$ & $\mathcal U_S$ & \colorbox{navybg}{\cCOR} & \irnk \\
 & P(IK)~\citep{kadavathLanguageModelsMostly2022} & \accW & $\bullet_{\text{sem}}$ & $\varnothing$ & $\mathcal U_S$ & \colorbox{navybg}{\cCOR} & \ical \\
 & Probe Reasoning~\citep{zhangReasoningModelsKnow2025} & \accW & $\bullet_{\text{sem}}$ & $\varnothing$ & $\mathcal U_S$ & \colorbox{navybg}{\cCOR} & \irnk\ical \\
 & LLM Internal States~\citep{jiLLMInternalStates2024} & \accW & $\bullet_{\text{sem}}$ & $\varnothing$ & $\mathcal U_S$ & \colorbox{navybg}{\cCOR} & \idec \\
 & LAFaCT~\citep{wangLAFaCTAttributionbasedLocalization2026} & \accW & $\bullet_{\text{sem}}$ & $\varnothing$ & $\mathcal U_S$ & \colorbox{navybg}{\cCOR} & \irnk \\
 & KEEN~\citep{gottesmanEstimatingKnowledgeLarge2024} & \accW & $\bullet_{\text{sem}}$ & $\varnothing$ & $\mathcal U_S$ & \colorbox{navybg}{\cCOR} & \ireg \\
 & HALLUCANA~\citep{liHALLUCANAFixingLLM2025} & \accW & $\bullet_{\text{sem}}$ & $\varnothing$ & $\mathcal U_S$ & \colorbox{navybg}{\cCOR}\colorbox{yellowbg}{\cUNS} & \irnk\ireg \\
 & Linguistic Calibration~\citep{mielkeReducingConversationalAgents2022} & \accW & $\bullet_{\text{sem}}$ & $\varnothing$ & $\mathcal U_S$ & \colorbox{navybg}{\cCOR} & \ical\iscr \\
 & CCPS~\citep{khanmohammadiCalibratingLLMConfidence2025} & \accW & $\bullet_{\text{dis}}\bullet_{\text{sem}}$ & $\varnothing$ & $\mathcal U_S$ & \colorbox{navybg}{\cCOR} & \irnk\ical \\
 & HaRP~\citep{huHARPHallucinationDetection2026} & \accW & $\bullet_{\text{sem}}$ & $\varnothing$ & $\rho_S$, $\mathcal U_S$ & \colorbox{navybg}{\cCOR} & \irnk \\
\midrule
\multirow{22}{*}{\textbf{Selected}} & Geometric Deviation~\citep{duGeometricDeviationUnsupervised2026} & \accW & $\circ$ & $\varnothing$ & $\varnothing$ & \colorbox{yellowbg}{\cUNS} & \irnk\idec \\
 & Feature-Gaps~\citep{bakmanUncertaintyFeatureGaps2025} & \accW & $\circ$ & $\varnothing$ & $\mathcal U_S$ & \colorbox{navybg}{\cCOR} & \irnk \\
 & CCS~\citep{burnsDiscoveringLatentKnowledge2023} & \accW & $\circ$ & $\varnothing$ & $\mathcal U_S$ & \colorbox{navybg}{\cCOR} & \idec \\
 & Supervised UE~\citep{liuUncertaintyEstimationQuantification2024} & \accW & $\bullet_{\text{dis}}$ & $\varnothing$ & $\mathcal U_S$ & \colorbox{navybg}{\cCOR} & \irnk\ical\iscr \\
 & Distinguish Known~\citep{ahdritzDistinguishingKnowableUnknowable2024} & \accW & $\bullet_{\text{dis}}$ & $\varnothing$ & $\mathcal U_S$ & \colorbox{navybg}{\cCOR} & \irnk \\
 & Persona Probe~\citep{joshiPersonasWayModel2024} & \accW & $\bullet_{\text{sem}}$ & $\varnothing$ & $\mathcal U_S$ & \colorbox{navybg}{\cCOR} & \idec \\
 & CoT Success Predictor~\citep{afzalKnowingSayingLLM2025} & \accW & $\bullet_{\text{sem}}$ & $\varnothing$ & $\mathcal U_S$ & \colorbox{navybg}{\cCOR} & \idec \\
 & SAPLMA~\citep{azariaInternalStateLLM2023} & \accW & $\bullet_{\text{sem}}$ & $\varnothing$ & $\mathcal U_S$ & \colorbox{navybg}{\cCOR} & \irnk\idec \\
 & MHAD~\citep{zhangDetectingHallucinationLarge2025} & \accW & $\bullet_{\text{sem}}$ & $\varnothing$ & $\mathcal U_S$ & \colorbox{navybg}{\cCOR} & \irnk \\
 & Intrinsic Encoding~\citep{luoTwoPathwaysTruthfulness2026} & \accW & $\bullet_{\text{sem}}$ & $\varnothing$ & $\mathcal U_S$ & \colorbox{navybg}{\cCOR} & \irnk \\
 & Factuality Probe~\citep{hanSimpleFactualityProbes2025} & \accW & $\bullet_{\text{sem}}$ & $\varnothing$ & $\mathcal U_S$ & \colorbox{navybg}{\cCOR} & \irnk \\
 & HaloScope~\citep{duHaloScopeHarnessingUnlabeled2024} & \accW & $\bullet_{\text{sem}}$ & $\varnothing$ & $\mathcal U_S$ & \colorbox{navybg}{\cCOR} & \irnk \\
 & Steer Latents~\citep{parkSteerLLMLatents2025} & \accW & $\bullet_{\text{sem}}$ & $\varnothing$ & $\mathcal U_S$ & \colorbox{navybg}{\cCOR} & \irnk \\
 & SATMD~\citep{vazhentsevTokenLevelDensityBasedUncertainty2025} & \accW & $\bullet_{\text{sem}}$ & $\varnothing$ & $\mathcal U_S$ & \colorbox{navybg}{\cCOR} & \irnk \\
 & SEP~\citep{hanSemanticEntropyProbes2024} & \accW & $\bullet_{\text{sem}}$ & $\varnothing$ & $\mathcal U_S$ & \colorbox{navybg}{\cCOR} & \irnk \\
 & Sparse SEP~\citep{hanSemanticEntropyNeurons2024} & \accW & $\bullet_{\text{sem}}$ & $\varnothing$ & $\mathcal U_S$ & \colorbox{navybg}{\cCOR} & \irnk \\
 & IRIS~\citep{sreyUnsupervisedHallucinationDetection2025} & \accW & $\bullet_{\text{sem}}$ & $\varnothing$ & $\mathcal U_S$ & \colorbox{navybg}{\cCOR} & \idec \\
 & Calibrate Verbal Unc.~\citep{jiCalibratingVerbalUncertainty2025} & \accW & $\bullet_{\text{sem}}$ & $\varnothing$ & $\mathcal U_S$ & \colorbox{navybg}{\cCOR} & \irnk \\
 & PEP~\citep{mrykhinPromptEmbeddingProbes2026} & \accW & $\bullet_{\text{dis}}\bullet_{\text{sem}}$ & $\varnothing$ & $\mathcal U_S$ & \colorbox{navybg}{\cCOR} & \irnk \\
 & Knowledge Awareness~\citep{ferrandoKnowThisEntity2025} & \accW & $\bullet_{\text{sem}}$ & $\varnothing$ & $\rho_S$, $\mathcal U_S$ & \colorbox{navybg}{\cCOR} & \irnk \\
 & HalluSAE~\citep{chenHalluSAEDetectingHallucinations2026} & \accW & $\bullet_{\text{sem}}$ & $\varnothing$ & $\rho_S$, $\mathcal U_S$ & \colorbox{navybg}{\cCOR} & \irnk\idec \\
 & RAGLens~\citep{xiongFaithfulRetrievalAugmentedGeneration2026} & \accW & $\bullet_{\text{sem}}$ & $\varnothing$ & $\rho_S$, $\mathcal U_S$ & \colorbox{tealbg}{\cGRD} & \irnk\idec \\
\midrule
\multirow{6}{*}{\textbf{All}} & Chain-of-Embedding~\citep{wangLatentSpaceChainofEmbedding2025} & \accW & $\circ$ & $\varnothing$ & $\varnothing$ & \colorbox{navybg}{\cCOR} & \irnk \\
 & Layer Ens. LR~\citep{dakhmoucheCanLinearProbes2025} & \accW & $\bullet_{\text{sem}}$ & $\varnothing$ & $\mathcal U_S$ & \colorbox{navybg}{\cCOR} & \irnk\ical \\
 & CLAP~\citep{sureshCrossLayerAttentionProbing2025} & \accW & $\bullet_{\text{sem}}$ & $\varnothing$ & $\mathcal U_S$ & \colorbox{navybg}{\cCOR} & \irnk \\
 & GHOST~\citep{maoDigitalDunningKrugerEffect2026} & \accW & $\bullet_{\text{sem}}$ & $\varnothing$ & $\mathcal U_S$ & \colorbox{navybg}{\cCOR} & \irnk \\
 & ReProbe~\citep{niReProbeEfficientTestTime2026} & \accW & $\bullet_{\text{sem}}$ & $\varnothing$ & $\mathcal U_S$ & \colorbox{navybg}{\cCOR} & \irnk \\
 & SIVR~\citep{sreyLearningUncertaintySequential2026} & \accW & $\bullet_{\text{dis}}\bullet_{\text{sem}}$ & $\varnothing$ & $\mathcal U_S$ & \colorbox{navybg}{\cCOR} & \irnk \\
\bottomrule
\end{tabular}
\end{table}

\subsubsection{Fixed-layer Representation}\label{sec:lens-rep-fixed}

The simplest reading point is a fixed layer. Relative Mahalanobis distance~\citep{renOutofDistributionDetectionSelective2023} measures the distance between $h^{(\ell)}$ and a background distribution. U-Space~\citep{braunUSpaceUncoveringWhen2026} reads late-layer $h^{(\ell)}$ through Jacobian-based $\rho_S$ onto four uncertainty directions, and combines it with \emph{cross-lens} mean token entropy. INSIDE~\citep{chenLLMsInternalStates2024} reads $h^{(\ell)}$ at a middle layer for responses sampled at $Z=y$ and computes the eigenvalues of their covariance matrix. SemGrad~\citep{liGradientsRespectSemantics2026} reads late-layer $h^{(\ell)}$ at the most stable token position under paraphrases and computes the gradient of the distributional log-likelihood.

The other fixed-layer methods train a \emph{probe} $\mathcal U_S$, usually a linear model or a shallow multi-layer perceptron, to predict a label from $h^{(\ell)}$. They differ mainly in where the label comes from. A probe trained on meaning labels is a surrogate for the semantics, and one trained on lexical labels is a surrogate for the distribution.

\paragraph{LLM supervision.}
The supervision can come from the model itself. In P(IK)~\citep{kadavathLanguageModelsMostly2022}, a value head predicts the probability that the model knows the answer. The label is derived from repeated sampling, and a question counts as known when the sampled answers are correct.

The label can also be assigned by an external LLM judge. This is used in intermediate reasoning steps with no reference label. Probe Reasoning~\citep{zhangReasoningModelsKnow2025} fits $\mathcal U_S$  to predict whether each step is correct. 

\paragraph{Reference supervision.} 
The most common supervision signal is the ground truth. LLM Internal States~\citep{jiLLMInternalStates2024} fits $\mathcal U_S$ on the hidden state directly. LAFaCT~\citep{wangLAFaCTAttributionbasedLocalization2026} localizes the factual tokens and trains $\mathcal U_S$ on their $h^{(\ell)}$. KEEN~\citep{gottesmanEstimatingKnowledgeLarge2024} and HALLUCANA~\citep{liHALLUCANAFixingLLM2025} regress against a continuous FActScore~\citep{minFActScoreFinegrainedAtomic2023} instead of a binary correctness label. CCPS~\citep{khanmohammadiCalibratingLLMConfidence2025} pairs each state with an adversarially perturbed one for a robust probe. Linguistic Calibration~\citep{mielkeReducingConversationalAgents2022} fine-tunes the LLM to verbalize a confidence aligned with the representational $\mathcal U_S$ output, and the verbalized semantics is a \emph{cross-lens} signal.
HaRP~\citep{huHARPHallucinationDetection2026} applies a readout map $\rho_S$ that projects $h^{(\ell)}$ onto the weakly expressed component for prediction.

\subsubsection{Selected-layer Representation}\label{sec:lens-rep-selected}

A large family sweeps over the layers and keeps the one with the best performance. Geometric Deviation~\citep{duGeometricDeviationUnsupervised2026} is a native estimator, reading the distance from $h^{(\ell)}$ at the selected layer to the centroid of representations of known questions as answerability of the question.

\paragraph{LLM supervision.}
The supervision can come from the model itself. Feature-Gaps~\citep{bakmanUncertaintyFeatureGaps2025} compares the representation between the given input and a specified input and reads uncertainty from the gap. CCS~\citep{burnsDiscoveringLatentKnowledge2023} fits a probe against the logistic consistency of the model's internal states.

Supervised UE~\citep{liuUncertaintyEstimationQuantification2024} and Distinguish Known~\citep{ahdritzDistinguishingKnowableUnknowable2024} fit a probe on the states of an auxiliary model to recover the output distribution of the target model. Their estimand is defined over $V$. Persona Probe~\citep{joshiPersonasWayModel2024} predicts truthfulness judged by LLM\@. CoT Success Predictor~\citep{afzalKnowingSayingLLM2025} predicts whether the reasoning will succeed with LLM judge.

\paragraph{Reference supervision.}
The largest group fits $\mathcal U_S$ against a reference meaning-level correctness label, estimating semantical uncertainty. SAPLMA~\citep{azariaInternalStateLLM2023} first uses a multi-layer perceptron probe as $\mathcal U_S$. Later work changes the part of the state used by the probe. MHAD~\citep{zhangDetectingHallucinationLarge2025} restricts the probe to a selected subset of neurons. PEP~\citep{mrykhinPromptEmbeddingProbes2026} reads the prompt states alone. Intrinsic Encoding~\citep{luoTwoPathwaysTruthfulness2026} fits separate probes on the prompt and the response and combines their outputs. Factuality Probe~\citep{hanSimpleFactualityProbes2025} applies the probe to retrieval-augmented generation. HaloScope~\citep{duHaloScopeHarnessingUnlabeled2024} and Steer Latents~\citep{parkSteerLLMLatents2025} first fit the probe against pseudo-labels and then align it with the reference labels.

Other methods transform $h^{(\ell)}$ before fitting $\mathcal U_S$. SATMD~\citep{vazhentsevTokenLevelDensityBasedUncertainty2025} computes token-level Mahalanobis distances of selected layers. Knowledge Awareness~\citep{ferrandoKnowThisEntity2025}, HalluSAE~\citep{chenHalluSAEDetectingHallucinations2026}, and RAGLens~\citep{xiongFaithfulRetrievalAugmentedGeneration2026} read the features of a sparse autoencoder (SAE)~\citep{hubenSparseAutoencodersFind2024}. The SAE applies a readout map $\rho_S$ to the $h^{(\ell)}$.

\paragraph{Proxy supervision.}
The label can also be another estimator's output, usually a \emph{cross-lens} signal. Semantic Entropy Probes (SEP)~\citep{hanSemanticEntropyProbes2024} and Sparse SEP~\citep{hanSemanticEntropyNeurons2024} fit 
$\mathcal U_S$ against semantic entropy~\citep{kuhnSemanticUncertaintyLinguistic2023}. IRIS~\citep{sreyUnsupervisedHallucinationDetection2025} uses verbalized uncertainty. Calibrate Verbal Uncertainty~\citep{jiCalibratingVerbalUncertainty2025} combines semantic entropy and the verbalized statement into one label.

\subsubsection{All-layer Representation}\label{sec:lens-rep-all}

The first two groups commit to one single reading point. The third group reads the field $\{h^{(\ell)}\}_{\ell=1}^{L}$ and reads the evolution of the representation through the network. Chain-of-Embedding~\citep{wangLatentSpaceChainofEmbedding2025} describes this trajectory as the magnitude and the angle between adjacent layers. 

\paragraph{LLM supervision.}
ReProbe~\citep{niReProbeEfficientTestTime2026} uses an LLM judge to label reasoning step correctness.

\paragraph{Reference supervision.}
Stacked over $L$ layers, the observation is a high-dimensional and noisy $(L, d)$ array, so the methods compress it into features before fitting $\mathcal U_S$ against a ground-truth label. Layer Ensembled Logistic Regression~\citep{dakhmoucheCanLinearProbes2025} uses the layer-wise increment of the residual stream. Cross-Layer Attention Probing (CLAP)~\citep{sureshCrossLayerAttentionProbing2025} projects each layer's state into a low-dimensional space. SIVR~\citep{sreyLearningUncertaintySequential2026} summarizes the trajectory by layer-wise covariance and circular covariance statistics. GHOST~\citep{maoDigitalDunningKrugerEffect2026} records the cosine similarity and the distance between adjacent layers. The form of $\mathcal U_S$ ranges from logistic regression to random forests and transformers.

\begin{takeaway}{Representational estimators are mostly surrogates}
    All representational methods need white-box access, and most read a single forward pass and target correctness. The native estimators measure spread in hidden state space by a closed-form function or an unsupervised probe. The surrogate estimators fit a model against a semantical or distributional label.
\end{takeaway}

\subsection{Structural Lens}\label{sec:lens-str}

The structural lens reads the component writes $O_\text{str}=\Delta h_c^{(\ell)}$ and their intermediate states, the attention weights $a_c^{(\ell)}$ of a head and the activation $m^{(\ell)}$ of the FFN\@. The write shows what the component adds to the residual stream, and the intermediate state records how the component computes that write. Table~\ref{tab:str-survey} organizes the methods by the component they read. Most structural estimators fit $\mathcal U_S$ on the component writes against a label of correctness or grounding. 

\begin{table}[!htb]
\centering
\small
\caption{Structural methods survey. \textit{Access}: \accB{} black-box, \accG{} gray-box, \accW{} white-box. \textit{N/S}: Native and surrogate estimators in Section~\ref{sec:lens-axis}, $\circ$ native, $\bullet_{\text{dis}}$ surrogate for the distributional lens, $\bullet_{\text{sem}}$ surrogate for the semantical lens. $Z$: The variation axis in Section~\ref{sec:variation-axis}. $S$: The auxiliary structure in Section~\ref{sec:auxiliary-axis}. \textit{Target} in Section~\ref{sec:eval-target}: \colorbox{navybg}{\cCOR} correctness, \colorbox{tealbg}{\cGRD} grounding. \textit{Eval. Metrics} in Section~\ref{sec:eval-metrics}: \irnk ranking, \ical calibration, \idec decision, \ireg regression.}
\label{tab:str-survey}
\begin{tabular}{@{} l p{7cm} c >{\scriptsize}c >{\footnotesize}c >{\footnotesize}c c l @{}}
\toprule
\textsc{Comp.} & \textsc{Method} & \textsc{Access} & \textsc{N/S} & $Z$ & $S$ & \textsc{Target} & \textsc{Eval} \\
\midrule
\multirow{20}{*}{\textbf{MHA}} & LLM Check~\citep{sriramananLLMCheckInvestigatingDetection2024} & \accW & $\circ$ & $\varnothing$ & $\mathcal U_S$ & \colorbox{navybg}{\cCOR}\colorbox{tealbg}{\cGRD} & \irnk\idec \\
 & TOHA~\citep{bazarovaHallucinationDetectionLLMs2026} & \accW & $\circ$ & $\varnothing$ & $\mathcal U_S$ & \colorbox{tealbg}{\cGRD} & \irnk \\
 & RFS Guard~\citep{liuRFSGuardDetectingReasoning2026} & \accW & $\circ$ & $\varnothing$ & $\mathcal U_S$ & \colorbox{navybg}{\cCOR} & \irnk\idec \\
 & Attention Head Entropy~\citep{ostmeierGradientStableAttentionHeads2026} & \accW & $\bullet_{\text{dis}}$ & $\varnothing$ & $\mathcal U_S$ & \colorbox{navybg}{\cCOR} & \irnk\ical \\
 & SAT Probe~\citep{yuksekgonulAttentionSatisfiesConstraintSatisfaction2024} & \accW & $\bullet_{\text{dis}}$ & $\varnothing$ & $\mathcal U_S$ & \colorbox{navybg}{\cCOR} & \irnk \\
 & UQAC~\citep{liLanguageModelUncertainty2025} & \accW & $\bullet_{\text{dis}}$ & $\varnothing$ & $\mathcal U_S$ & \colorbox{navybg}{\cCOR} & \ical\irnk \\
 & AGSER~\citep{liuAttentionguidedSelfreflectionZeroshot2025} & \accW & $\bullet_{\text{dis}}$ & $x'$ & $Z_S$ & \colorbox{navybg}{\cCOR} & \irnk \\
 & Attn. Divergence Signals~\citep{vandijkDetectingHallucinationsLarge2026} & \accW & $\bullet_{\text{sem}}$ & $\varnothing$ & $\mathcal U_S$ & \colorbox{navybg}{\cCOR} & \irnk\ical \\
 & LapEigvals~\citep{binkowskiHallucinationDetectionLLMs2025} & \accW & $\bullet_{\text{sem}}$ & $\varnothing$ & $\mathcal U_S$ & \colorbox{navybg}{\cCOR} & \irnk \\
 & Uncertainty Head~\citep{shelmanovHeadPredictHead2025} & \accW & $\bullet_{\text{sem}}$ & $\varnothing$ & $\mathcal U_S$ & \colorbox{navybg}{\cCOR} & \irnk \\
 & LaaB~\citep{miLogicalConsistencyBridge2026} & \accW & $\bullet_{\text{sem}}$ & $\varnothing$ & $\mathcal U_S$ & \colorbox{navybg}{\cCOR} & \idec \\
 & Focus~\citep{zhangEnhancingUncertaintyBasedHallucination2023} & \accW & $\bullet_{\text{sem}}$ & $\varnothing$ & $\mathcal U_S$ & \colorbox{navybg}{\cCOR} & \ical\ireg \\
 & Contextualized Seq. Likelihood~\citep{linContextualizedSequenceLikelihood2024} & \accW & $\bullet_{\text{sem}}$ & $\varnothing$ & $\mathcal U_S$ & \colorbox{navybg}{\cCOR} & \irnk \\
 & Lookback Ratio~\citep{chuangLookbackLensDetecting2024} & \accW & $\bullet_{\text{sem}}$ & $\varnothing$ & $\mathcal U_S$ & \colorbox{tealbg}{\cGRD} & \irnk \\
 & HalluZig~\citep{samagaHalluZigHallucinationDetection2026} & \accW & $\bullet_{\text{sem}}$ & $\varnothing$ & $\mathcal U_S$ & \colorbox{navybg}{\cCOR}\colorbox{tealbg}{\cGRD} & \irnk\idec \\
 & CHARM~\citep{frascaNeuralMessagePassingAttention2026} & \accW & $\bullet_{\text{sem}}$ & $\varnothing$ & $\mathcal U_S$ & \colorbox{navybg}{\cCOR}\colorbox{tealbg}{\cGRD} & \irnk \\
 & RAUQ~\citep{vazhentsevEfficientHallucinationDetection2026} & \accW & $\bullet_{\text{sem}}$ & $\varnothing$ & $\mathcal U_S$ & \colorbox{navybg}{\cCOR}\colorbox{tealbg}{\cGRD} & \irnk \\
 & TAD~\citep{vazhentsevUnconditionalTruthfulnessLearning2025} & \accW & $\bullet_{\text{sem}}$ & $\varnothing$ & $\mathcal U_S$ & \colorbox{navybg}{\cCOR}\colorbox{tealbg}{\cGRD} & \irnk \\
 & Information Flow~\citep{xuInformationFlowReveals2026} & \accW & $\bullet_{\text{sem}}$ & $\varnothing$ & $\mathcal U_S$ & \colorbox{tealbg}{\cGRD} & \irnk \\
 & ITI~\citep{liInferenceTimeInterventionEliciting2023} & \accW & $\bullet_{\text{sem}}$ & $\varnothing$ & $\rho_S$, $\mathcal U_S$ & \colorbox{navybg}{\cCOR} & \idec \\
\midrule
\multirow{4}{*}{\textbf{FFN}} & Factoscope~\citep{heLLMFactoscopeUncovering2024} & \accW & $\bullet_{\text{sem}}$ & $\varnothing$ & $\mathcal U_S$ & \colorbox{navybg}{\cCOR} & \idec \\
 & UQ Trajectory~\citep{eusebiReadingCalibratedUncertainty2026} & \accW & $\bullet_{\text{sem}}$ & $\varnothing$ & $\mathcal U_S$ & \colorbox{navybg}{\cCOR} & \irnk \\
 & Grad Detect~\citep{kamatGradDetectGradientBased2026} & \accW & $\bullet_{\text{sem}}$ & $\varnothing$ & $\mathcal U_S$ & \colorbox{navybg}{\cCOR} & \irnk\idec \\
 & CRV~\citep{zhaoVerifyingChainofThoughtReasoning2026} & \accW & $\bullet_{\text{sem}}$ & $\varnothing$ & $\mathcal U_S$ & \colorbox{navybg}{\cCOR} & \irnk \\
\midrule
\multirow{4}{*}{\textbf{Joint}} & ICR Probe~\citep{zhangICRProbeTracking2025} & \accW & $\bullet_{\text{sem}}$ & $\varnothing$ & $\mathcal U_S$ & \colorbox{navybg}{\cCOR} & \irnk \\
 & TPA~\citep{luTPANextToken2026} & \accW & $\bullet_{\text{sem}}$ & $\varnothing$ & $\mathcal U_S$ & \colorbox{tealbg}{\cGRD} & \irnk \\
 & ReDeEP~\citep{sunReDeEPDetectingHallucination2025} & \accW & $\bullet_{\text{sem}}$ & $\varnothing$ & $\mathcal U_S$ & \colorbox{tealbg}{\cGRD} & \irnk \\
 & Trace Profiles~\citep{grunefeldTracingUncertaintyLanguage2026a} & \accW & $\bullet_{\text{sem}}$ & $\varnothing$ & $\mathcal U_S$ & \colorbox{navybg}{\cCOR} & \irnk \\
\bottomrule
\end{tabular}
\end{table}

\subsubsection{MHA Observation}\label{sec:lens-str-mha}

Most MHA methods read the attention weights $a_c^{(\ell)}$, or the attention pattern $A_c^{(\ell)}$ across positions.

\paragraph{MHA pattern.}

Three native estimators measure the spread of the attention pattern. LLM-Check~\citep{sriramananLLMCheckInvestigatingDetection2024} computes the log-determinant of the attention kernel diagonals over all heads of a selected layer. TOHA~\citep{bazarovaHallucinationDetectionLLMs2026} casts the pattern as a graph and reads the topological divergence between the prompt and response graphs. RFS Guard~\citep{liuRFSGuardDetectingReasoning2026} pools the pattern of the reasoning relevant heads selected by \emph{cross-lens} representational signal. They all use a closed-form statistic of the pattern. $S=\mathcal U_S$ records that a validation step chooses which heads to read.

Attention Head Entropy~\citep{ostmeierGradientStableAttentionHeads2026} and SAT Probe~\citep{yuksekgonulAttentionSatisfiesConstraintSatisfaction2024} fit a logistic regression on the per-head entropy and the attention weights, respectively. The training label is lexical correctness on most of their datasets, so both are surrogates for the distributional lens. UQAC~\citep{liLanguageModelUncertainty2025} back-traces an attention chain and aggregates their probabilities. It selects heads by attention entropy and filters them by \emph{cross-lens} hidden-state similarity. AGSER~\citep{liuAttentionguidedSelfreflectionZeroshot2025} uses an attention saliency map to read the importance of each part of the query.

A larger group targets meaning-level correctness. Internal Attention Divergence Signals~\citep{vandijkDetectingHallucinationsLarge2026} fit $\mathcal U_S$ on the Kullback--Leibler divergence between attention patterns. LapEigvals~\citep{binkowskiHallucinationDetectionLLMs2025} applies it to the top-$k$ Laplacian eigenvalues of the attention graph. Uncertainty Head~\citep{shelmanovHeadPredictHead2025} trains a transformer on the attention features and the \emph{cross-lens} top-$k$ logits. LaaB~\citep{miLogicalConsistencyBridge2026} aligns the model's self-assessment with \emph{cross-lens} signals, like lookback ratio, hidden state, and logit-lens distribution. Focus~\citep{zhangEnhancingUncertaintyBasedHallucination2023} and Contextualized Sequence Likelihood~\citep{linContextualizedSequenceLikelihood2024} are \emph{cross-lens} methods that use the attention pattern to reweight the token probabilities.

The following methods target grounding. Lookback Ratio~\citep{chuangLookbackLensDetecting2024} fits a probe on the ratio of attention weights on the context versus generated tokens. HalluZig~\citep{samagaHalluZigHallucinationDetection2026} extends the attention graph of TOHA~\citep{bazarovaHallucinationDetectionLLMs2026} to cross layer patterns, and CHARM~\citep{frascaNeuralMessagePassingAttention2026} trains a message-passing network on it. RAUQ~\citep{vazhentsevEfficientHallucinationDetection2026} and TAD~\citep{vazhentsevUnconditionalTruthfulnessLearning2025} are \emph{cross-lens}, combining the attention pattern with the token probabilities. RAUQ uses attention-weighted aggregation over selected heads, and TAD uses a learnt regression.

\paragraph{MHA update.}
Information Flow~\citep{xuInformationFlowReveals2026} decomposes the MHA write into the contributions of each token position and traces them across layers. Inference-Time Intervention (ITI)~\citep{liInferenceTimeInterventionEliciting2023} fits a probe on each head's output and steers the outputs along the probe directions at inference. 

\subsubsection{FFN Observation}\label{sec:lens-str-ffn}

Factoscope~\citep{heLLMFactoscopeUncovering2024} fits $\mathcal U_S$ on the activation $m^{(\ell)}$, and \emph{cross-lens} logits, top-$k$ tokens and probabilities, to predict factuality. UQ Trajectory~\citep{eusebiReadingCalibratedUncertainty2026} summarizes FFN update geometry across layers, and fits a sparse probe on the features. Grad Detect~\citep{kamatGradDetectGradientBased2026} back-propagates the gradient to layer-wise FFN update, and compares the gradient with reference gradient. Circuit-based Reasoning Verification (CRV)~\citep{zhaoVerifyingChainofThoughtReasoning2026} analyzes the reasoning circuits through attribution graph, pruned by the \emph{cross-lens} logits, and fits a probe on the features.

\subsubsection{Joint Observation}

The last family reads both components. ICR Probe~\citep{zhangICRProbeTracking2025} reads the discrepancy between the hidden state projection after the MHA write alone and after both writes. \emph{Cross-lens} TPA~\citep{luTPANextToken2026} fits a classifier on the component writes and token distribution. ReDeEP~\citep{sunReDeEPDetectingHallucination2025} proposes an attention-based score and a \emph{cross-lens} distributional score, which are combined for hallucination detection. Trace Profiles~\citep{grunefeldTracingUncertaintyLanguage2026a} trains a classifier on the model reasoning trace features extracted from the gradient and \emph{cross-lens} token probabilities.

\begin{takeaway}{Structural estimators often target grounding}
    All structural methods need white-box access, and some add cross-lens signals. The native estimators measure spread on the attention pattern. The surrogate estimators fit a model against correctness or grounding labels. Structural methods are more often used for grounding prediction than methods at the other lenses.
\end{takeaway}

\paragraph{Summary.} The lens native--surrogate axis, variation axis $Z$, and auxiliary structure axis $S$ all vary across the lens hierarchy. Every semantical estimator and most distributional estimators are native, while 34/41 representational methods and 25/28 structural methods are surrogates for semantics or distributions. The $Z$ and $S$ axes have the same pattern across the lens hierarchy. Outer-lens estimators typically resample outputs, inputs, or parameters, and $S$ is either $Z_S$ or $\psi_S$. Inner-lens estimators usually operate on a single forward pass, with $S$ as $\mathcal U_S$. The evaluation targets and metrics are shown in Section~\ref{sec:evaluation}.

% !TEX root = tmlr.tex
\section{Evaluation}\label{sec:evaluation}

This section shows UQ evaluation along the \emph{target} that the uncertainty score is evaluated against in Section~\ref{sec:eval-target}, the \emph{criterion} that decides whether the target is met in Section~\ref{sec:eval-criteria}, and the \emph{metrics} that quantify how well the score tracks it in Section~\ref{sec:eval-metrics}, followed by the benchmarks used for evaluations in Section~\ref{sec:eval-benchmarks}.

\subsection{Evaluation Target}\label{sec:eval-target}

Since we treat uncertainty as the dispersion of the model's predictive state, it has no ground truth. A UQ method is evaluated as a proxy for a \emph{target}, a property of the inference instance $\langle x, f_\theta, y\rangle$. Section~\ref{sec:survey} tags each surveyed method with one of four evaluation targets: \colorbox{navybg}{\cCOR} correctness, \colorbox{tealbg}{\cGRD} grounding, \colorbox{yellowbg}{\cUNS} underspecification, and \colorbox{skybg}{\cFTH} faithfulness.

\paragraph{Correctness \colorbox{navybg}{\cCOR}} is a property of the output $y$ alone. It indicates whether $y$ matches a ground-truth answer $y^*$ under a criterion $C(y,y^*)$ in Section~\ref{sec:eval-criteria}. An uncertainty score evaluated against correctness is read as the model's \emph{confidence}. Correctness is the default target in the literature and asks: \textit{``How well does the uncertainty estimate tell whether the response is correct?''}

\paragraph{Grounding \colorbox{tealbg}{\cGRD}} is a property of the output $y$, the context in the input $x$, and the model $f_\theta$ for causal reliance. It covers both contextual support, whether $y$ is supported by the context under a criterion $C(y,x)$, and causal reliance, whether $f_\theta$ uses the provided context or its own parametric knowledge to produce $y$. Grounding is used in summarization~\citep{chuangLookbackLensDetecting2024}, translation, and retrieval-augmented generation~\citep{xiongFaithfulRetrievalAugmentedGeneration2026}. Correctness and grounding correspond to the two hallucination types: factuality hallucination, where the output conflicts with known facts, and faithfulness hallucination, where it conflicts with the context~\citep{huangSurveyHallucinationLarge2025, kangUncertaintyQuantificationHallucination2026}. Grounding asks: \textit{``How well does the uncertainty estimate tell whether the response is grounded in the context?''}

\paragraph{Underspecification \colorbox{yellowbg}{\cUNS}} is a property of the input $x$ alone. It specifies whether $x$ admits a valid answer. It covers three cases. Answerability asks whether a valid answer exists given the context~\citep{yinLargeLanguageModels2023}. Ambiguity asks whether the question has more than one valid reading~\citep{coleSelectivelyAnsweringAmbiguous2023}. Novelty asks whether the input is out of distribution. Its label comes from dataset annotation. Underspecification asks: \textit{``How well does the uncertainty estimate tell whether the input is underspecified?''}

\paragraph{Faithfulness \colorbox{skybg}{\cFTH}} is a property of the output $y$ and the model $f_\theta$. It asks whether $y$ reflects the model's own knowledge or reasoning. It compares two uncertainty signals from the same model, such as verbalized confidence against sampling consistency~\citep{yonaCanLargeLanguage2024, xuSaySelfTeachingLLMs2024}, or the final output against the reasoning trace~\citep{daUnderstandingUncertaintyLLM2025, tanneruQuantifyingUncertaintyNatural2024}. Faithfulness asks: \textit{``How well do the model's own uncertainty signals agree with one another?''}

\subsection{Evaluation Criteria}\label{sec:eval-criteria}

An evaluation criterion is a function $C(\cdot,\cdot)$ that decides whether an instance has the target property. For correctness, $C(y, y^*)$ checks the output $y$ against the ground-truth answer $y^*$. For grounding, $C(y, x)$ checks $y$ against the context in $x$, which labels contextual support, and $C(y, (x, \theta))$ checks if the $y$ use the context in $x$ or the paramtric knowledge in $f_\theta$, which labels casual reliance. For underspecification, the label is annotated on $x$, and for faithfulness, the criterion compares two uncertainty signals.

Criteria in the literature measure either token-level overlap between the output and the reference or their similarity in meaning. The distinction between the two approaches is whether $C(\cdot,\cdot)$ reads the lexical form. This is the information discarded by the lens map $g_R$. We divide criteria into distributional and semantical, summarized below and detailed in Appendix~\ref{sec:appendix-criteria}.

\paragraph{Distributional Criteria.}
A distributional criterion measures the lexical overlap between the output and the reference. Common choices are \emph{exact match}, \emph{BLEU}~\citep{papineniBleuMethodAutomatic2002}, and \emph{ROUGE-1/2/L}~\citep{linROUGEPackageAutomatic2004}. If the uncertainty esimator is designed to predict targets defined by a distributional criterion, it is either a native distributional estimator or a surrogate for distribution. By Corollary~\ref{claim:lens-sufficiency}, a semantical estimator cannot fully recover a distributional label, since the lexical form is discarded. 

\paragraph{Semantical Criteria.}
A semantical criterion assesses a meaning-level property. In multiple-choice question answering, \emph{option match} checks the answer against the gold option. In open-ended generation, \emph{NLI entailment}~\citep{williamsBroadCoverageChallengeCorpus2018} and \emph{LLM-as-judge}~\citep{zhengJudgingLLMasaJudgeMTBench2023} judge meaning with a model. \emph{FActScore}~\citep{minFActScoreFinegrainedAtomic2023} and \emph{AlignScore}~\citep{zhaAlignScoreEvaluatingFactual2023} check factual consistency against a knowledge source. \emph{Human annotation}~\citep{wangCLUEConceptLevelUncertainty2024} uses human interpretation of meaning. Each of these criteria acts as a semantic relation $R$. A label defined through $R$ can be estimated with any lens.

\subsection{Evaluation Metrics}\label{sec:eval-metrics}

Once the criterion has labeled each instance, a metric measures how well the uncertainty scores track those labels. For a binary target, we call an instance \emph{positive} when it has the target property, i.e., the output is correct or grounded, or the input is well specified, and \emph{negative} otherwise. A good uncertainty score is low on positive instances and high on negative ones. We group metrics into seven families: ranking \irnk, calibration \ical, scoring \iscr, decision \idec, regression \ireg, coverage \icov, and faithfulness \ifth. The first three are the standard UQ evaluation families, and we refer to \citet{wangAleatoricEpistemicExploring2025} for a detailed discussion. The other four are tied to specific methods, targets, or tasks. We summarize each family below and describe the metrics in Appendix~\ref{sec:appendix-metrics}.

\paragraph{Ranking \irnk.} Ranking metrics measure whether the uncertainty score places positive instances ahead of negative ones. Ranking metrics include the area under the receiver operating characteristic curve (\emph{AUROC}), the area under the precision--recall curve (\emph{AUPRC}), the area under the accuracy--rejection curve (\emph{AUARC}), the area under the risk--coverage curve (\emph{AURC}), and the prediction rejection ratio (\emph{PRR}). They summarize the separation over all decision thresholds. Ranking is often reported as calibration~\citep{coxMappingMeaningAddressing2025, jiCalibratingVerbalUncertainty2025}, but they answer different questions~\citep{liLanguageModelUncertainty2025}. Ranking depends only on the order of the scores, whereas calibration depends on their values.\footnote{A constant uncertainty score equal to the empirical accuracy achieves zero calibration error (ECE) but gives no useful ranking, and a score can rank near-optimally while being miscalibrated~\citep{eusebiReadingCalibratedUncertainty2026}.}

\paragraph{Calibration \ical.} Calibration metrics measure the agreement between the probability and the empirical frequency of positive instances. The expected, maximum, and adaptive calibration errors (\emph{ECE}, \emph{MCE}, \emph{ACE}) summarize how well the scores align with the true probability of correctness across confidence bins.

\paragraph{Scoring \iscr.} Proper scoring rules assess calibration and the concentration of the predicted distribution. The \emph{Brier score} applies to binary labels, and the continuous ranked probability score (\emph{CRPS}) applies to continuous labels.

\paragraph{Decision \idec.} Some papers threshold the uncertainty score for downstream decisions, such as agentic abstention~\citep{luoAgenticAbstentionAgents2026, duanUPropInvestigatingUncertainty2025}, and answer abstention~\citep{fengDontHallucinateAbstain2024}. They report classification metrics such as \emph{accuracy} and \emph{F1 score}. A decision metric is a single operating point on a ranking curve. We keep it as a separate family because it usually serves a downstream action, and the threshold is usually chosen for that action. It is also widely reported in the surveyed literature.

\paragraph{Regression \ireg.} Regression metrics apply when the criterion returns a continuous score, for instance FActScore~\citep{minFActScoreFinegrainedAtomic2023}. They measure the correlation between the uncertainty score and the criterion score, using the Pearson correlation coefficient (\emph{PCC}) or Spearman's rank correlation coefficient (\emph{SCC}). Binarizing the score at a threshold turns the regression metrics into ranking metrics.

\paragraph{Coverage \icov.} Coverage metrics are specific to conformal prediction~\citep{wangConUConformalUncertainty2024, mohriLanguageModelsConformal2024, suAPIEnoughConformal2024}, which outputs a prediction set guaranteed to contain an acceptable response with a chosen probability. \emph{Coverage probability} checks the guarantee holds empirically, and \emph{average prediction set size} measures how efficient the set is.

\paragraph{Faithfulness \ifth.} Faithfulness metrics apply to the faithfulness target. They measure the agreement between two uncertainty signals of the same model, typically one expressed in the output and one derived from the model's computation. Faithfulness metrics include \emph{rank correlation}, the conditional mean faithfulness gap (\emph{cMFG})~\citep{yonaCanLargeLanguage2024, liuReinforcementLearningMetacognitive2026, liuMetaFaithFaithfulNatural2025}, and \emph{CoT intervention}.

\begin{table}[htbp]
\centering
\caption{Evaluation targets, criteria, and metric families. \textit{Criterion}: dis = distributional criterion, sem = semantical criterion. \textit{Metric family}: \irnk ranking, \ical calibration, \iscr scoring, \idec decision, \ireg regression, \icov coverage, \ifth faithfulness. \textit{Ground truth}: Whether the target requires a ground-truth label.}
\label{tab:eval-metrics}
\renewcommand{\arraystretch}{1.3}
\begin{tabular}{l l l c r}
\toprule
\textsc{Target} & \textsc{Criterion} & \textsc{Metric Family} & \textsc{Ground Truth} & \textsc{\# Papers} \\
\midrule
Correctness \colorbox{navybg}{\cCOR}          & dis, sem & \irnk\ical\iscr\idec\ireg\icov & \checkmark & 146 \\
Grounding \colorbox{tealbg}{\cGRD}           & dis, sem & \irnk\idec                     & \checkmark & 11  \\
Underspecification \colorbox{yellowbg}{\cUNS} & sem      & \irnk\idec                     & \checkmark & 3   \\
Faithfulness \colorbox{skybg}{\cFTH}       & sem      & \ifth                          & --         & 6   \\
\bottomrule
\end{tabular}
\end{table}

Table~\ref{tab:eval-metrics} summarizes how different metric families and criteria are used for each target. Among the 159 surveyed papers, 146 target correctness, and 81 of these report ranking metrics only. The grounding target sits almost entirely at the structural lens, and all faithfulness methods read the semantical lens. Corollary~\ref{claim:lens-sufficiency} accounts for this pattern. A correctness label is defined on the output, so any inner lens can be read for it. Grounding methods typically assess causal reliance, which requires attribution to individual component writes that $g_\text{sum}$ discards, although they are evaluated against contextual support labels. Faithfulness focuses on expressed uncertainty in semantical objects. Across all targets, ranking is the most common metric family, followed by calibration, and a growing number of papers evaluate decision metrics on downstream use.

\begin{takeaway}{Target, criterion, and metric form the evaluation protocol}
    A UQ evaluation protocol evaluates the uncertainty score against a target: correctness, grounding, underspecification, or faithfulness. A distributional or semantical criterion decides whether the target is met. A metric in one of seven families measures how well the score tracks the target.
\end{takeaway}

\subsection{Benchmarks}\label{sec:eval-benchmarks}

We close the section with the benchmarks used in the surveyed papers, grouped by input--output format: multiple-choice question answering (MCQA), short-form question answering (SFQA), long-form question answering (LFQA), open-ended generation (OEG), and contextual generation (CG).

We group by format because different formats require different evaluation criteria. MCQA is judged by semantical option match. SFQA is sometimes scored with distributional criteria, while LFQA, OEG, and CG usually use semantical criteria. Benchmarks with gold answers support the correctness target. Benchmarks with context support grounding. Grounding benchmarks include open-book QA in all three question formats and contextual generation tasks such as summarization, translation, and retrieval-augmented generation. A few benchmarks have underspecification labels, marking unanswerable questions or if an agent should ask for clarification. Faithfulness needs no ground truth and can be evaluated on any benchmark.

Table~\ref{tab:eval-benchmarks} lists the benchmarks by format. Within each row, the benchmarks after \colorbox{tealbg}{\cGRD} also support grounding, and those after \colorbox{yellowbg}{\cUNS} support underspecification. Correctness \colorbox{navybg}{\cCOR} and faithfulness \colorbox{skybg}{\cFTH} apply to all benchmarks and are not marked.

\begin{table}[!htb]
\centering
\caption{Benchmarks by input--output format. Multiple-choice question answering (MCQA), short-form question answering (SFQA), long-form question answering (LFQA), open-ended generation (OEG), and contextual generation (CG)\@. \colorbox{tealbg}{\cGRD} marks grounding applicable benchmarks, \colorbox{yellowbg}{\cUNS} marks underspecification applicable benchmarks, \colorbox{navybg}{\cCOR} and \colorbox{skybg}{\cFTH} omitted.}
\label{tab:eval-benchmarks}
\renewcommand{\arraystretch}{1.15}
\small
\begin{tabular}{@{} p{1.6cm} p{\dimexpr\textwidth-1.6cm-2\tabcolsep\relax} @{}}
\toprule
\textsc{Format} & \textsc{Benchmarks} \\
\midrule
MCQA & MMLU(-Pro)~\citep{hendrycksMeasuringMassiveMultitask2021, wangMMLUProMoreRobust2024}, MedQA~\citep{jinWhatDiseaseDoes2021}, MedMCQA~\citep{palMedMCQALargescaleMultiSubject2022}, GPQA~\citep{reinGPQAGraduateLevelGoogleProof2024}, CommonsenseQA~\citep{talmorCommonsenseQAQuestionAnswering2019}, BoolQ~\citep{clarkBoolQExploringSurprising2019}, ARC~\citep{clarkThinkYouHave2018}, HellaSwag~\citep{zellersHellaSwagCanMachine2019}, FEVER~\citep{thorneFEVERLargescaleDataset2018}, Winogrande~\citep{sakaguchiWinoGrandeAdversarialWinograd2019}, PIQA~\citep{biskPIQAReasoningPhysical2019}, GLUE~\citep{wangGLUEMultiTaskBenchmark2018}, BigBench~\citep{srivastavaImitationGameQuantifying2023},\newline \colorbox{tealbg}{\cGRD} SciQ~\citep{welblCrowdsourcingMultipleChoice2017}, CosmosQA~\citep{huangCosmosQAMachine2019}, RACE~\citep{laiRACELargescaleReAding2017}, OpenBookQA~\citep{mihaylovCanSuitArmor2018}, HaluEval~\citep{liHaluEvalLargeScaleHallucination2023} \\
SFQA & TriviaQA~\citep{joshiTriviaQALargeScale2017}, NQ(-open)~\citep{kwiatkowskiNaturalQuestionsBenchmark2019}, PopQA~\citep{mallenWhenNotTrust2023}, BioASQ~\citep{tsatsaronisOverviewBIOASQLargescale2015}, TruthfulQA~\citep{linTruthfulQAMeasuringHow2022}, SimpleQA~\citep{weiMeasuringShortformFactuality2024}, MS MARCO~\citep{bajajMSMARCOHuman2018}, WebQuestions~\citep{berantSemanticParsingFreebase2013}, LAMA~\citep{petroniLanguageModelsKnowledge2019}, PubMedQA~\citep{jinPubMedQADatasetBiomedical2019}, MGSM~\citep{shiLanguageModelsAre2023},\newline \colorbox{tealbg}{\cGRD} CoQA~\citep{reddyCoQAConversationalQuestion2019}, HotpotQA~\citep{yangHotpotQADatasetDiverse2018}, SQuAD~\citep{rajpurkarSQuAD100000Questions2016}, HaluEval~\citep{liHaluEvalLargeScaleHallucination2023},\newline \colorbox{yellowbg}{\cUNS} QuAC~\citep{choiQuACQuestionAnswering2018}, SelfAware~\citep{yinLargeLanguageModels2023} \\
LFQA & ELI5(-Category)~\citep{fanELI5LongForm2019, atakishiyevExplainableArtificialIntelligence2024}, MedLFQA~\citep{jeongOLAPHImprovingFactuality2024},\newline \colorbox{tealbg}{\cGRD} Qasper~\citep{dasigiDatasetInformationSeekingQuestions2021}, NarrativeQA~\citep{kociskyNarrativeQAReadingComprehension2018} \\
OEG & FActScore~\citep{minFActScoreFinegrainedAtomic2023}, WikiBio GPT-3~\citep{manakulSelfCheckGPTZeroResourceBlackBox2023}, Chatbot Arena~\citep{chiangChatbotArenaOpen2024}, FAVA~\citep{mishraFinegrainedHallucinationDetection2024}, HumanEval~\citep{chenEvaluatingLargeLanguage2021}, MBPP~\citep{austinProgramSynthesisLarge2021},\newline \colorbox{tealbg}{\cGRD} MT-Bench~\citep{chenMTBenchMultimodalTime2026} \\
CG & \colorbox{tealbg}{\cGRD} XSum~\citep{narayanDontGiveMe2018}, SamSum~\citep{gliwaSAMSumCorpusHumanannotated2019}, TL;DR~\citep{stiennonLearningSummarizeHuman2020}, SummEval~\citep{fabbriSummEvalReevaluatingSummarization2021}, FRANK~\citep{pagnoniUnderstandingFactualityAbstractive2021}, CNN/DailyMail~\citep{hermannTeachingMachinesRead2015}, WMT~\citep{bojarFindings2017Conference2017, barraultFindings2019Conference2019, barraultFindings2020Conference2020}, HaluEval~\citep{liHaluEvalLargeScaleHallucination2023}, RAGTruth~\citep{niuRAGTruthHallucinationCorpus2024} \\
\bottomrule
\end{tabular}
\end{table}

There are also benchmarks used for chain-of-thought reasoning, like GSM8K~\citep{cobbeTrainingVerifiersSolve2021}, MATH~\citep{hendrycksMeasuringMathematicalProblem2021}, CommonsenseQA~\citep{talmorCommonsenseQAQuestionAnswering2019}, BoolQ~\citep{clarkBoolQExploringSurprising2019}, TheoremQA~\citep{chenTheoremQATheoremdrivenQuestion2023}, and SVAMP~\citep{patelAreNLPModels2021}. Others target a specific domain, like MedLFQA~\citep{jeongOLAPHImprovingFactuality2024}, MedQA~\citep{jinWhatDiseaseDoes2021}, MedMCQA~\citep{palMedMCQALargescaleMultiSubject2022}, and BioASQ~\citep{tsatsaronisOverviewBIOASQLargescale2015}. Some papers construct their own trivia datasets~\citep{heLLMFactoscopeUncovering2024, yuksekgonulAttentionSatisfiesConstraintSatisfaction2024, ferrandoKnowThisEntity2025}.

% !TEX root = tmlr.tex
\section{Discussion}\label{sec:discussion}

\subsection{Open Problems}\label{sec:open-problems}

\paragraph{Evaluation Target for Uncertainty Quantification.} Most UQ methods are designed as \emph{correctness} predictors. Correctness is labeled only at the output by distributional or semantical criteria. It is an adequate target when only the output matters. When the concern is the model's intrinsic uncertainty, new evaluation targets are needed, and Section~\ref{sec:future-directions} discusses them. \citet{beigiRethinkingUncertaintyCritical2024} raise the same concern and call for UQ to go beyond confidence prediction. The other three targets in Section~\ref{sec:eval-target} are relatively under-explored. Grounding covers both contextual support, which can be judged from the output and the context, and causal reliance, whether the model uses the provided context or its own parametric knowledge. Causal reliance requires attribution at the structural lens~\citep{xuInformationFlowReveals2026, sunReDeEPDetectingHallucination2025}, but most grounding methods are evaluated against contextual support labels. Faithfulness and underspecification need no correctness label. Faithfulness compares two uncertainty signals of the same model, and underspecification is labeled on the input~\citep{renOutofDistributionDetectionSelective2023}. \citet{kirchhofPositionUncertaintyQuantification2025} name underspecification uncertainty as a target in agentic frameworks, and \citet{xiaSurveyUncertaintyEstimation2025} call for benchmarks with ambiguous inputs for evaluation.

\paragraph{LLM Internal Uncertainty.} Since most estimators target output-level correctness, most representational and structural methods fit an external model $\mathcal U_S$ against output-level labels. They are surrogate estimators for the distribution or semantics. The few inner-lens native estimators measure the spread of $O_{\text{rep}}$ or $O_{\text{str}}$ but are still evaluated against the output. This raises the questions of whether native representational and structural estimators are necessary, what target is visible at inner lens and is lost through $g_\text{sum}$ and $g_\text{unemb}$, and how to design them without output-level labels. Section~\ref{sec:future-directions} explores these questions.

\paragraph{Limitations of Four Lens Observation.} The four lenses in Section~\ref{sec:lenses} give a unified view of where uncertainty can be observed in the model's computation. Corollary~\ref{claim:information-loss} shows that every lens map is lossy. This concerns the information available in the observation, not the information usable by an estimator, and how the lost information can be exploited remains open. The experiment in Section~\ref{sec:lenses-poc} measures only the correctness signal that one probe family can use at each lens.

\subsection{Future Directions}\label{sec:future-directions}

\paragraph{Uncertainty Quantification for Agentic Frameworks.} Beyond correctness, uncertainty quantification can serve as a decision signal within agentic frameworks~\citep{luoAgenticAbstentionAgents2026, kirichenkoAbstentionBenchReasoningLLMs2025, davidovKnowingWhenQuit2026}. Uncertainty can guide the decision whether to answer or request clarification~\citep{kirchhofPositionUncertaintyQuantification2025}, whether to rely on retrieved context or parametric knowledge~\citep{sunReDeEPDetectingHallucination2025}, and whether to invoke extended reasoning~\citep{daUnderstandingUncertaintyLLM2025}. Each decision draws on a different lens.  Retrieval routing requires structural attribution, reasoning decisions require hidden-state analysis, while selective answering requires distributional or semantic signals~\citep{fengDontHallucinateAbstain2024, duGeometricDeviationUnsupervised2026}. \citet{shorinwaSurveyUncertaintyQuantification2025} also ask for benchmarks that evaluate UQ for LLMs in multi-episode interaction scenarios. Along this direction, a small but growing line of work examines the uncertainty of agentic systems~\citep{zhaoUncertaintyPropagationLLM2025, duanUPropInvestigatingUncertainty2025, chenEveryResponseCounts2026}. Future work can study how uncertainty propagates through an agentic system and how to design an estimator for each decision.

\paragraph{Native Uncertainty Estimators in the Inner Lenses.} We return to the three questions raised in Section~\ref{sec:open-problems}. On necessity, component attribution depends on $O_{\text{str}}$, so no outer-lens estimator can recover it. For such targets, such as causal reliance in grounding, native inner-lens estimators are necessary. On targets, recent work shows that structural estimators can measure the causal contribution of individual components~\citep{mengLocatingEditingFactual2022}, including whether the output relies on the provided context or on parametric knowledge~\citep{xuInformationFlowReveals2026, sunReDeEPDetectingHallucination2025}, and representational estimators may track whether the reasoning trajectory is faithful to the final answer~\citep{lanhamMeasuringFaithfulnessChainofThought2023}. On design, supervised probes need reference labels~\citep{azariaInternalStateLLM2023} and generalize poorly~\citep{ch-wangAndroidsKnowTheyre2024,sunReDeEPDetectingHallucination2025}. A more promising direction reads the model computation directly, with no labels or training~\citep{bazarovaHallucinationDetectionLLMs2026}. Such estimators can be validated without labels through causal mechanistic validation~\citep{geigerCausalAbstractionsNeural2021, bugaudStatisticalFrameworkMechanistic2026}. If an estimator reads a causally relevant signal, intervening on that signal should change the model's behavior~\citep{kumaranCausalEvidenceThat2026} or its verbalized confidence~\citep{zhaoWiredOverconfidenceMechanistic2026}.

\paragraph{Exploiting the Lens Hierarchy.} The lens hierarchy in Section~\ref{sec:lenses-hierarchy} is an information ordering and is validated empirically through decodability. It should also be exploited directly. \citet{bakmanUncertaintyFeatureGaps2025} derive an upper bound on distributional epistemic uncertainty in terms of hidden-state variation. In Section~\ref{sec:lenses-poc}, the $g_\text{unemb}$ and $g_R$ gaps for correctness prediction depend on the task and the model. It remains unclear when or why they widen, or whether they matter for targets beyond correctness. Readout maps offer another way to exploit the hierarchy. The J-lens~\citep{gurnee2026verbalizable} and R-lens~\citep{blankRlensMakingJlens2026} are readout maps $\rho$ at the distributional lens, and the U-space~\citep{braunUSpaceUncoveringWhen2026} and SAE~\citep{hubenSparseAutoencodersFind2024} are readout maps $\rho$ at the representational lens. A readout map does not change the lens, but it can make the same observation more faithful or easier to interpret~\citep{blankRlensMakingJlens2026}.

% !TEX root = tmlr.tex
\section{Conclusions}\label{sec:conclusion}

Large language models can produce unreliable outputs, and uncertainty quantification can be used to predict the correctness, grounding, faithfulness, and underspecification of their responses given the query. This survey organizes UQ methods through four observation lenses: structural, representational, distributional, and semantical, and characterizes them by the lens they read, the variable they resample, and the auxiliary structure they use. The lens determines the sufficiency of an estimator for a target, and the resampled variable and auxiliary structure determine its efficiency. We also distinguish evaluation targets, criteria, and metrics to clarify different UQ evaluations. Future work can use uncertainty estimates to guide decisions in agentic systems and investigate how uncertainty is expressed within LLMs.

%% ─────────────────────────────────────────────────────────────────────

\subsubsection*{AI Use Statement}
In this work, we used generative AI tools to refine the framework and notation, brainstorm, pre-classify surveyed methods, and design experiments. We did not use them to originate the central research idea, select surveyed papers, state or prove claims, draft the manuscript, create synthetic data, translate, or interpret results. We additionally used them to implement experiments, create or modify schematic figures, and polish wording. All AI-assisted outputs were reviewed by the authors: experiment code was tested for correctness, figures were inspected, surveyed method classifications were decided by the authors, and edited text was critically revised. We take responsibility for the final content of this work, including all text, claims, and artifacts produced with the aid of generative AI\@.

% \subsubsection*{Broader Impact Statement}
% In this optional section, TMLR encourages authors to discuss possible repercussions of their work,
% notably any potential negative impact that a user of this research should be aware of.
% Authors should consult the TMLR Ethics Guidelines available on the TMLR website
% for guidance on how to approach this subject.

% in many journals. This is optional and at the discretion of the authors. Only add
% this information once your submission is accepted and deanonymized.

% Only add this information once your submission is accepted and deanonymized.

\bibliography{tmlr}
\bibliographystyle{tmlr}

\appendix
% !TEX root = tmlr.tex
\section{Appendix}\label{sec:appendix}

\subsection{Literature Selection}\label{sec:appendix-selection}

\paragraph{Round 1: related surveys.} We searched Google Scholar for ``uncertainty survey'' and ``uncertainty review'' and kept the surveys, reviews, and position papers on uncertainty quantification. For each retained work, we also checked the surveys it cites and added those in scope. These works form Table~\ref{tab:survey-comparison}. We do not cover surveys of mechanistic interpretability that do not address uncertainty quantification, such as \citet{barezUnderstandingAIState2026, bereskaMechanisticInterpretabilityAI2024}.

\paragraph{Round 2: methods covered by the surveys.} From the surveys of Round~1, we collected every UQ method applied to LLMs. For each method, we checked the baselines it compares against and added the important baselines that no survey in Round~1 covers.

\paragraph{Round 3: recent conference papers.} We then screened the proceedings of ACL, EMNLP, NAACL, ICLR, ICML, and NeurIPS\@. For each venue, we selected the two most recent editions with proceedings available in August 2026 and kept papers whose title or abstract contains ``uncertainty'' or ``hallucination''. We added the important baselines of each retained paper.

\paragraph{Inclusion criteria.} We include a method if it produces an uncertainty score for an LLM\@. We exclude work that applies an existing UQ score to hallucination detection or mitigation to any other task without proposing a new score. Methods for code generation and multimodal models fall outside our main scope, and we include only some of them. A baseline counts as important when several retained methods compare against it or when it was published at a top conference or journal.

\subsection{Proof-of-Concept Experiment}\label{sec:appendix-poc}

This appendix documents the experiments in Section~\ref{sec:lenses-poc}, covering generation, labeling, feature construction, probe fitting, evaluation, and additional results. 

\subsubsection{Model and Datasets Description}\label{sec:appendix-poc-models-datasets}

\paragraph{Models.} We evaluate three open-weight models from two model families and two post-training regimes. Our primary model is Qwen2.5-7B-Instruct~\citep{qwenQwen25TechnicalReport2025}. We also evaluate Qwen2.5-7B, the pretrained base model, to measure the effect of instruction tuning. Llama-3.1-8B-Instruct~\citep{grattafioriLlama3Herd2024} is included to test generalization across model families.

\paragraph{Datasets.} We use three datasets, all standard benchmarks for correctness prediction and all used extensively in prior UQ works. We draw $N = 3{,}000$ questions per dataset under a fixed sampling seed, held constant across models. TriviaQA~\citep{joshiTriviaQALargeScale2017} contains open-domain question-answering trivia questions paired with gold answers and alias sets. We use the \texttt{rc.nocontext} configuration of \texttt{mandarjoshi/trivia\_qa}, validation split\footnote{\url{https://huggingface.co/datasets/mandarjoshi/trivia\_qa}}. PopQA~\citep{mallenWhenNotTrust2023} is an entity-centric open-domain QA dataset. We use \texttt{akariasai/PopQA}, test split\footnote{\url{https://huggingface.co/datasets/akariasai/PopQA}}. MMLU~\citep{hendrycksMeasuringMassiveMultitask2021} is a multiple-choice benchmark. We use the \texttt{all} configuration of \texttt{cais/mmlu}, test split\footnote{\url{https://huggingface.co/datasets/cais/mmlu}}, output rendered as $\{A, B, C, D\}$.

\paragraph{Inference.} For each question, greedy decoding produces a deterministic answer $a_0$ with $50$ new tokens. For the semantical lens, we additionally draw $k = 20$ stochastic samples using temperature $T = 1.0$ and top-$p = 0.9$. A single teacher-forced forward pass over $[\text{prompt}; a_0]$ provides the hidden states, per-(layer, head) attention weights, and per-token logits used by the other three lenses. Questions are presented using a $5$-shot prompt containing in-domain exemplars, fixed within each dataset. All models are loaded in \texttt{bfloat16}.

\paragraph{Correctness labels.} Correctness is a binary target, with \emph{incorrect} as the positive class. The chance-level AUPRC equals the rate of incorrect answers. For TriviaQA and PopQA, we first compare $a_0$ with the set of gold aliases using normalized string matching, with strings lowercased, articles and punctuation removed. If no match is found, we test bidirectional NLI entailment between $a_0$ and each alias using \texttt{cross-encoder/nli-deberta-v3-large} with threshold $\tau = 0.5$, which applies to fewer than 4\% of questions across all three models. For MMLU, correctness is determined by option letter matching.

\subsubsection{Feature Extraction in Four Lenses}\label{sec:appendix-poc-features}
We extract four strictly nested feature sets, $F_\text{sem} \subset F_\text{dis} \subset F_\text{rep} \subset F_\text{str}$. Each inner set contains every outer set as a prefix, so the difference in predictive performance between a lens and the next outer lens is the marginal contribution of the new features.

\paragraph{Semantical lens.}
The greedy answer and the $k = 20$ sampled answers are grouped into meaning-equivalent clusters using bidirectional NLI entailment~\citep{kuhnSemanticUncertaintyLinguistic2023, farquharDetectingHallucinationsLarge2024}. Shannon entropy over the semantic clusters gives the semantic entropy. We also include the cluster count~\citep{linGeneratingConfidenceUncertainty2024}, the mass of the largest cluster, and the top-$5$ cluster masses, constructing eight semantical features in total. 

\paragraph{Distributional lens.}
Distributional features are computed from the logits of the greedy answer. They capture mean token entropy, mean max softmax probability, top-1 logit margin~\citep{hendrycksBaselineDetectingMisclassified2017}, length-normalized log-probability~\citep{malininUncertaintyEstimationAutoregressive2021}, and variance and skewness of the token-entropy profile, adding ten more features. For MMLU, we renormalize the softmax distribution over the four option-letter token IDs and then compute the features.

\paragraph{Representational lens.}
This lens uses the hidden state of the final answer token at a single readout layer~\citep{azariaInternalStateLLM2023}. The layer is selected separately for each model through a validation sweep and is then frozen. The raw hidden-state vector is compressed using Principal Component Analysis (PCA)~\citep{jolliffePrincipalComponentAnalysis2016}.

\paragraph{Structural lens.}
Structural features include attention entropy~\citep{ostmeierGradientStableAttentionHeads2026}, lookback ratio~\citep{chuangLookbackLensDetecting2024}, and the $\ell_2$ magnitude of each layer's residual update. The resulting feature vector is compressed using PCA\@. The pooling procedure discards head-level detail, so these features give a lower bound on the structural information available.

\subsubsection{Baseline Description}\label{sec:appendix-poc-baseline}
\paragraph{Probes.}
For each lens, we use an $\ell_2$-regularized logistic regression model followed by a one-hidden-layer MLP with 256 units, ReLU activations, and 0.2 dropout for TriviaQA and PopQA\@. For MMLU, we use only the logistic-regression layer. Training uses early stopping on the validation split, and results are averaged over 5 random seeds. The measured increments are marginal \emph{decodable} information, usable by this probe family.

\paragraph{Hyperparameter Selection.}
We jointly select the readout layer, representational PCA dimension, and structural PCA dimension in one sweep, choosing the combination that maximizes the validation performance of DecBase on $F_\text{str}$. This ensures that the same configuration is used consistently across the representational and structural lenses. All three hyperparameters are frozen before test-set evaluation and remain fixed throughout the experiments.
Both Qwen variants use readout layer 17, a representational PCA dimension of 128, and a structural PCA dimension of 128. Llama also selects layer 17, but uses larger representational and structural PCA dimensions of 512 and 256.

\paragraph{Evaluation metrics.} We report AUROC on the full population and AUPRC under confidence cut-off. AUROC is independent of class balance, suitable for comparing lenses on a fixed population but misleading under a confidence cut-off. AUPRC is interpreted against the chance baseline shown in each panel.

\subsubsection{Uncertainty Estimators Description}\label{sec:appendix-poc-estimators}
We implement a minimal variant of RAUQ by selecting, per layer, the head with the highest attention to preceding tokens. Each estimator is detailed in its lens section of the survey in Section~\ref{sec:survey}. SE and INSIDE read the samples with $Z=y$, and TE and RAUQ read the greedy pass with $Z=\varnothing$. 

\subsubsection{Additional Results}\label{sec:appendix-poc-additional-results}

\paragraph{DecBase across Confidence Levels.}
Figure~\ref{fig:correct-confidence-qwen-base} and Figure~\ref{fig:correct-confidence-llama-inst} show the DecBase performance across confidence levels for the other two models. On open-ended QA, all four DecBases fall broadly as the confidence cutoff rises, and the ordering of the lenses holds. On MMLU, the ordering breaks down. The MMLU sanity check behaves the same way across model families and instruction tuning.

\begin{figure}[ht]
    \centering
    \includegraphics[width=\linewidth]{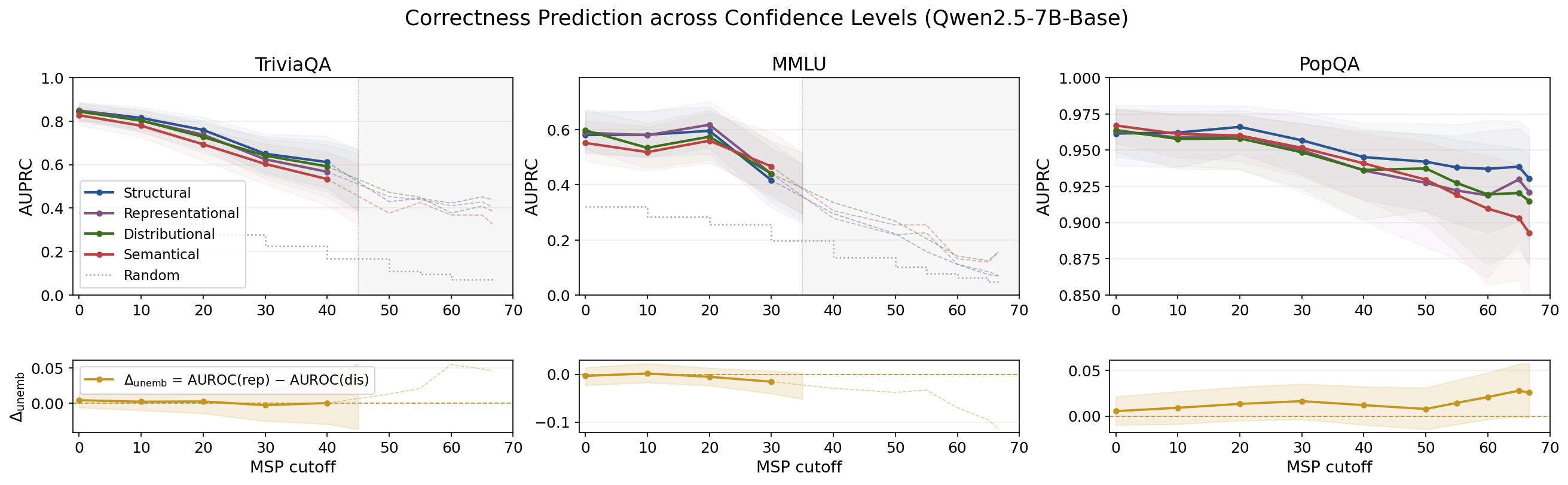}
    \caption{DecBase performance of Qwen-Base across confidence levels. Top: AUPRC as a function of the MSP lower-percentile cutoff, with 95\% confidence intervals. The gray dashed line marks the chance-level baseline, and gray shading marks cutoffs with fewer than 50 incorrect answers. Bottom: marginal gain of $g_\text{unemb}$.}
    \label{fig:correct-confidence-qwen-base}
\end{figure}
\begin{figure}[ht]
    \centering
    \includegraphics[width=\linewidth]{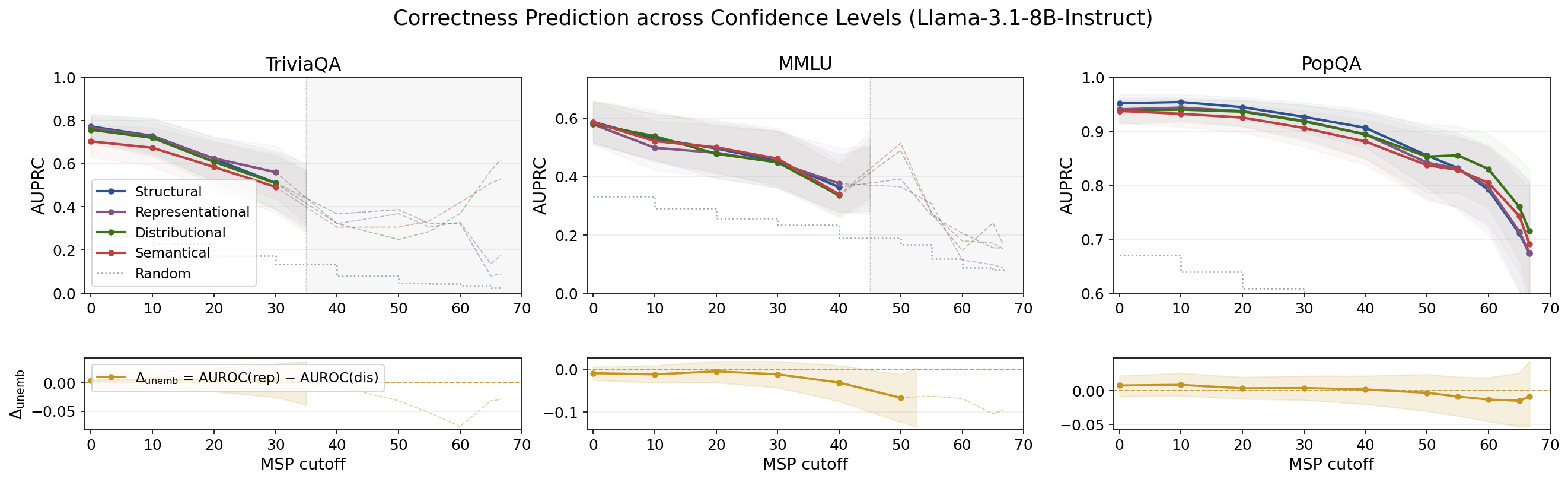}
    \caption{DecBase performance of Llama-Inst across confidence levels. Top: AUPRC as a function of the MSP lower-percentile cutoff, with 95\% confidence intervals. The gray dashed line marks the chance-level baseline, and gray shading marks cutoffs with fewer than 50 incorrect answers. Bottom: marginal gain of $g_\text{unemb}$.}
    \label{fig:correct-confidence-llama-inst}
\end{figure}

\paragraph{Variance Split of Lens Information Gaps.}
For $g_R$, we report lexical occupancy, the fraction of questions whose sampled outputs vary in surface form within one meaning. On TriviaQA, the lexical occupancy is 60.7\% for Qwen2.5-7B-Instruct, 67.1\% for Qwen2.5-7B (base), and 58.9\% for Llama-3.1-8B-Instruct. Lexical variation is most frequent in the base model, whose outputs have not been regularized.

To decompose the $g_\text{unemb}$ gap, we compute the thin SVD of the unembedding matrix $W_U$ through the Gram matrix $G = W_U^\top W_U$. We select the smallest rank $r$ that explains $99\%$ of the row-space variance. The hidden state is decomposed into $h_\text{top}$, its projection onto the top-$r$ right singular vectors, and the tail $h_\text{tail} = h - h_\text{top}$. The tail lies in the low-singular-value subspace of $W_U$ and are weakly expressed in the output. Table~\ref{tab:poc-variance-split} reports this decomposition.

\begin{table}[htbp]
\centering
\caption{Hidden-state variance decomposition on TriviaQA\@. Share of $\lVert h \rVert^2$ in the top-$r$ and tail singular subspaces of $W_U$, with retained rank $r$ relative to hidden dimension $d$.}\label{tab:poc-variance-split}
\vspace{4pt}
\small
\begin{tabular}{l c c c}
\toprule
\textbf{Model} & $h_\text{top}$ \textbf{fraction} & $h_\text{tail}$ \textbf{fraction} & \textbf{rank} $r$ / \textbf{dim} $d$ \\
\midrule
Qwen2.5-7B-Instruct   & $77.7\%$ & $22.3\%$ & $3439$ / $3584$ \\
Qwen2.5-7B (base)     & $77.7\%$ & $22.3\%$ & $3444$ / $3584$ \\
Llama-3.1-8B-Instruct & $95.6\%$ & $4.4\%$  & $3962$ / $4096$ \\
\bottomrule
\end{tabular}
\end{table}

The two Qwen variants have nearly identical decompositions. Instruction tuning does not reshape this partition in the Qwen family. Llama places more of its hidden-state norm in the top-$r$ component.

\subsection{Evaluation Criteria Description}\label{sec:appendix-criteria}

This appendix describes the evaluation criteria in Section~\ref{sec:eval-criteria} in detail. The distributional criteria measure lexical correctness, while the semantical criteria measure semantic correctness. 

\paragraph{Distributional Criteria} are used in natural language generation evaluation. 
\begin{itemize}
    \item \textbf{Exact match} considers an output correct only when it exactly matches the reference string. 
    \item \textbf{BLEU} measures shared $n$-grams between an output and reference, with a penalty for overly short outputs~\citep{papineniBleuMethodAutomatic2002}.
    \item \textbf{ROUGE-1/2/L} measures overlap between an output and reference using unigrams, bigrams, or longest common subsequences~\citep{linROUGEPackageAutomatic2004}.
\end{itemize}

Other less-often used distributional criteria include Word Error Rate (WER), which counts word-level insertions, deletions, and substitutions~\citep{morrisWERRILMER2004} towards reference, METEOR, which aligns words using stemming, and synonym matches~\citep{banerjeeMETEORAutomaticMetric2005}, and SQuAD F1, which computes token-level precision and recall between the output and reference~\citep{rajpurkarSQuAD100000Questions2016}.

\paragraph{Semantical Criteria} concern the meaning of output.
\begin{itemize}
    \item \textbf{Option match} checks whether the selected option is the same as the reference option. It is a special case of exact match restricted to the MCQA setting. Since the answer collapses to a single option letter, the match verifies meaning rather than surface form.
    \item \textbf{NLI entailment} uses a natural-language-inference model to find bidirectional entailment between the output and the gold reference~\citep{williamsBroadCoverageChallengeCorpus2018}.
    \item \textbf{LLM-as-judge} prompts an LLM to judge directly whether the output is correct or grounded in the supplied context~\citep{zhengJudgingLLMasaJudgeMTBench2023}.
    \item \textbf{FActScore} decomposes the output into atomic factual claims and checks each against a knowledge source, reporting the fraction supported as a continuous score~\citep{minFActScoreFinegrainedAtomic2023}.
    \item \textbf{AlignScore} trains an alignment function across several factual-consistency tasks and applies it to score an output against a reference~\citep{zhaAlignScoreEvaluatingFactual2023}.
    \item \textbf{Human annotation} has annotators judge directly whether the output is correct or grounded in the context~\citep{wangCLUEConceptLevelUncertainty2024}.
\end{itemize}

Other semantical criteria include BERTScore~\citep{zhangBERTScoreEvaluatingText2020}, FactCC~\citep{kryscinskiEvaluatingFactualConsistency2020} for source consistency, BLEURT~\citep{sellamBLEURTLearningRobust2020} for learned text-quality prediction, COMET~\citep{reiCOMETNeuralFramework2020} for learned translation-quality estimation, XCOMET~\citep{guerreiroXcometTransparentMachine2024} for translation scoring with error localization, and MetricX~\citep{juraskaMetricX24GoogleSubmission2024} for learned translation-quality scoring.

\subsection{Evaluation Metrics Description}\label{sec:appendix-metrics}

This appendix describes the evaluation metrics in Section~\ref{sec:eval-metrics} in detail, organized in seven families. In the paragraphs below, $c = C(y,y^*) \in \{0,1\}$ denotes the binary correctness or grounding label from Section~\ref{sec:eval-criteria}, and $U(x)$ denotes the uncertainty score, with higher $U$ meaning more uncertain. We omit the lens subscript $\lambda$ for brevity.

\paragraph{Ranking \irnk} measures the ability of uncertainty estimators to rank correct answers before incorrect ones.
\begin{itemize}
    \item \textbf{AUROC} is the area under the receiver operating characteristic curve. $\mathrm{AUROC}$ is estimated as the area under the curve traced by the true-positive rate and false-positive rate for varying decision thresholds. The AUROC represents the probability that a randomly selected positive instance is ranked higher than a randomly selected negative instance.
    \item \textbf{AUARC} is the area under the accuracy--rejection curve. It orders samples by uncertainty score $U(x)$, rejects an increasing fraction of the least certain, and plots the accuracy of the retained samples against the rejected fraction.
    \item \textbf{AURC} is the area under the risk--coverage curve, which plots the error rate of the retained samples against the retained fraction. Since the error rate is one minus the accuracy, $\mathrm{AURC} = 1 - \mathrm{AUARC}$, and the two metrics carry the same information~\citep{nadeemAccuracyRejectionCurvesARCs2009}.
    \item \textbf{AUPRC} is the area under the precision--recall curve, preferred over AUROC on imbalanced datasets.
    \item \textbf{PRR} is the prediction rejection ratio. It normalizes the model's risk--coverage curve against a random-rejection baseline and an oracle that rejects every incorrect answer first. The PRR is defined as $\mathrm{PRR} = \frac{\mathrm{AURC} - \mathrm{AURC}_{rnd}}{\mathrm{AURC}_{\text{oracle}} - \mathrm{AURC}_{\text{rnd}}}$.
\end{itemize}
Some other ranking metrics report a single point on one of these curves. C@Acc is the coverage rate at a given accuracy threshold~\citep{coleSelectivelyAnsweringAmbiguous2023}. FPR@\%C is the false positive rate at a fixed coverage level. TPR@k\%FPR is the true positive rate at a fixed low false positive rate. Both are single points on the ROC curve. 

\paragraph{Calibration \ical} measures whether the uncertainty score can be read as the probability of positive responses. 
\begin{itemize}
    \item \textbf{ECE} is the expected calibration error. It partitions predictions into equal-width confidence bins $B_1,\dots, B_M$ and averages the gap between confidence and accuracy within each bin, weighted by bin size. $\mathrm{ECE} = \sum_{m=1}^M \frac{|B_m|}{n}\,\big|\mathrm{acc}(B_m) - \mathrm{conf}(B_m)\big|$.
    \item \textbf{MCE} is the maximum calibration error, which takes the maximum of the absolute differences. $\mathrm{MCE} = \max_m \big|\mathrm{acc}(B_m)-\mathrm{conf}(B_m)\big|$, reporting the single worst-calibrated bin.
    \item \textbf{ACE} is the adaptive calibration error. It computes the same quantity as ECE with an equal number of samples in each bin, avoiding the noise induced by sparse bins.
\end{itemize}
The less-used calibration metrics include ROB, the ratio of overconfidence bias, which is the fraction of incorrect answers whose confidence exceeds a fixed high threshold (e.g., 80\%).

\paragraph{Scoring \iscr} measures the quality of the uncertainty score as a probabilistic prediction.
\begin{itemize}
    \item \textbf{Brier score} is the mean squared error between the predicted probability of correctness $1-U(x)$, with $U$ normalized to $[0,1]$, and the true label, $\mathrm{BS} = \frac{1}{n}\sum_i \big(1-U(x_i) - c_i\big)^2$. 
    \item \textbf{CRPS} is the continuous ranked probability score, comparing the predicted cumulative distribution function (CDF) $F$ with the true outcome $y$. $\mathrm{CRPS} = \int_{-\infty}^{\infty} \big(F(z) - \mathbbm{1}[z\ge y]\big)^2\,dz$.
\end{itemize}

\paragraph{Decision \idec} measures the ability of the uncertainty score to make binary decisions for downstream tasks.
\begin{itemize}
    \item \textbf{Accuracy} is the fraction of correctly classified samples. $\mathrm{Acc} = \frac{1}{n}\sum_i \mathbbm{1}\big[d(x_i) = c_i\big]$, for a decision rule $d(x) = \mathbbm{1}[U(x) < \tau]$ at a fixed threshold $\tau$.
    \item \textbf{F1 score} is the harmonic mean of precision $P = TP/(TP+FP)$ and recall $R = TP/(TP+FN)$ at the same threshold, where $TP$ is the number of true positives, $FP$ for false positives, and $FN$ for false negatives. $F1 = 2PR/(P+R)$.
\end{itemize}
Other classification metrics are used the same way, including precision, recall, balanced accuracy for imbalanced datasets, and weighted-F1. There are also task-specific metrics like Pass@k for code generation~\citep{xuUncertaintyQuantificationLLMbased2026}. All decision metrics are threshold-dependent.

\paragraph{Regression \ireg} measures the correlation between the score and a continuous target. 
\begin{itemize} 
    \item \textbf{PCC} is the Pearson correlation coefficient. It is the linear correlation between the uncertainty score and the continuous target $Y$. $r = \mathrm{Cov}(U(x),Y)/(\sigma_U\sigma_Y)$. 
    \item \textbf{SCC} is Spearman's rank correlation coefficient. It applies the Pearson formula to the ranks of $U(x)$ and $Y$ instead of their raw values, asking whether a higher score corresponds to a higher criterion label.
\end{itemize}
Other regression metrics include MSE, the mean squared gap between $U(x)$ and the label.

\paragraph{Coverage \icov} measures the conformal prediction guarantee of uncertainty methods.
\begin{itemize}
    \item \textbf{Coverage probability} is the fraction of test samples whose true label falls inside the prediction set. It is an empirical verification of the theoretical guarantee over actual data. 
    \item \textbf{Average prediction set size} is the mean size of the prediction set. It is an efficiency measure, since a method can trivially reach full coverage by including all the labels in the set.
\end{itemize}
Size-stratified coverage conditions the coverage rate on the realized set size. AUC over $\alpha$ calculates the area under the curve of set size against the miscoverage level $\alpha$. Claim-level conformal methods also report the fraction of claims retained.

\paragraph{Faithfulness \ifth} measures the alignment between two uncertainty scores.
\begin{itemize}
    \item \textbf{Rank correlation} applies Spearman's correlation between the two estimates directly.
    \item \textbf{cMFG}~\citep{yonaCanLargeLanguage2024} is the conditional mean faithful generation. $\mathrm{cMFG} = \mathbb{E}_{i,v} \big[\mathtt{f}_M(y_i;x_i) \mid \mathrm{conf}_M(y_i;x_i) = v\big]$ where $\mathtt{f}_M(y; x)$ is one minus the gap between how decisively the model expresses $y$ for input $x$ and how confident it actually is, as estimated by sampling.
    \item \textbf{CoT intervention}~\citep{lanhamMeasuringFaithfulnessChainofThought2023} examines how the model predictions change when the chain-of-thought is intervened. 
\end{itemize}

\end{document}